\documentclass{article} % For LaTeX2e
\usepackage{iclr2027_conference,times}

\usepackage{amsmath,amsfonts,bm}

\def\eqref#1{equation~\ref{#1}}
\def\1{\bm{1}}

\DeclareMathAlphabet{\mathsfit}{\encodingdefault}{\sfdefault}{m}{sl}
\SetMathAlphabet{\mathsfit}{bold}{\encodingdefault}{\sfdefault}{bx}{n}

\usepackage{hyperref}
\usepackage{graphicx}
\usepackage{url}
\usepackage{booktabs}

\usepackage{wrapfig}

\usepackage{booktabs}
\usepackage{multirow}
\usepackage{xcolor}
\usepackage{pifont}
\usepackage{fontawesome5}
\usepackage{booktabs}
\usepackage{multirow}
\usepackage[table]{xcolor}
\usepackage{array}

\usepackage{titletoc} 
\usepackage{hyperref} 

\usepackage[most]{tcolorbox}
\usepackage{inconsolata}

\newtcolorbox{promptbox}[1][]{title=#1,
    enhanced,
    breakable,
    colback=gray!4,
    colframe=gray!45,
    boxrule=0.5pt,
    arc=1.5mm,
    left=7mm,
    right=5mm,
    top=3mm,
    bottom=3mm,
    fontupper=\small\ttfamily,
    title=#1,
    coltitle=black,
    colbacktitle=gray!15,
    fonttitle=\small\bfseries,
    attach boxed title to top left={
        yshift=-2mm,
        xshift=2mm
    },
    boxed title style={
        boxrule=0pt,
        arc=1mm,
        left=1.5mm,
        right=1.5mm,
        top=0.5mm,
        bottom=0.5mm
    }
}

\title{LIBERO-RECOVER: Beyond Task Success Towards Failure Recovery in Robotic Manipulation Models}

\author{
Lin Liu \\ School of Information and Communication Engineering \\
Dalian University of Technology \& Beta Infinity
\And
Lu Zhang\thanks{Corresponding author.} \& Huchuan Lu \\
School of Information and Communication, \\
Engineering Dalian University of Technology
\And
Wu Yang, Yuzheng Zhuang, \\
\textbf{Shuai Tao} \& \textbf{Wulong Liu} \\
Beta Infinity
\And
Ziying Song\thanks{Corresponding author.} \\
Nanyang Technological University
\And
Zhicheng Bao \& Caiyan Jia \\
Beijing Jiaotong University
}

\iclrfinalcopy % Uncomment for camera-ready version, but NOT for submission.
\begin{document}

\maketitle

\begin{abstract}
Vision-Language-Action (VLA) or World Action (WAM) models have recently demonstrated remarkable performance in robotic manipulation. On LIBERO, SOTA method have achieved nearly 100\% success rates, seemingly suggesting that the models are ready for deployment in real world. However, near perfect performance on existing benchmarks can be misleading: success under ideal conditions does not imply real world robustness. Existing benchmarks primarily evaluate task completion from predefined initial states, while real world interactions inevitably involve failures such as failed grasps, collisions, and unintended object movements. A robot must therefore not only execute tasks successfully, but also recognize and recover from failures to continue the task. Yet this capability remains largely unmeasured, revealing a critical gap between benchmark performance and real world reliability. To address this gap, we introduce LIBERO-Recover Benchmark, a large scale benchmark for failure recovery in robotic manipulation. Built upon LIBERO, we collect real execution failures from SOTA embodied models and construct 2,000+ scenarios across four recovery levels: (1) Action Retry, (2) Action Adaptation, (3) Object State Recovery, and (4) Environmental Recovery. We evaluate four core capabilities: spatial understanding, object structure reasoning, interaction understanding, and topological reasoning. As the first large-scale benchmark for embodied failure recovery, LIBERO-Recover shifts evaluation from \emph{Can the robot succeed?''} to \emph{Can the robot recover after failure?''}, promoting robust and generalizable embodied agents. The project is available in https://liulin815.github.io/LIBERO-Recovery.
\end{abstract}
% 基于LIBERO，我们从SOTA模型中收集故障轨迹，包括π、GR00T和Cosmos-Policy，并在四个难度级别构建1000个场景:(1)动作重试，(2)动作适应，(3)对象状态恢复和(4)阻碍任务完成的环境恢复。我们进一步评估了四个核心功能：实施例理解、对象结构推理和拓扑交互推理。FAIL-RECOVER 将评估从“机器人能否成功？”转移到“机器人发生故障后是否恢复？”，为评估和改进具身代理的鲁棒性提供了新的测试平台。

\section{Introduction}
% 单个物体失败Vision-Language-Action (VLA) 模型最近已成为连接具身代理中的感知、语言理解和动作执行的基本范式 [17]。该领域进展的中心催化剂是开发标准化基准[7，13，15，12，16，19]，这使得竞争方法的可重现、公平和定量比较成为可能。其中，LIBERO [13] 迅速成为 VLA 最广泛使用的评估套件，作为性能报告的事实标准。如图6所示，大规模预训练模型和任务特定系统都在LIBERO上进行了基准测试，使其成为该领域评估的“普通货币”。因此，LIBERO 的方法严谨性和可靠性直接影响如何定义和追求 VLA 中的研究方向
% 近年来，视觉语言动作（Vision-Language-Action, VLA）及世界动作模型（World Action Model, WAM）在机器人操作中取得了令人瞩目的进展。尤其是在早期的操作基准 LIBERO 上，先进模型的成功率已逼近 100%，甚至已有工作不再报告 LIBERO 的结果，认为其任务已难以有效区分模型性能。然而，无论是早期的 LIBERO、Simpler-Env，还是后续的 RoboTwin 2.0、RoboCasa，现有基准的核心评测范式始终是：从预定义的初始状态出发，模型能否一次性成功完成任务。这种评测主要衡量模型在相对理想、受控条件下的任务执行能力，而与真实世界的机器人部署存在明显差距。现实环境中的操作具有高度不确定性，抓取失败、碰撞、物体偏移等情况不可避免，即使人类也无法保证每次操作都一次成功。因此，对于真正面向现实环境的机器人而言，关键问题不应仅仅是“能否完成任务”，更应是**“任务失败后，能否理解当前状态、恢复受损环境，并继续完成任务”。我们认为，这种失败后的持续任务能力（failure recovery）**
VLAs and WAMs have achieved remarkable progress in robotic manipulation. On early benchmarks such as LIBERO~\cite{libero}, SOTA models have approached 100\% success rates, with some recent works even omitting LIBERO results~\cite{libero}, as its tasks are no longer sufficiently challenging to differentiate model capabilities. Yet, from LIBERO~\cite{libero} and SimplerEnv~\cite{simplenv} to RoboTwin2.0~\cite{robotwin} and RoboCasa~\cite{robocasa}, the prevailing evaluation paradigm remains largely unchanged: models start from predefined initial states and are evaluated on single-attempt task completion. This setting assumes largely idealized execution, whereas real-world manipulation inevitably involves failed grasps, collisions, and unintended object displacement. Robust robots must therefore not only complete tasks, but also recognize failures, recover disrupted states, replan, and resume execution.

\begin{figure*}[!t]
\centering
\includegraphics[width=1.0\textwidth]{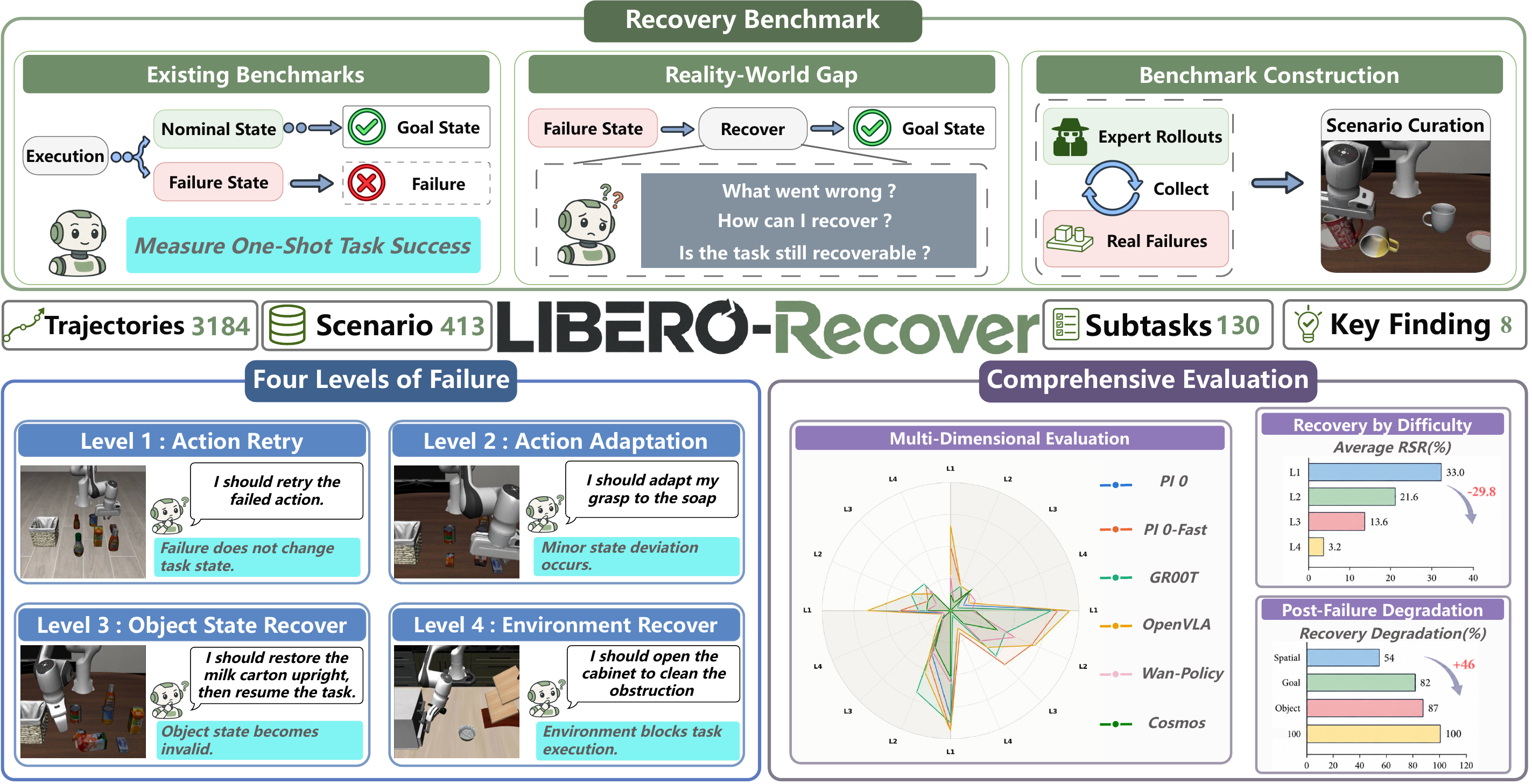}
\caption{\textbf{Overview of LIBERO-Recover.} LIBERO-Recover collects failure scenarios from SOTA models and expert recovery trajectories via human teleoperation. The plots summarize recovery performance across difficulty levels and performance degradation after failure.}
\label{fig:motivation}
\end{figure*}

Recent benchmarks have extended LIBERO toward more challenging and realistic settings. LIBERO-Pro~\cite{libero-pro} introduces more complex task variations, LIBERO-Plus~\cite{libero-plus} evaluates robustness across multiple dimensions, LIBERO-X~\cite{libero-x} considers variations in task and environmental conditions, and LIBERO-Safety~\cite{liberosafety} focuses on safety-critical failures and undesired interactions.

% 失败下重新规划任务并完成这一能力，本质上可以视为考核具身操作模型鲁棒性的一部分，近年来在已经存在工作进行过扩展和研究，例如：LIBERO-pro,LIBERO-plus,LIBERO-X以及LIBERO-Safety
% 然而，现有 benchmark 通常仅通过改变物体的初始位置或外观来构造测试场景，仅人工构造，而并未真实模拟任务执行过程中真实产生的失败状态。因此，这类评测本质上考察的是模型对物体位置、外观及环境变化的泛化能力，而非面对已经发生的操作失败时，能否理解受损状态、采取纠正动作并恢复任务执行。
However, these benchmarks primarily assess generalization under predefined perturbations or challenging conditions, rather than recovery from naturally occurring execution failures. Failure recognition, state recovery, and post-failure replanning therefore remain insufficiently evaluated. To address this gap, we propose LIBERO-Recover, a benchmark that collects failure states and their consequences from embodied model executions to evaluate failure recovery. Unlike existing benchmarks that construct test environments by manually perturbing object poses, appearances, or scene layouts, LIBERO-Recover naturally collects failure states and their consequences arising from real embodied models. Its evaluation protocol assesses four core capabilities—embodiment understanding, object structure reasoning, topological interaction reasoning, and failure recovery—across 130 subtasks and 16 evaluation dimensions. Experiments show that existing VLA models' failure recovery capabilities remain substantially limited, revealing a significant gap between high task success rates and robustness under real-world interactions. Our key contributions are summarized as follows.

\begin{itemize}
    \item We show that LIBERO evaluates one-shot task success under idealized conditions, potentially overestimating the robustness of embodied models to real-world failures.

    \item We introduce \textbf{LIBERO-Recover}, a failure recovery benchmark built from naturally occurring failures during embodied model interactions, covering \textbf{4 task categories, 130 subtasks, and 16 level combinations} to assess four core recovery capabilities.

    \item We evaluate leading VLA models and reveal a substantial performance gap between standard task execution and failure recovery, highlighting critical limitations in failure-state understanding and recovery planning.
\end{itemize}

\section{LIBERO-RECOVER Benchmark}
\label{main}

\subsection{Problem Formulation}
We consider a manipulation task specified by a language instruction $l$, with a goal condition $\mathcal{G}$. Starting from an initial state $s_0$, a manipulation policy $\pi_\theta$ generates a sequence of actions based on the robot's observations, resulting in an execution trajectory:
\begin{equation}
\tau = (s_0,a_0,s_1,a_1,\ldots,s_T).
\end{equation}

\textbf{Conventional Task Execution}. Standard evaluation measures whether a policy completes the task from a valid initial state $s_0$, reaching a terminal state $s_T$ with $\mathcal{G}(s_T,l)=1$, without separately assessing post-failure recovery.

\textbf{Failure Recovery}. Our setting instead considers an execution that deviates from the intended task progression and reaches an intermediate state $s_f$. We refer to $s_f$ as a \emph{failure state} when the original task goal is not satisfied at this point,
\begin{equation}
\mathcal{G}(s_f,l)=0,
\end{equation}
but the task remains recoverable through subsequent interaction. The key difference is that the policy can no longer simply continue the original action sequence. Instead, it must determine how the preceding failure has altered the current task configuration and generate a new sequence of actions that restores the necessary state before continuing toward the original goal. We therefore decompose the execution into a pre-failure trajectory and a recovery trajectory:
\begin{equation}
s_0 \rightarrow \cdots \rightarrow s_f
\rightarrow
s_{f+1} \rightarrow \cdots \rightarrow s_T,
\end{equation}
where the post-failure actions constitute a recovery policy $\pi_{\mathrm{rec}}$. A failure state is considered \emph{recoverable} if:
\begin{equation}
\mathcal{G}(s_f,l)=0
\quad\text{and}\quad
\exists,\pi_{\mathrm{rec}}
\ \text{s.t.}
\mathcal{G}(s_T,l)=1.
\end{equation}
Failure recovery requires robots to understand, adapt, and recover from unexpected states.

\subsection{Failure or State Variation?}
\label{sec:failure_def}

Not every state change during execution constitutes a failure. We define a state $s_f$ as a failure state when the preceding action induces a task-relevant change that invalidates the current execution plan:
\begin{equation}
\mathcal{F}(s_t,a_t,s_f)
\iff
\underbrace{s_t\xrightarrow{a_t}s_f}_{\text{execution-induced}}
\land
\underbrace{\mathcal{G}(s_f,l)=0}_{\text{task-relevant}}
\land
\underbrace{\pi_\theta(s_f)\not\Rightarrow\mathcal{G}}_{\text{plan-invalidating}}.
\end{equation}
Here, the last condition means that directly continuing the current execution is insufficient to complete the original task, requiring adaptation or recovery. A valid failure state must nevertheless remain recoverable,
\begin{equation}
\exists\,\pi_{\mathrm{rec}}
\quad\text{s.t.}\quad
s_f \xrightarrow{\pi_{\mathrm{rec}}} s_T,
\quad
\mathcal{G}(s_T,l)=1.
\end{equation}
Thus, state changes that preserve task executability are treated as normal variations, whereas execution-induced changes that invalidate the current plan and require recovery are treated as failures.

\subsection{Failure Taxonomy and Difficulty Levels}
\label{sec:taxonomy}

\begin{figure*}[t]
\centering
\includegraphics[width=1.0\textwidth]{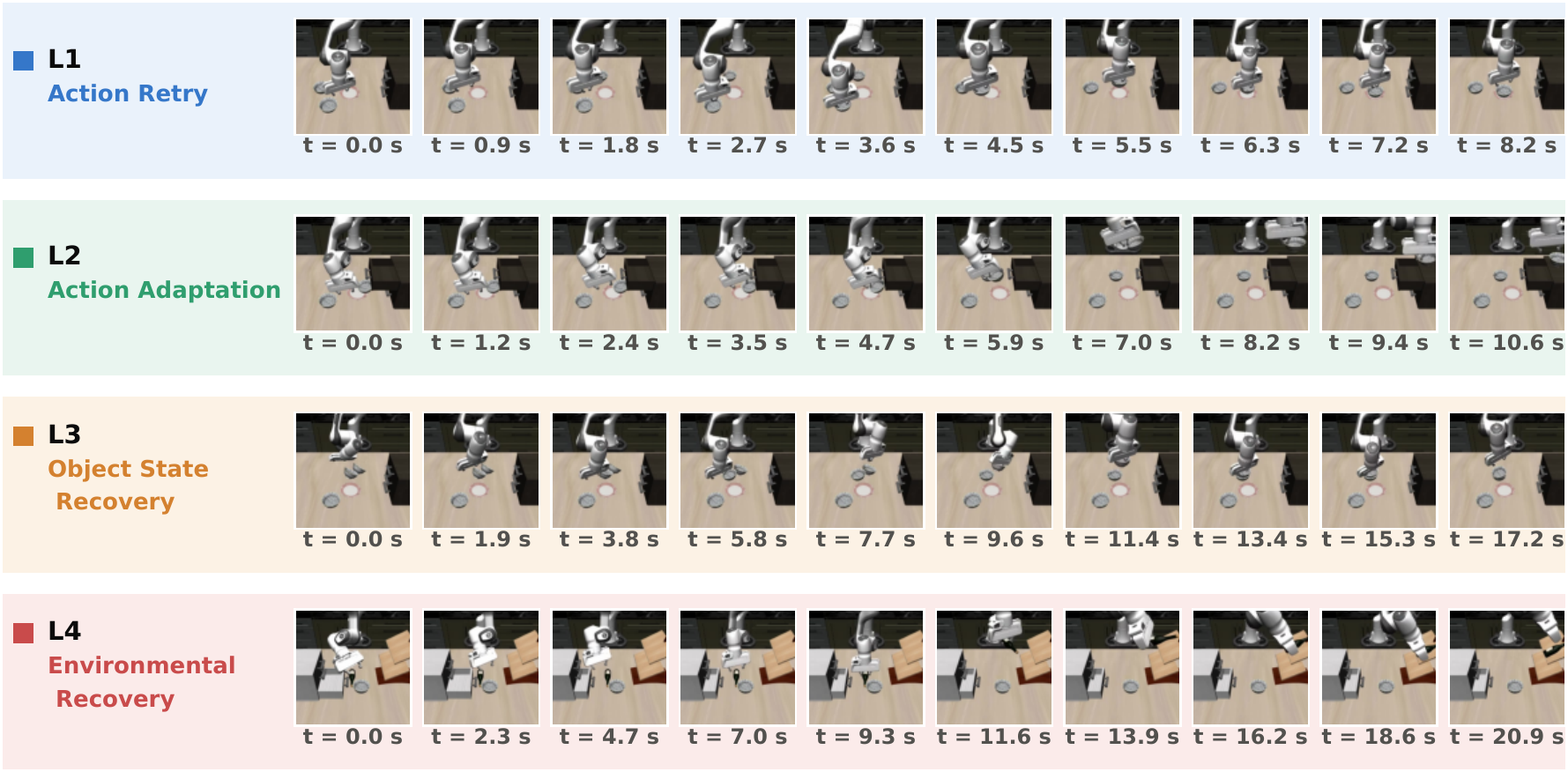}
\caption{\textbf{Four recovery difficulty levels of LIBERO-Recover.} Each level demands progressively deeper state reasoning, from re-executing a failed action ($L_{1}$) to restoring environmental states that block task completion ($L_{4}$).}
\label{fig:taxonomy}
\end{figure*}

We organize recoverable failures into four difficulty levels according to the corrective actions and state reasoning required to resume task execution, as illustrated in Fig.~\ref{fig:taxonomy}. The hierarchy distinguishes failures that can be resolved by retrying or adapting actions from those requiring object state restoration or the removal of environmental obstructions.

\textbf{$L_{1}$ Action Retry}. The failure does not materially alter the task configuration, i.e., $s_f\approx s_{\mathrm{expected}}$. Recovery therefore only requires retrying the failed action, such as reattempting a grasp when the target object remains unchanged.

\textbf{$L_{2}$ Action Adaptation}. The failure causes a limited change in the object configuration, such that $s_f\neq s_{\mathrm{expected}}$, but the object remains directly usable. The model must adapt its action to the observed state rather than blindly repeating the failed action:
\begin{equation}
a'=\pi(o_f,l)\neq a_{\mathrm{failed}}.
\end{equation}

\textbf{$L_{3}$ Object State Recovery}. The failure substantially alters a task-relevant object, making direct task completion impossible. The model must first restore the object's required state and then resume the original task. For example, after a failed grasp displaces a bowl, the robot must first recover the bowl's pose before continuing the instructed manipulation.

\textbf{$L_{4}$ Environmental Recovery}. The failure affects an object or state that is not directly manipulated by the original task but subsequently prevents task execution. Recovery therefore requires reasoning about both task-relevant and task-irrelevant states and their interaction topology. The model must identify the environmental obstruction, restore the configuration, and resume the original task.
\subsection{Recovery Scenario Construction}
\label{sec:construction}

\textbf{Scenario Definition.} Rather than manually perturbing task states to synthesize failures, LIBERO-Recover grounds every scenario in a failure that actually occurred during embodied model execution. Formally, each recovery scenario is defined as
\begin{equation}
(\mathcal{I},s_0,\tau_{\mathrm{fail}},s_f,g,r),
\end{equation}
where $\mathcal{I}$ denotes the task instruction, $s_0$ the original initial state, $\tau_{\mathrm{fail}}$ the failure trajectory, $s_f$ the resulting failure state, $g$ the original task goal, and $r$ the required recovery behavior. 

\begin{figure*}[t]
\centering
\includegraphics[width=1.0\textwidth]{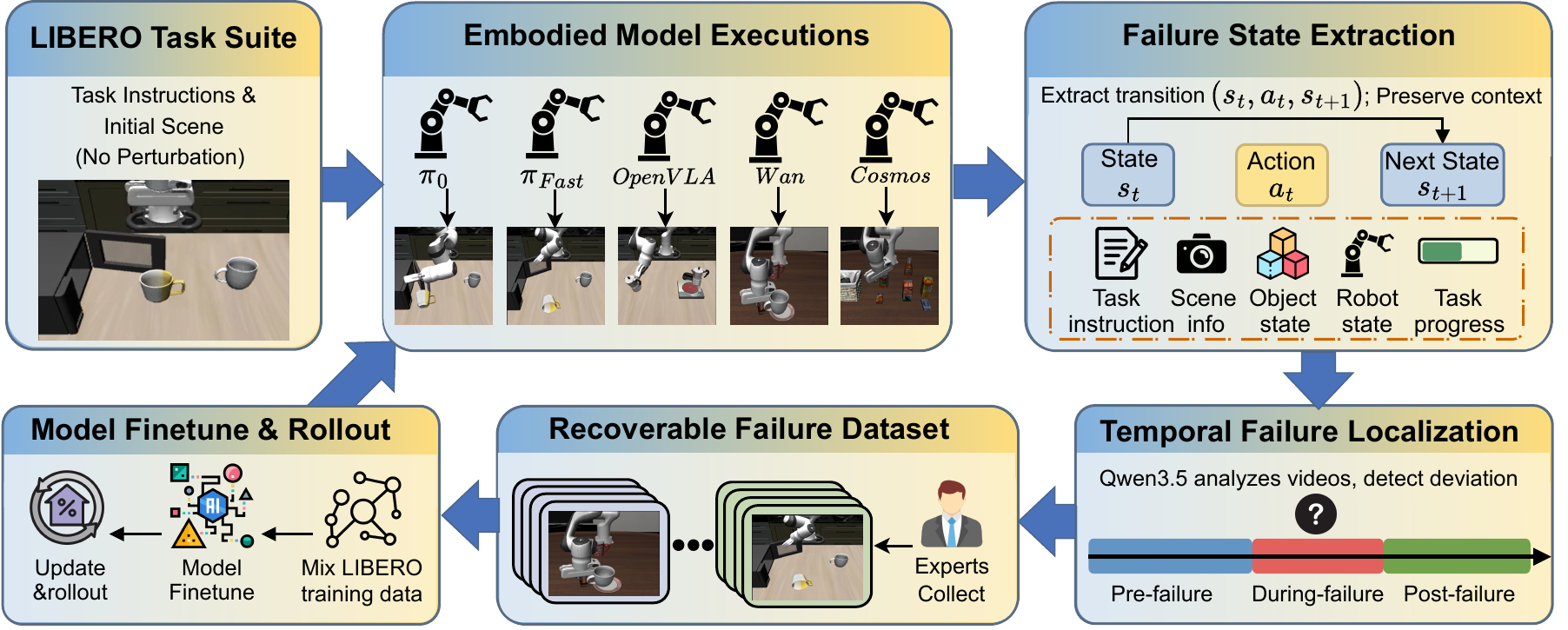}
\caption{The construction pipeline of LIBERO-Recover.}
\label{fig:pipeline}
\end{figure*}

\textbf{Construction Pipeline.} As illustrated in Fig.~\ref{fig:pipeline}, each scenario is constructed through a four-stage pipeline, with the source tasks, failure generators, and annotation models introduced stage by stage.

% 阶段 1：任务执行。我们从涵盖 130 个子任务的四个 LIBERO 任务套件开始，保留其原始指令和初始构型。部署了包括 $\pi_{0}$、$\pi_{0.5}$、OpenVLA、GR00T、Wan-Policy 和 Cosmos-Policy 在内的多种具身策略作为“失败生成器”。从 $s_0$ 开始的每次执行都会产生一条完整的失败轨迹 $\tau_{\mathrm{fail}}$，仿真器状态被渲染为视频以供后续时序分析。执行期间未对物体、环境状态或初始构型进行任何人工扰动，也未注入任何失败。
\textbf{Stage 1: Task Execution.} We start from four LIBERO task suites covering 130 subtasks, preserving their original instructions and initial configurations. Multiple embodied policies, including $\pi_{0}$, $\pi_{0.5}$, OpenVLA, GR00T, Wan-Policy, and Cosmos-Policy, are deployed as \emph{failure generators}. Each execution from $s_0$ produces a complete failure trajectory $\tau_{\mathrm{fail}}$, and the simulator states are rendered into videos for subsequent temporal analysis. No objects, environment states, or initial configurations are manually perturbed, and no failures are injected during execution.

\textbf{Stage 2: Failure Localization.} For each execution video, we employ Qwen3.5-27B to analyze the entire execution together with the task instruction and to temporally localize where the execution deviates from the intended progression. Each video is decomposed into three stages,
\begin{equation}
V=V_{\mathrm{pre}}\cup V_{\mathrm{fail}}\cup V_{\mathrm{post}},
\end{equation}
corresponding to the execution before, during, and after the failure-causing interaction. Based on the predicted temporal boundaries, we retrieve the corresponding robot simulation states and object poses directly from the simulator, and extract the localized failure scenario.
\begin{wrapfigure}{r}{0.40\textwidth}
    \centering
    \includegraphics[width=0.40\textwidth]{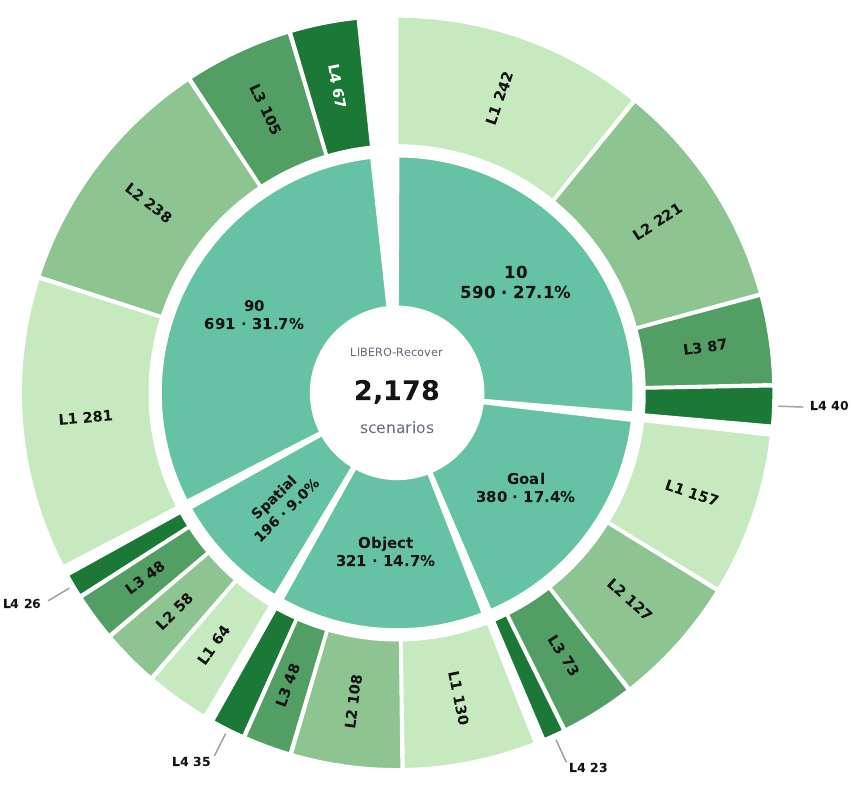}
    \caption{Distribution of LIBERO-Recover scenarios across task suites and recovery difficulty levels, including LIBERO-Spatial, LIBERO-Object, LIBERO-Goal, LIBERO-10 (LIBERO-100) and LIBERO-90 (LIBERO-100). For detailed analysis, LIBERO-100 is further partitioned into the LIBERO-10 and LIBERO-90 subsets.}
    \label{fig:stats}
\end{wrapfigure}
\textbf{Stage 3: Failure Characterization.} For each candidate failure scene, we render its execution trajectory into a video and provide the video and corresponding task instruction to Qwen3.5-27B. The model is then prompted to assess the post-failure consequences and determine the corresponding recovery type and difficulty level according to the taxonomy defined in Sec.~\ref{sec:taxonomy}.

\textbf{Stage 4: Model Finetune \& Rollout.} We fine-tuned the model using manually collected recovery data from failure scenarios, and deployed it under the original LIBERO settings to collect further failure trajectories. Scenarios sampled for recovery data were excluded from the final evaluation set.

\subsection{Benchmark Statistics}
\label{sec:stats}
% 10 578，goal 368，object 379，spatial 184，90 678
% 10 239, goal 154, object 127, spatial 61, 90 277
% 10 218, goal 124, object 105, spatial 55, 90 235
% 10 84, goal 70, object 45, spatial 45, 90 102
% 10 37, goal 20, object 32, spatial 23, 90 64
% 整体上，5个task内，四个level呈现不均匀分布，其中难度越高占据整体的比例越低，这是来自真机模型真实采集的结果，为了保证数据分布和模型真实执行的分布保持一致，我们没有进行认为的调整比例。此外，在难度上升的同时，对应难度下，Post-Task-Failure比例对应上升，这是由于在失败后模型不具备恢复能力，而在反复重试，通常会导致更严重的后果
The resulting benchmark contains \textbf{2178} recovery scenarios derived from \textbf{130 subtasks} across \textbf{four task categories}, covering \textbf{four recovery levels} and \textbf{16 evaluation dimensions}. In total, failures are generated by \textbf{six embodied models}. The overall benchmark statistics are summarized in Fig.~\ref{fig:stats}. Across the five evaluation suites, the four recovery levels exhibit an inherently imbalanced distribution, with higher-difficulty failures occurring less frequently. This distribution emerges naturally from real-world model executions, and we therefore preserve it without manual rebalancing to faithfully reflect the empirical failure distribution. Notably, the proportion of Post-Task-Failure cases increases with recovery difficulty, as models lacking recovery capability tend to repeatedly retry failed actions, often leading to progressively more severe consequences. 

\section{Datasets and Evaluation Metrics}
\label{sec:dataset_con}
% 为了提升模型从失败中恢复的能力，我们招募了4名人类遥操员，从413失败恢复场景中利用spacemouse采集了3184条恢复的轨迹，其中413个场景由benchmark中的场景中5个task中均匀采样得到，来保证分布一致、
\subsection{LIBERO-Recover Dataset}
To improve failure recovery, four human teleoperators collected 3,184 recovery trajectories across 413 failure scenarios using a SpaceMouse, uniformly sampled from the five task suites. These trajectories are used for model fine-tuning. Finetuned models are used to re-generate failure scenarios.
\subsection{Evaluation Metrics}
To assess models' ability to handle failures, we propose the following metrics for evaluation:

\textbf{Recovery Success Rate (RSR)}, which measures whether a policy can successfully complete the original task after entering a failure state. For $N$ failure scenarios, we define
\begin{equation}
\mathrm{RSR}
=
\frac{1}{N}
\sum_{i=1}^{N}
\mathbb{I}\left[\mathcal{G}(s_T^i,l_i)=1\right],
\end{equation}
where $s_T^i$ denotes the terminal state of the recovered episode. A higher RSR indicates stronger failure recovery capability. To characterize the effect of recovery difficulty, we additionally report
$\mathrm{RSR}_{L_{1}}$, $\mathrm{RSR}_{L_{2}}$, $\mathrm{RSR}_{L_{3}}$, and $\mathrm{RSR}_{L_{4}}$ for the four recovery levels.

\textbf{Recovery Degradation (RD)}. To quantify execution capability lost after failure, we divide each episode into three phases: \emph{Pre-Failure}, \emph{During-Failure}, and \emph{Post-Failure}. Let $M_{\mathrm{pre}}$ and $M_{\mathrm{post}}$ denote the execution performance before and after failure. We define \emph{Recovery Degradation (RD)} as
\begin{equation}
\mathrm{RD}
=1-
\frac{M_{\mathrm{post}}+\epsilon}
{M_{\mathrm{pre}}+\epsilon},
\end{equation}
where $\epsilon$ ensures numerical stability. Lower RD indicates better retention of execution capability, while higher RD indicates greater degradation.

\textbf{Recovery Consistency (RC)}, which measures the stability of recovery across different failure states within the same task. For each task $k$, let $R_k$ denote its recovery success rate across all associated failure scenarios. We define
\begin{equation}
\mathrm{RC}
=
1-
\operatorname{Std}_{k}(R_k),
\end{equation}
where $\operatorname{Std}_{k}(\cdot)$ denotes the standard deviation of task-level recovery success rates. Higher RC indicates more consistent recovery across failure states.

\section{EXPERIMENTS}
\label{exp}

\subsection{EXPERIMENTS SETUP}
We evaluate six representative models: OpenVLA-OFT~\cite{openvla}, $\pi_0$-Fast~\cite{pi0_fast}, GR00T-N1.5~\cite{gr00t}, $\pi_0$~\cite{pi_0}, Wan2-Policy~\cite{wan}, and Cosmos-Predict2-Policy~\cite{cosmos-policy}. VLA methods use VLM backbones to encode visual observations and language instructions, while WAM methods generate latent action representations followed by an action head. All implementations follow the official StarVLA framework~\cite{starvla2026}. Following LIBERO~\cite{libero}, each task is evaluated over 10 trials with slight object-position perturbations, with a maximum rollout horizon of 520 control steps. Initial scene configurations are recorded to ensure consistent starting conditions. Additional experimental details are provided in the appendix.

\begin{figure*}[h]
\vspace{-5mm}
\centering
\includegraphics[width=1.0\textwidth]{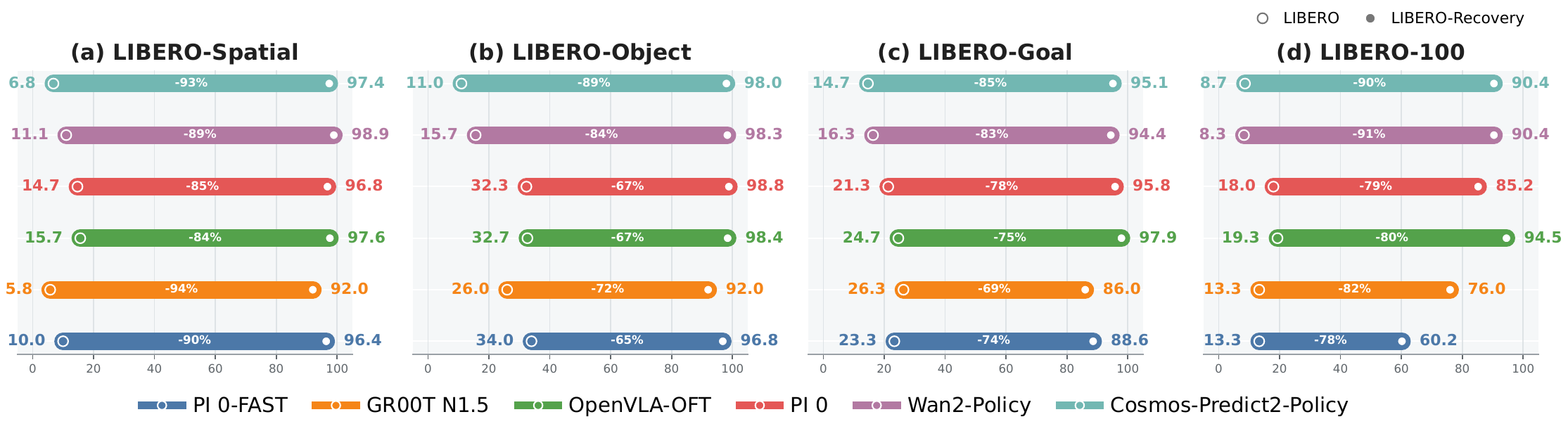}
\caption{The performance of VLAS and WAMS on LIBERO and LIBERO-Recover.}
\label{fig:statu}
\vspace{-4mm}
\end{figure*}

\subsection{MAIN RESULTS}

\subsubsection{Multi-level Evaluation}

\textbf{Key Finding 1: Success in predefined scenarios does not guarantee robust generalization to real failures.} As shown in Fig.~\ref{fig:statu}, performance on predefined scenarios does not reliably reflect embodied models' capabilities under real execution. All models suffer over 50\% performance drops when exposed to naturally occurring failures. More importantly, performance advantages on standard benchmarks do not consistently transfer to failure scenarios. For example, Wan2-Policy outperforms GR00T-N1.5 by \textcolor{blue}{+14.40\%} on LIBERO-100, yet falls \textcolor{red}{-5.0\%} behind it under corresponding failures. Similarly, $\pi_0$ and $\pi_0$-Fast exhibit substantial performance reversals on LIBERO-Spatial. These results demonstrate that standard benchmark rankings are poor predictors of robustness to real execution failures and recovery capability.
% 如图所示，在预定义场景上性能的优劣无法真实的反映具身操作模型在现实场景中的能力，例如，所有的方法在各个场景上均出现了超过50%+能力的下降。另一个事实是，在这种预定义场景下模型之前的性能优劣，反映不到真实的失败场景之中，例如：wan2-policy 与GR00T N1.5 在 LIBERO-100上的+14.40%的优势在对应的失败场景中转化为了-5.0%的劣势。又或者PI 0和PI 0 -fAST在 LIBERO-SPAtial的上的性能波动，这均预示着。。。。

% \begin{figure*}[h]
% \centering
% \includegraphics[width=1.0\textwidth]{images/performance_radar.png}
% \caption{The Multi-level Evaluation on LIBERO-Recover.}
% \label{fig:stat}
% \end{figure*}
% ============================================================
% Required packages in preamble:
%
% ============================================================

\begin{table*}[t]
\vspace{-5mm}
\centering
\caption{The Multi-level Evaluation on LIBERO-Recover.
Darker cells indicate higher recovery success rates. Best results in each
difficulty level are shown in bold. PI (PI-0), FAST (PI-0 FAST), GR00T (GR00T N1.5), Wan.(Wan2-Policy) and Cos.(Cosmos-Predict2-Policy).}
\label{tab:recovery_difficulty}

\scriptsize
\setlength{\tabcolsep}{6.0pt}
\renewcommand{\arraystretch}{0.72}

% ------------------------------------------------------------
% Heatmap colors
% ------------------------------------------------------------
\definecolor{rsr0}{RGB}{255,255,255}
\definecolor{rsr1}{RGB}{242,248,242}
\definecolor{rsr2}{RGB}{220,240,222}
\definecolor{rsr3}{RGB}{190,225,194}
\definecolor{rsr4}{RGB}{150,205,158}
\definecolor{rsr5}{RGB}{105,180,119}
\definecolor{rsr6}{RGB}{65,150,82}

\resizebox{\linewidth}{!}{%
\begin{tabular}{@{}l@{\hspace{4pt}}cccc@{\hspace{8pt}}
                cccc@{\hspace{8pt}}
                cccc@{\hspace{8pt}}
                cccc@{}}
\toprule

\cellcolor{white}
& \multicolumn{4}{c}{\textbf{LIBERO-Spatial}}
&
\multicolumn{4}{c}{\textbf{LIBERO-Object}}
&
\multicolumn{4}{c}{\textbf{LIBERO-Goal}}
&
\multicolumn{4}{c}{\textbf{LIBERO-100}}
\\[-1pt]

\cmidrule{2-5}
\cmidrule{6-9}
\cmidrule{10-13}
\cmidrule{14-17}

\cellcolor{white}\textbf{Method}
& \textbf{L1} & \textbf{L2} & \textbf{L3} & \textbf{L4}
& \textbf{L1} & \textbf{L2} & \textbf{L3} & \textbf{L4}
& \textbf{L1} & \textbf{L2} & \textbf{L3} & \textbf{L4}
& \textbf{L1} & \textbf{L2} & \textbf{L3} & \textbf{L4}
\\
\midrule

\cellcolor{white}PI
& \cellcolor{rsr4}\textbf{40.8}
& \cellcolor{rsr2}11.2
& \cellcolor{rsr2}13.7
& \cellcolor{rsr1}6.0
& \cellcolor{rsr6}\textbf{56.4}
& \cellcolor{rsr4}\textbf{41.9}
& \cellcolor{rsr3}24.6
& \cellcolor{rsr1}8.4
& \cellcolor{rsr5}50.6
& \cellcolor{rsr3}21.9
& \cellcolor{rsr1}5.9
& \cellcolor{rsr1}2.7
& \cellcolor{rsr4}\textbf{39.4}
& \cellcolor{rsr3}\textbf{20.6}
& \cellcolor{rsr2}10.1
& 0.0
\\

\cellcolor{white}FAST
& \cellcolor{rsr3}29.3
& \cellcolor{rsr2}13.1
& \cellcolor{rsr1}6.9
& \cellcolor{rsr1}3.2
& \cellcolor{rsr5}49.2
& \cellcolor{rsr4}41.7
& \cellcolor{rsr4}\textbf{35.2}
& \cellcolor{rsr2}\textbf{12.2}
& \cellcolor{rsr6}\textbf{56.9}
& \cellcolor{rsr3}25.8
& \cellcolor{rsr1}5.6
& 0.0
& \cellcolor{rsr3}24.5
& \cellcolor{rsr2}10.8
& \cellcolor{rsr2}16.2
& 0.0
\\

\cellcolor{white}GR00T
& \cellcolor{rsr1}7.4
& \cellcolor{rsr2}13.2
& \cellcolor{rsr1}7.3
& \cellcolor{rsr1}3.4
& \cellcolor{rsr5}47.9
& \cellcolor{rsr3}26.8
& \cellcolor{rsr3}30.6
& \cellcolor{rsr1}1.8
& \cellcolor{rsr5}51.9
& \cellcolor{rsr4}\textbf{41.8}
& \cellcolor{rsr1}5.8
& 0.0
& \cellcolor{rsr2}14.3
& \cellcolor{rsr3}\textbf{20.6}
& \cellcolor{rsr2}\textbf{17.6}
& 0.0
\\

\cellcolor{white}OFT
& \cellcolor{rsr4}\textbf{40.8}
& \cellcolor{rsr2}12.1
& \cellcolor{rsr2}12.8
& \cellcolor{rsr1}8.6
& \cellcolor{rsr6}\textbf{56.4}
& \cellcolor{rsr4}\textbf{41.9}
& \cellcolor{rsr3}25.1
& \cellcolor{rsr1}7.9
& \cellcolor{rsr6}\textbf{56.9}
& \cellcolor{rsr3}30.8
& \cellcolor{rsr1}\textbf{8.9}
& \cellcolor{rsr1}\textbf{3.9}
& \cellcolor{rsr4}\textbf{39.4}
& \cellcolor{rsr3}\textbf{20.6}
& \cellcolor{rsr2}10.1
& 0.0
\\

\cellcolor{white}Wan.
& \cellcolor{rsr2}16.7
& \cellcolor{rsr1}3.6
& \cellcolor{rsr2}11.8
& \cellcolor{rsr2}\textbf{12.4}
& \cellcolor{rsr1}9.3
& \cellcolor{rsr3}33.1
& \cellcolor{rsr3}20.1
& 0.0
& \cellcolor{rsr3}34.4
& \cellcolor{rsr3}20.8
& \cellcolor{rsr1}4.2
& \cellcolor{rsr1}2.9
& \cellcolor{rsr2}10.7
& \cellcolor{rsr1}7.6
& \cellcolor{rsr2}13.9
& 0.0
\\

\cellcolor{white}Cos.
& \cellcolor{rsr1}7.4
& \cellcolor{rsr1}4.4
& \cellcolor{rsr2}\textbf{14.2}
& 0.0
& \cellcolor{rsr2}10.8
& \cellcolor{rsr3}24.5
& \cellcolor{rsr1}8.5
& 0.0
& \cellcolor{rsr3}30.7
& \cellcolor{rsr2}19.4
& \cellcolor{rsr1}5.7
& 0.0
& \cellcolor{rsr2}11.9
& \cellcolor{rsr2}10.9
& \cellcolor{rsr2}11.2
& 0.0
\\

\bottomrule
\end{tabular}%
}

\vspace{-4mm}

\end{table*}

\textbf{Key Finding 2: Existing manipulation models excel at local correction but struggle with state recovery.} All models achieve substantially higher success rates on $L_{1}$/$L_{2}$ failures than on L3/L4, with performance consistently decreasing as recovery difficulty increases. This progressive degradation suggests that current manipulation models can locally adjust their immediate actions but struggle to reason about and recover the underlying task state after failure. L1 and L2 can often be addressed by adapting the next action based on visual observations, whereas L3 and L4 require structured reasoning about changes in object states, spatial relations, and interaction dependencies, as well as sufficient scene memory. The substantial gap between L2 and L3 therefore reveals a transition from \emph{action-level correction} to \emph{state-level recovery}, which remains a major limitation of existing manipulation models.

\begin{figure*}[h]
\centering
\includegraphics[width=1.0\textwidth]{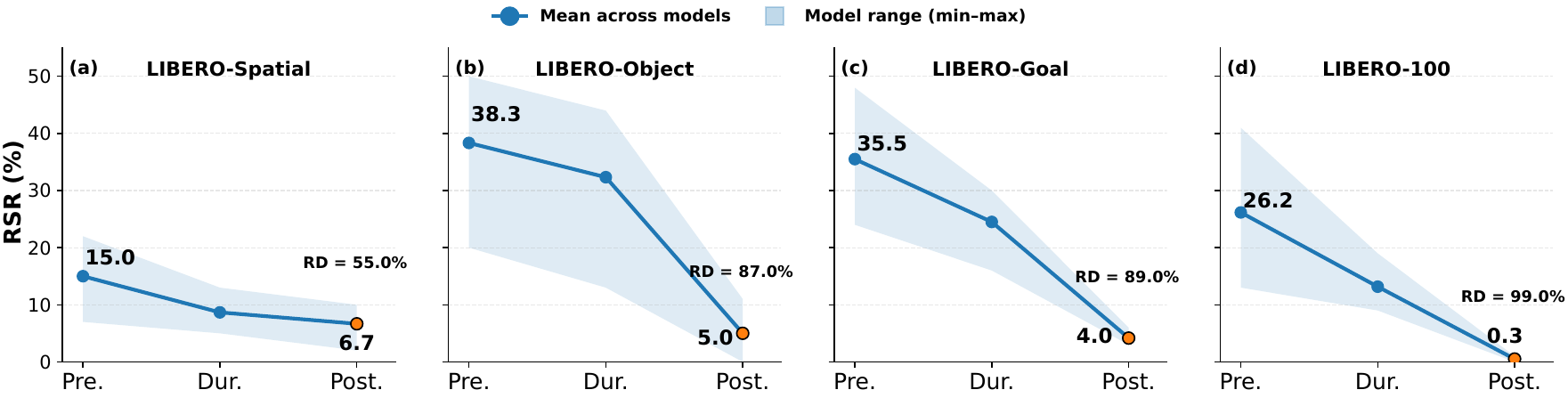}
\caption{Temporal Manifestation of Failures (RD) in VLAs and WAMs on LIBERO-Recover.}
\label{fig:RD}
\end{figure*}

\begin{figure*}[h]
\vspace{-4mm}
\centering
\includegraphics[width=1.0\textwidth]{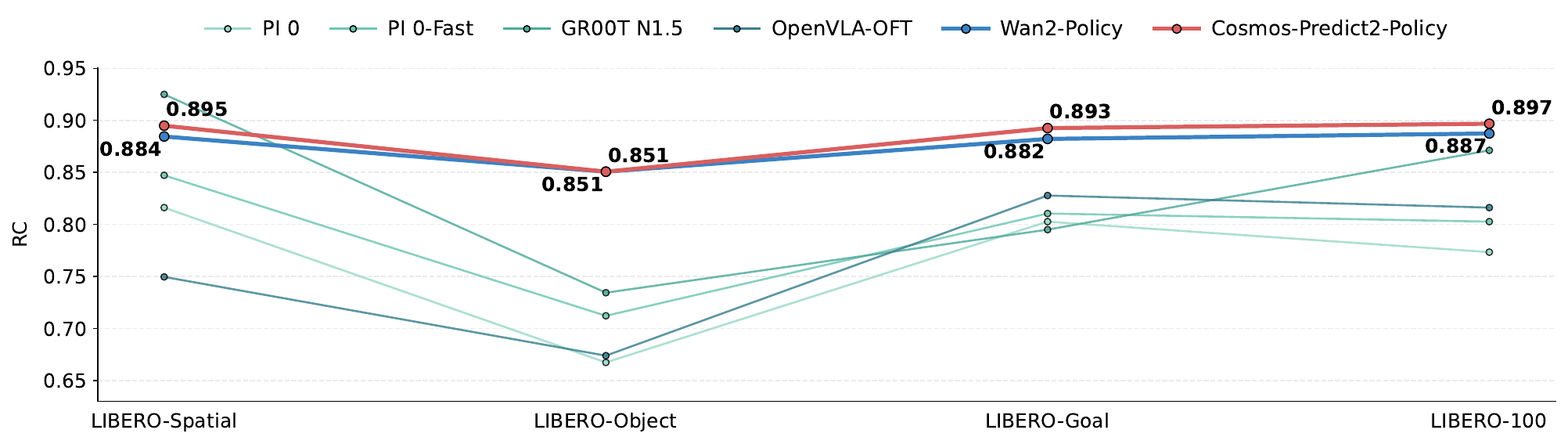}
\caption{The \textbf{RC} performace of VLAS and WAMS on LIBERO-Recover.}
\vspace{-4mm}
\label{fig:RC}
\end{figure*}

\subsubsection{Multi-label Evaluation}

\begin{wraptable}{r}{0.52\textwidth}
\centering
\setlength{\tabcolsep}{2pt}
\caption{Performance on LIBERO and LIBERO-Recover after training with mixed LIBERO-Recover expert data.}
\label{tab:libero_recovery}
\resizebox{\linewidth}{!}{
\begin{tabular}{lcccc}
\toprule
\multicolumn{5}{c}{\textbf{LIBERO}} \\
\midrule
Method & LIBERO-Spatial & LIBERO-Object & LIBERO-Goal & LIBERO-100 \\
\midrule
GR00T N1.5      & 0.920   & 0.920   & 0.860   & 0.760 \\
GR00T N1.5 (Combined) & 0.931 & 0.905 & 0.882 & 0.770 \\
\midrule
OpenVLA-OFT      & 0.976   & 0.984   & 0.979   & 0.945 \\
OpenVLA-OFT (Combined) & 0.982 & 0.994 & 0.970 & 0.916 \\
\midrule
\midrule
\multicolumn{5}{c}{\textbf{LIBERO-Recover}} \\
\midrule
Method      & LIBERO-Spatial & LIBERO-Object & LIBERO-Goal & LIBERO-100 \\
\midrule
GR00T N1.5      & 0.057   & 0.260   & 0.263   & 0.133 \\
GR00T N1.5 (Combined) & 0.116 & 0.262 & 0.264 & 0.206 \\
\midrule
OpenVLA-OFT      & 0.156   & 0.326   & 0.246   & 0.193 \\
OpenVLA-OFT (Combined) & 0.178 & 0.333 & 0.270 & 0.236 \\
\bottomrule
\end{tabular}}
\end{wraptable}

\textbf{Key Finding 3: Failures induce compounding distribution shifts that progressively degrade model performance.} All evaluated manipulation models exhibit substantial performance degradation after failure across all four LIBERO suites. As shown in Fig.~\ref{fig:RD}, averaged across models, the success rate drops from 15.0\% to 6.7\% on LIBERO-Spatial, 35.5\% to 4.0\% on LIBERO-Goal, 38.3\% to 5.0\% on LIBERO-Object, and 26.2\% to only 0.3\% on LIBERO-100, corresponding to recovery degradation rates of 55.0\%, 87.0\%, 89.0\%, and 99.0\%, respectively. Notably, performance further decreases from the failure stage to the post-failure stage, suggesting that repeated unsuccessful interactions may progressively drive the execution into increasingly unfamiliar states. These results indicate that failure is not merely a transient disturbance, but can induce a compounding distribution shift that substantially impairs subsequent task execution.

% \begin{table}[]
% \centering
% \caption{Comparison with other simulation benchmarks.}
% \resizebox{\textwidth}{!}{
% \begin{tabular}{lcccccc}
% \toprule
% Method                 & LIBERO-Spatial & LIBERO-Object & LIBERO-Goal & LIBERO-100 & Overall RC & Mean RC \\
% \midrule
% PI 0                   & 0.8162           & 0.6674          & 0.8027        & 0.7734     & 0.7045       & 0.7649    \\
% PI 0-Fast              & 0.8472         & 0.7122        & 0.8105      & 0.8027     & 0.7505     & 0.7931  \\
% GR00T N1.5             & 0.9250         & 0.7344        & 0.7950      & 0.8712     & 0.7980     & 0.8314  \\
% OpenVLA-OFT            & 0.7496         & 0.6739        & 0.8278      & 0.8162     & 0.7491     & 0.7669  \\
% Wan2-Policy            & 0.8845         & 0.8507        & 0.8822      & 0.8874     & 0.8698     & 0.8762  \\
% Cosmos-Predict2-Policy & 0.8949         & 0.8507        & 0.8926      & 0.8968     & 0.8837     & 0.8875 \\
% \bottomrule
% \end{tabular}}
% \end{table}

\textbf{Key Finding 4: WAM policies exhibit more stable recovery}. WAM policies consistently achieve higher Recovery Consistency (RC) than conventional VLAs, with Cosmos-Predict2-Policy and Wan2-Policy reaching 0.8837 and 0.8698, respectively, higher compared with conventional VLAs as shown in Fig.~\ref{fig:RC}. A possible explanation is the difference in training objectives: conventional VLAs primarily learn action imitation from successful trajectories, whereas world-model-based policies explicitly model action-conditioned state transitions, encouraging the model to reason about how actions affect the current state and future outcomes. Such a transition-aware representation may provide a stronger inductive bias for adapting to unexpected failure states, leading to more consistent recovery behavior.

\begin{figure*}[h]
\centering
\vspace{-5mm}
\includegraphics[width=1.0\textwidth]{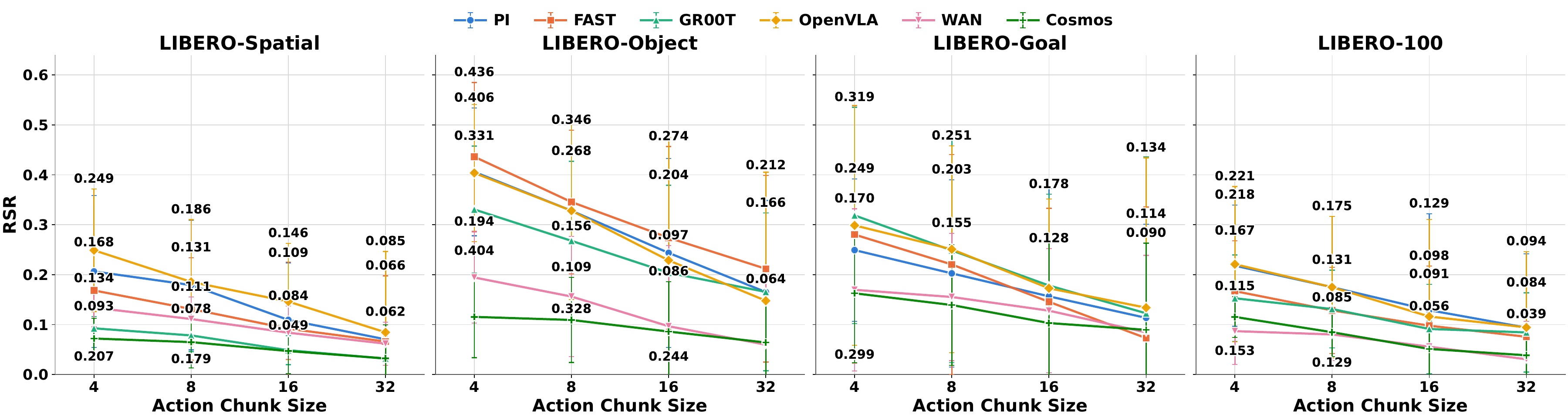}
\caption{The impact of Chunk Size on Model Recovery Capability.}
\label{fig:stat}
\vspace{-4mm}
\end{figure*}

\subsubsection{How to enhance model failure handling?}

\textbf{Key Finding 5: Smaller action chunks improve recovery by enabling more frequent state feedback.} Across all four LIBERO subsets, recovery success consistently degrades as the action chunk size increases from 4 to 32, with the most substantial drop generally occurring at larger chunk sizes. This trend is shared across different policies, indicating that recovery performance is highly sensitive to the temporal granularity of action execution. Longer action chunks commit the robot to a sequence of actions before the next policy update, making it harder to correct deviations caused by failures. In contrast, smaller chunks enable more frequent feedback and action adjustment, allowing the policy to better track the evolving post-failure state. These results suggest that effective failure recovery requires not only a capable policy, but also sufficiently fine-grained closed-loop control.

\textbf{Key Finding 6: Recovery training does not effectively transfer to failures encountered during standard task execution.} Standard LIBERO evaluation is not failure free: even when starting from valid initial states, policies can err during execution and fail to complete the task. This raises a natural question: can training on failure-recovery scenarios teach a policy to recover from failures that arise during standard task execution, thereby improving its overall LIBERO success rate? To examine this, we jointly train the policies on the original LIBERO training data and LIBERO-Recover data, and evaluate them on both standard LIBERO tasks and LIBERO-Recover scenarios.

\begin{wraptable}{r}{0.52\textwidth}
\centering
\setlength{\tabcolsep}{2pt}
\caption{Impact of temporal prompts on model recovery capability.}
\label{tab:libero_recovery}
\resizebox{\linewidth}{!}{
\begin{tabular}{lcccc}
\toprule
\multicolumn{5}{c}{\textbf{LIBERO-Recover}} \\
\midrule
Method      & LIBERO-Spatial & LIBERO-Object & LIBERO-Goal & LIBERO-100 \\
\midrule
GR00T N1.5      & 0.057   & 0.260   & 0.263   & 0.133 \\
GR00T N1.5 (temporal) & 0.066 & 0.260 & 0.264 & 0.166 \\
\midrule
OpenVLA-OFT      & 0.156   & 0.326   & 0.246   & 0.193 \\
OpenVLA-OFT (temporal) & 0.193 & 0.333 & 0.300 & 0.247 \\
\bottomrule
\end{tabular}}
\end{wraptable}

Interestingly, the results do not support this expectation. Adding LIBERO-Recover data consistently improves performance on LIBERO-Recover, but produces little or even negative change on standard LIBERO. For GR00T N1.5, mixed training improves the average recovery success rate from 17.8\% to 21.2\%, while the average LIBERO success rate only increases from 86.5\% to 87.2\%. For OpenVLA-OFT, recovery success increases from 20.8\% to 25.4\%, whereas the average LIBERO success rate slightly decreases from 97.1\% to 96.6\%. Thus, the recovery data clearly provides useful supervision for the recovery scenarios on which it is evaluated, yet this learned recovery capability provides little benefit when failures arise naturally during standard LIBERO execution.

This result reveals a failure-recovery transfer gap. Exposing a policy to failure states is not sufficient to make it reliably recover from its own execution errors during ordinary task execution. \textbf{\emph{This suggests that effective recovery may depend not only on learning how to act from a failed state, but also on recognizing and responding to failures that emerge online during the policy's own execution trajectory}}.

\textbf{Key Finding 7: Temporal context improves failure recovery}. Providing the task initial frame as temporal context consistently improves recovery performance across both models, with the largest gains on more challenging tasks. For GR00T N1.5, temporal context improves the overall recovery success rate from 17.8\% to 18.9\%. The effect is substantially stronger for OpenVLA-OFT, whose average recovery success increases from 20.8\% to 26.8\%, including a notable improvement on LIBERO-100 from 19.3\% to 24.7\% and on LIBERO-Goal from 24.6\% to 30.0\%. These improvements suggest that recovery benefits from access to the task's temporal context rather than relying solely on the current failure state. The initial frame provides a reference for comparing the current scene against the intended task configuration, allowing the policy to identify state deviations and infer what has been disrupted by the failure. This indicates that effective failure recovery requires not only understanding the current state, but also reasoning about how the current state differs from the expected state.

\begin{figure*}[h]
\vspace{-5mm}
\centering
\includegraphics[width=1.0\textwidth]{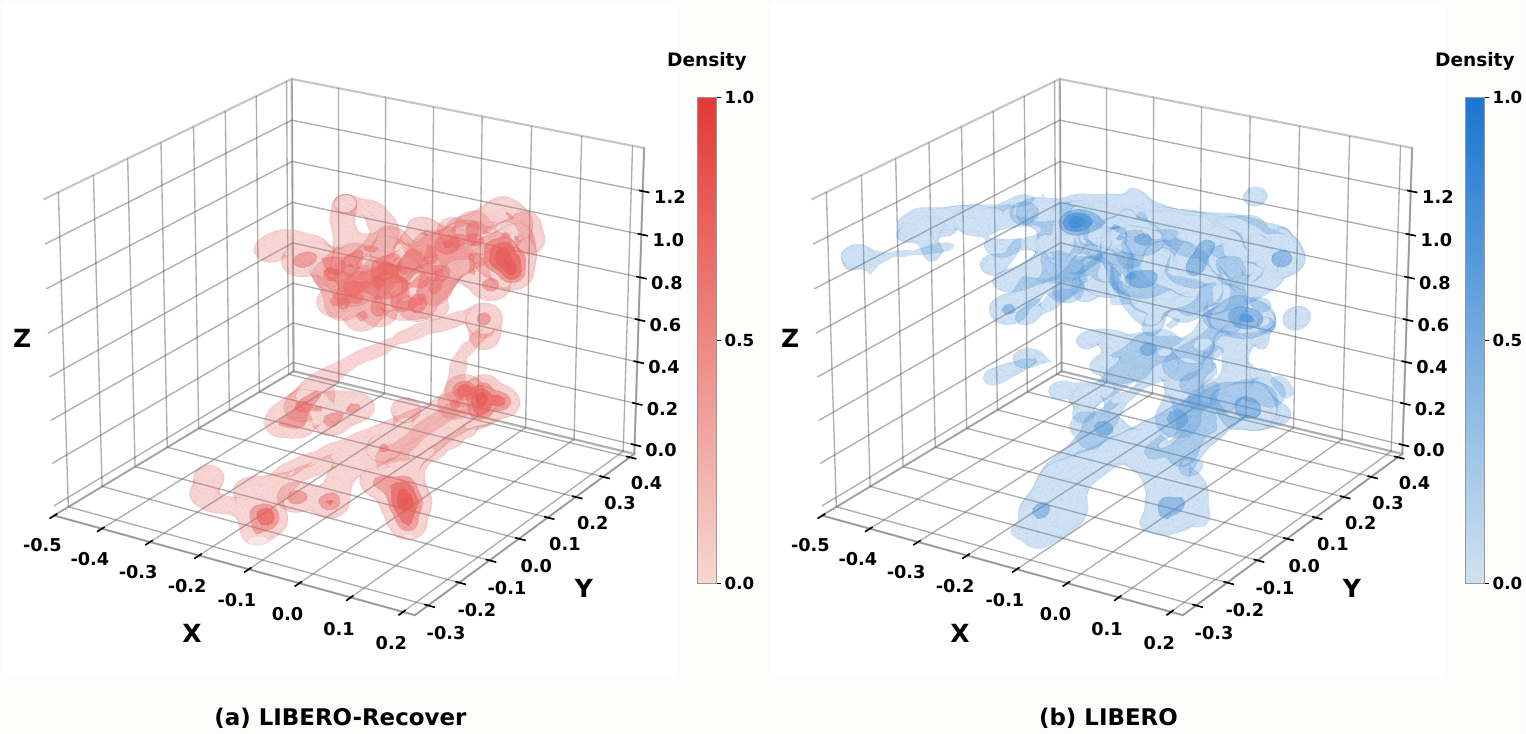}
\caption{Distribution visualization of training trajectories on LIBERO-Recover and LIBERO.}
\vspace{-4mm}
\label{fig:stat}
\end{figure*}

\textbf{Key Finding 8: Failure recovery occurs in a more concentrated state distribution.} The KDE visualization reveals that LIBERO-Recover exhibits a more concentrated state distribution than the original LIBERO data, with several distinct high-density regions, while the original data covers a more dispersed workspace. This suggests that failure recovery is concentrated around a limited number of critical states where failures are more likely to occur, such as grasping, placing, and object interaction. In contrast, standard demonstrations cover a broader range of successful execution states. This distributional difference indicates that recovery data provides concentrated information around critical state-transition regions that are underrepresented in standard demonstrations, highlighting the need to explicitly model and evaluate failure recovery.

\section{Conclusion}

% We introduced \textbf{LIBERO-Recover}, a benchmark for evaluating failure recovery in robotic manipulation models. By collecting naturally occurring failures from embodied model executions, LIBERO-Recover evaluates whether models can understand disrupted states, adapt their actions, and recover to complete the original task. Experiments across representative VLA and WAM models reveal a substantial gap between standard task success and failure recovery, with performance degrading sharply as recovery requires increasingly deeper state reasoning. Our analysis further shows that finer-grained action execution and temporal context can improve recovery, while current recovery training provides limited transfer to failures encountered during standard task execution. We hope LIBERO-Recover will encourage future research toward more robust and failure-aware embodied agents.

We introduced \textbf{LIBERO-Recover}, a benchmark for failure recovery in robotic manipulation. Using naturally occurring failures from embodied model executions, LIBERO-Recover evaluates whether models can understand disrupted states, adapt their actions, and recover to complete the original task. Experiments across VLA and WAM models reveal a substantial gap between standard task success and failure recovery, with performance degrading sharply as recovery requires deeper state reasoning. Our analysis shows that finer-grained action execution and temporal context can improve recovery, while recovery training provides limited transfer to failures during standard task execution. We hope LIBERO-Recover will encourage research toward robust and failure-aware embodied agents.

\subsection*{AI Use Statement}
During the preparation of this work, the authors used generative AI tools to assist with method code implementation, language polishing, and figure layout and color adjustments. Specifically, the open-source Qwen 3.5 model was applied to suggest the partitioning of the $V_{\text{pre}}$, $V_{\text{fail}}$, and $V_{\text{post}}$ sets, and all model-generated annotations were manually sampled and cross-verified by the authors to guarantee data quality. Generative AI was not used to develop theoretical models or conceptual frameworks, or to formulate or refine hypotheses. It was not used to provide essential elements for mathematical proofs, assist in writing proofs, or support qualitative or thematic data analysis.

All AI-assisted outputs were thoroughly reviewed and verified by the authors. Any code or annotations generated with the assistance of large language models were independently tested and verified for correctness. The authors take full responsibility for the final content of this work, including all text, claims, data annotations, and materials produced with the assistance of generative AI. Please refer to the Appendix for further details.

\subsection*{Reproducibility Statement}
Section~\ref{main} and Section~\ref{sec:dataset_con} details the methodology for constructing LIBERO-Recover and the data collection procedure. Subsequent experimental section~\ref{exp} and the Appendices~\ref{app:dataset_construction} and~\ref{app:exp_details} describe the training datasets, benchmark splits, evaluation metrics, and baseline configurations. The Appendix further provides full model configurations, optimization hyperparameters, evaluator definitions, and metric rules. Upon acceptance, we will publicly release all source code, datasets, evaluation protocols, and model checkpoints required to reproduce our findings.

\bibliography{iclr2027_conference}
\bibliographystyle{iclr2027_conference}

\appendix

\clearpage 
\appendix 
\startcontents[appendix] 
\section*{Appendix Contents} 
\printcontents[appendix]{}{1}{} 
\clearpage

\section{Related Work}

\subsection{Robot Manipulation Policies}

\begin{figure*}[h]
\centering
\includegraphics[width=1.0\textwidth]{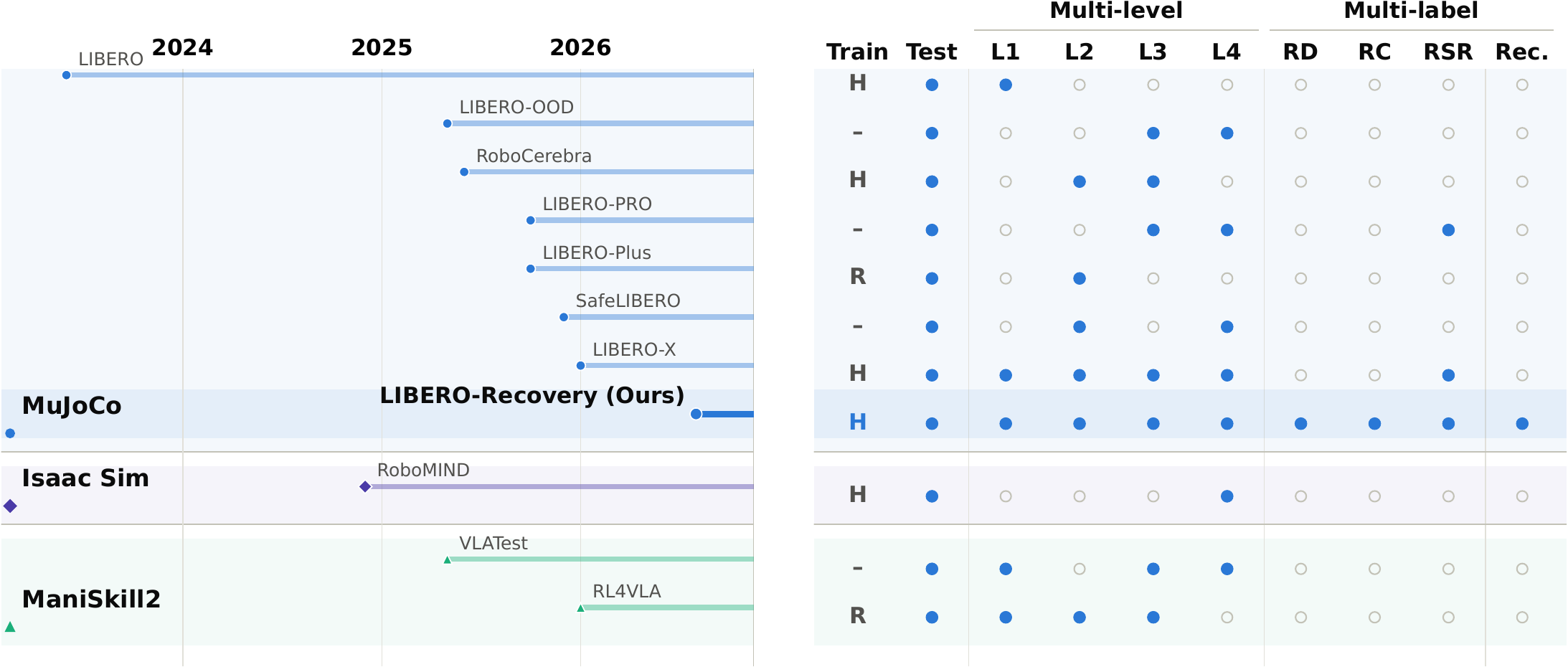}
\caption{\textbf{Comparison of LIBERO-Recover with existing benchmarks.} LIBERO-Recover is the first benchmark dedicated to failure recovery, where \textbf{Rec.}}
\label{fig:benchmark}
\end{figure*}

Robot manipulation policies have evolved from visual imitation learning toward general-purpose models that jointly understand vision, language, and actions. Representative VLA methods, including RT-1, RT-2, OpenVLA~\cite{openvla}, $\pi_0$~\cite{pi_0}, $\pi_{0.5}$~\cite{pi_05}, RoboMamba~\cite{robomamba}, and GR00T~\cite{gr00t}, leverage large-scale vision-language and robot datasets to improve generalization across tasks, objects, and embodiments.

More recently, world models have been incorporated into robot manipulation, leading to WAMs such as UniVLA~\cite{univla}, Video Policy~\cite{video_policy}, UWM~\cite{uwm}, FLARE~\cite{flare}, and Cosmos-Policy~\cite{cosmos}. By modeling actions and future states, these methods provide foresight into action consequences and improve long-horizon planning. However, existing VLA/WAM methods primarily focus on successful action prediction and execution, while systematic modeling of failures, unexpected state transitions, and post-failure recovery remains largely unexplored. Enabling robots to understand, adapt to, and recover from failures is therefore crucial for reliable manipulation.

\subsection{Robot Manipulation Benchmarks} 
Standardized benchmarks are essential for evaluating the generalization and robustness of robot manipulation policies. Early benchmarks such as RLBench~\cite{rlbench} and LIBERO~\cite{libero} established large-scale simulation tasks and unified evaluation protocols. Recent benchmarks, including CALVIN~\cite{calvin}, SimplerEnv~\cite{simplenv}, RoboCasa~\cite{robocasa}, RoboTwin~\cite{robotwin}, and LIBERO-Plus~\cite{libero-plus}, LIBERO-Pro~\cite{libero-pro}, LIBERO-X~\cite{libero-x}, and LIBERO-Safety~\cite{liberosafety}, further evaluate task generalization, scene diversity, long-horizon interaction, and robustness to controlled changes in object placement, appearance, robot pose, and task instructions. Despite these advances, existing benchmarks primarily rely on \emph{manually predefined distribution shifts} and evaluate whether policies can \emph{complete tasks under changed conditions, as shown in Fig.~\ref{fig:benchmark}}.

In contrast, real execution failures can induce complex and unstructured states, such as object displacement and unintended state changes, which remain largely unexplored. \textbf{LIBERO-Recover} addresses this gap by collecting naturally occurring failures and resulting state transitions from embodied model executions, constructing progressively challenging recovery scenarios to evaluate state understanding, action adaptation, interaction reasoning, and failure recovery.

% 数据集构建与标注
\section{Dataset Construction and Annotation}
\label{app:dataset_construction}

% 本附录介绍 LIBERO-Recover 的具体数据构建流程，包括失败阶段划分、失败时间定位、恢复难度与类别标注、遥操作采集以及轨迹录制和后处理。正文中已经给出了失败状态和四级恢复难度的形式化定义，本附录仅补充这些定义在数据构建过程中的具体实施方式。

% 失败阶段的操作性定义
\subsection{Operational Definition of Failure Stages}
\label{app:failure_stages}

% 出于时序分析的目的，每个执行视频被划分为三个阶段：失败前（Pre-Failure）、失败中（During-Failure）和失败后（Post-Failure）。
For temporal analysis, each execution video is divided into three phases: Pre-Failure, During-Failure, and Post-Failure.

% 失败前阶段。
\paragraph{Pre-Failure.}
% 失败前阶段包含第一次出现与任务相关偏差之前的观测。在此阶段，机器人遵循的执行过程仍与任务指令保持一致，且当前规划仍可执行。
The Pre-Failure phase contains the observations before the first task-relevant deviation. During this phase, the robot follows an execution that is still consistent with the task instruction, and the current plan remains executable.

% 失败中阶段。
\paragraph{During-Failure.}
% 失败中阶段始于机器人动作引起任务相关偏差的最早可观测时刻。它包括导致失败的因果交互，并在由此产生的状态在视觉上达到稳定或机器人停止尝试原动作时结束。
The During-Failure phase starts at the earliest observable moment at which the robot action causes a task-relevant deviation. It includes the causal interaction that produces the failure and ends when the resulting state becomes visually stable or the robot stops attempting the original action.

% 失败后阶段。
\paragraph{Post-Failure.}
% 失败后阶段始于失败状态稳定之后。它包含在失败之后以及恢复、终止或人工干预之前所能获取的观测。这些观测刻画了恢复策略在生成纠正动作之前必须理解的状态特征。
The Post-Failure phase begins after the failure state has become stable. It contains the observations available after the failure and before recovery, termination, or manual intervention. These observations characterize the state that a recovery policy must interpret before generating corrective actions.

% 失败时间戳对应于失败中阶段的起始点。所有时间戳均从录制视频的起始处开始计算，并保留一位小数。
The failure timestamp corresponds to the beginning of the During-Failure phase. All timestamps are measured from the beginning of the recorded video and rounded to one decimal place.

% 时序失败定位提示词
\subsection{Prompt for Temporal Failure Localization}
\label{app:failure_localization_prompt}

% 我们使用视觉语言模型从录制的机器人视频中定位执行失败的起始点。以下提示词被用作默认模板。论文中为清晰起见将原始提示词翻译为了英文；代码实现中使用了对应的中文版本。
We use a vision-language model to localize the onset of execution failures from recorded robot videos. The following prompt was used as the default template. The original prompt was translated into English for clarity in the paper; the implementation used the corresponding Chinese version.

\begin{promptbox}[Temporal Failure Localization Prompt]

\setlength{\parindent}{1.5em}
\setlength{\parskip}{0.4em}

You are an expert in analyzing robot manipulation videos.

Watch the entire video carefully and identify the earliest moment at which the robot experiences a task-relevant execution failure.

A failure is defined as the earliest observable moment when:

\hspace{1.5em}(1) the robot fails to grasp or place the manipulated object as intended;

\hspace{1.5em}(2) the robot action causes an object displacement or interaction that invalidates the current execution plan.

Return the earliest failure timestamp for the manipulated object.

\textbf{Requirements:}

\hspace{1.5em}1. Analyze the complete video before making a decision.

\hspace{1.5em}2. Report time in seconds from the beginning of the video.

\hspace{1.5em}3. Round the timestamp to one decimal place.

\hspace{1.5em}4. Do not provide explanations.

\hspace{1.5em}5. Output JSON only.

The output format must be exactly:

\hspace{2em}\{
    
\hspace{3em}"fail": 2.0

\hspace{2em}\}

\end{promptbox}

% 模型输出被解析为 JSON 对象。如果生成的回复包含额外文本，首先提取最终推理标记之后的内容；然后使用正则表达式回退机制来恢复第一个有效的 JSON 对象。无法解析为所需模式的样本将被丢弃并单独审查。
The model output is parsed as a JSON object. If the generated response contains additional text, the content after the final reasoning marker is first extracted; a regular-expression fallback is then used to recover the first valid JSON object. Samples that cannot be parsed into the required schema are discarded and reviewed separately.

% 定性示例
\paragraph{Qualitative Examples.}

\begin{figure}[t]
    \centering
    \includegraphics[width=\linewidth]{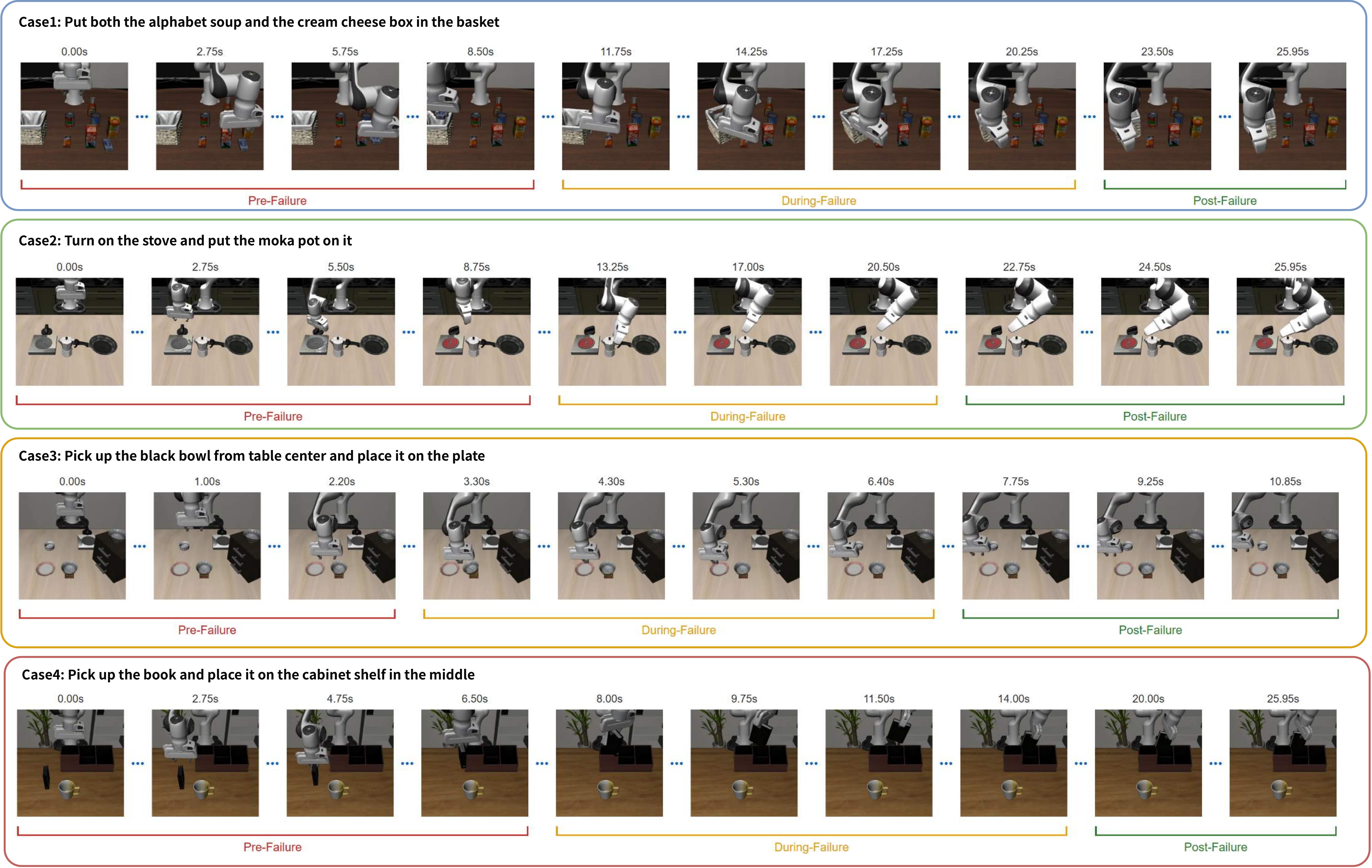}
    % 时序失败定位的定性示例。每行展示一次操作推演中的十个代表性关键帧，附带来自原始视频的时间戳。关键帧下方的括号标示了失败前、失败中和失败后，展示了从任务执行到失败发展及其最终状态的演变过程。关键帧采用非均匀采样，以突出相关的状态转变。
    \caption{Qualitative examples of temporal failure localization. Each row shows ten representative frames from a manipulation rollout, with timestamps from the original video. The brackets below the frames indicate Pre-Failure, During-Failure, and Post-Failure, illustrating the progression from task execution to failure development and its resulting state. Frames are sampled non-uniformly to highlight relevant transitions.}
    \label{fig:temporal_failure_localization}
\end{figure}

% 图~\ref{fig:temporal_failure_localization} 展示了四个操作任务中的时序失败定位：将两个物体放入篮子中、将摩卡壶放置在已开启的炉灶上、将碗转移到盘子上，以及将书本放置在橱柜隔板上。每行按时间顺序展示了十个代表性关键帧，并附有标明其在原始视频中位置的时间戳。阶段标注区分了失败发生前的执行过程、偏差形成的交互过程，以及最终形成的失败后构型。
Figure~\ref{fig:temporal_failure_localization} illustrates temporal failure localization across four manipulation tasks: placing two objects in a basket, placing a moka pot on an activated stove, transferring a bowl onto a plate, and placing a book on a cabinet shelf. Each row presents ten representative frames in chronological order, with timestamps indicating their positions in the original video. The phase annotations distinguish the execution preceding the failure, the interaction during which the deviation develops, and the resulting post-failure configuration.

\paragraph{Temporal Context Evidence.}
% 这些示例突显了理解执行失败所需的时序依据。夹爪与物体关系、物体位置以及任务进展的变化，有助于将进行中的操作尝试与可观测的失败区分开来。定位提示词返回失败起始点，对应于从失败前到失败中的转变；三阶段标注为可视化完整的失败过程提供了额外上下文。
These examples highlight the temporal evidence needed to interpret an execution failure. Changes in the gripper--object relationship, object position, and task progress help distinguish an ongoing manipulation attempt from an observable failure. The localization prompt returns the failure onset, corresponding to the transition from Pre-Failure to During-Failure; the three-phase annotations provide additional context for visualizing the complete failure process.

% 恢复难度与失败分类提示词
\subsection{Prompt for Recovery Difficulty and Failure Categorization}
\label{app:failure_categorization_prompt}

% 定位失败时间戳后，我们进一步使用三个互补字段对每个失败进行标注：恢复难度等级、主要失败类别以及恢复执行原始任务所需的最小纠正动作。恢复动作表示为简洁的可执行指令，而非因果解释。该额外字段使得标注更具可操作性，并支持遥操作员随后的工校验。
After localizing the failure timestamp, we further annotate each failure with three complementary fields: the recovery difficulty level, the primary failure category, and the minimum corrective action required to resume the original task. The recovery action is expressed as a concise executable instruction rather than a causal explanation. This additional field makes the annotation more operational and supports subsequent human verification by teleoperators.

% 以下提示词被用作默认模板。
The following prompt is used as the default template.

\begin{promptbox}[title={Recovery Difficulty, Failure Categorization, and Recovery Action Prompt}]

\setlength{\parindent}{1.5em}
\setlength{\parskip}{0.4em}

% 你是机器人操作与失败恢复领域的专家。
You are an expert in robot manipulation and failure recovery.

% 仔细观看完整视频，并根据恢复难度和主要失败类别对观察到的失败进行分类。然后描述机器人应执行的最小纠正动作，以从失败中恢复并重新执行原始任务。
Watch the entire video carefully and classify the observed failure according to its recovery difficulty and primary failure category. Then describe the minimum corrective action that the robot should execute to recover from the failure and resume the original task.

% 使用以下恢复等级：
Use the following recovery levels:

% L1 -- 动作重试：在物体或环境没有发生实质性改变的情况下，可通过基本重复相同的动作来纠正失败。
\hspace{1.5em}\textbf{L1 -- Action Retry:} the failure can be corrected by repeating essentially the same action, without any meaningful change to the object or environment.

% L2 -- 动作调整：机器人必须修改动作的执行方式，例如改变抓取位置、接近方向、时机或运动轨迹。
\hspace{1.5em}\textbf{L2 -- Action Adaptation:} the robot must modify the execution of the action, such as changing the grasp position, approach direction, timing, or motion trajectory.

% L3 -- 物体状态恢复：失败改变了物体的状态，机器人在继续执行原始任务之前必须先恢复或重新整理该物体。
\hspace{1.5em}\textbf{L3 -- Object State Recovery:} the failure changes the state of an object, and the robot must first restore or rearrange the object before continuing the original task.

% L4 -- 环境恢复：失败改变了周围环境或造成了阻碍，在任务继续进行之前必须清除阻碍或恢复环境。
\hspace{1.5em}\textbf{L4 -- Environmental Recovery:} the failure changes the surrounding environment or creates an obstruction that must be removed or restored before the task can continue.

% 从以下列表中恰好选择一个失败类别：
Choose exactly one failure category from the following list:

% 抓取失败
\hspace{1.5em}\texttt{grasp\_failure}

% 放置失败
\hspace{1.5em}\texttt{placement\_failure}

% 物体移位
\hspace{1.5em}\texttt{object\_displacement}

% 碰撞或交互失败
\hspace{1.5em}\texttt{collision\_or\_interaction\_failure}

% 环境阻碍
\hspace{1.5em}\texttt{environmental\_obstruction}

% 其他
\hspace{1.5em}\texttt{other}

% 要求：
\textbf{Requirements:}

% 1. 使用原始任务指令作为任务目标。
\hspace{1.5em}1. Use the original task instruction as the task objective.

% 2. 考虑成功恢复所需的最小纠正行为。
\hspace{1.5em}2. Consider the minimum corrective behavior required for successful recovery.

% 3. 恰好分配一个恢复等级和一个失败类别。
\hspace{1.5em}3. Assign exactly one recovery level and one failure category.

% 4. 将恢复动作描述为简洁的祈使指令。
\hspace{1.5em}4. Describe the recovery action as a concise imperative instruction.

% 5. 恢复动作应在必要时指定相关物体、目标、夹爪状态、空间调整以及执行顺序。
\hspace{1.5em}5. The recovery action should specify the relevant object, target, gripper state, spatial adjustment, and execution order when necessary.

% 6. 不要提供解释、原因、置信度得分或替代恢复方案。
\hspace{1.5em}6. Do not provide explanations, causes, confidence scores, or alternative recovery plans.

% 7. 仅输出有效 JSON。不要使用 Markdown 代码块标记。
\hspace{1.5em}7. Output valid JSON only. Do not use Markdown code fences.

% 输出格式必须严格为：
The output format must be exactly:

\hspace{2em}\{
\hspace{3em}"recovery\_level": "L2",
\hspace{3em}"failure\_category": "grasp\_failure",
\hspace{3em}"recovery\_action": "Reposition the gripper above the bowl, close the gripper to grasp it, and resume the original task."
\hspace{2em}\}

\end{promptbox}

% 生成的 JSON 标注既包含失败的语义描述，也包含可执行的恢复规范。例如，动作重试可以表示为重复失败的抓取，而物体状态失败可能需要先重新放置移位的物体，然后再继续执行原始指令。
The resulting JSON annotation contains both a semantic description of the failure and an executable recovery specification. For example, an action retry may be represented by repeating the failed grasp, whereas an object-state failure may require first repositioning the displaced object and then resuming the original instruction.

\begin{figure}[t]
    \centering
    \includegraphics[width=\linewidth]{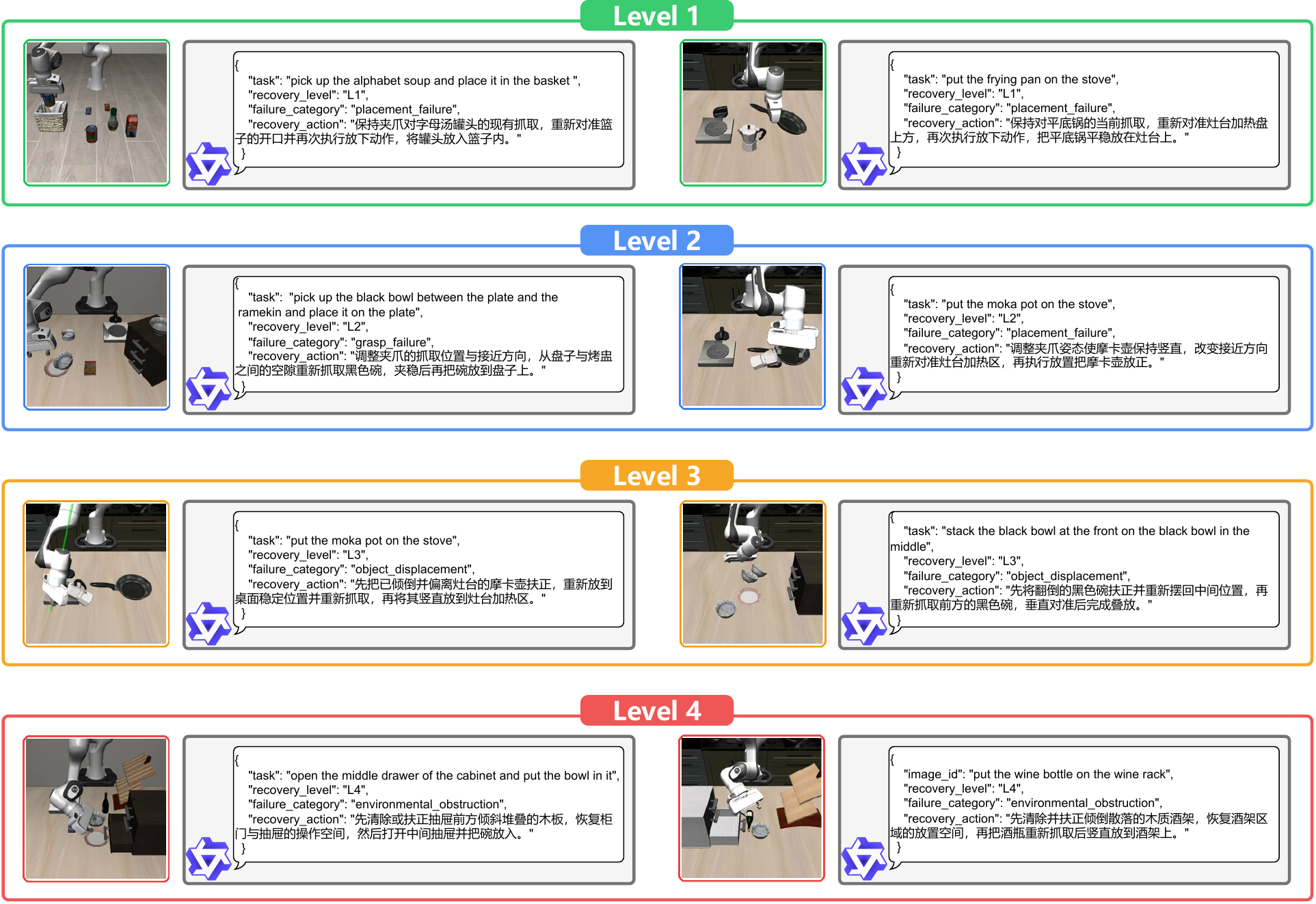}
    % 模型生成的恢复标注示例。每个示例展示了观察到的失败、预测的恢复等级与失败类别，以及 JSON 输出中生成的相应纠正动作。
    \caption{Examples of model-generated recovery annotations. Each example shows the observed failure, the predicted recovery level and failure category, and the corresponding corrective action generated in the JSON output.}
    \label{fig:failure_categorization_examples}
\end{figure}

% 生成的 JSON 标注随后由四名具有机器人操作经验的遥操作员进行复核。每位评审员都会获得完整视频、原始任务指令以及模型生成的 JSON 输出。评审员独立验证三个方面：恢复等级是否与最小纠正行为相匹配、失败类别是否描述了主要失败机制，以及所提出的恢复动作在物理上是否可执行且足以恢复执行原始任务。评审员可以接受标注、修改一个或多个字段，或将样本标记为存在歧义。评审员之间存在分歧的样本将提交裁决，最终标注记录经过复核的恢复等级、失败类别和恢复动作。
The generated JSON annotations are subsequently reviewed by four teleoperators with experience in robot manipulation. Each reviewer is provided with the complete video, the original task instruction, and the model-generated JSON output. The reviewers verify three aspects independently: whether the recovery level matches the minimum corrective behavior, whether the failure category describes the primary failure mechanism, and whether the proposed recovery action is physically executable and sufficient to resume the original task. Reviewers may accept the annotation, revise one or more fields, or flag the sample as ambiguous. Samples with reviewer disagreement are forwarded for adjudication, and the final annotation records the reviewed recovery level, failure category, and recovery action.

% 遥操作数据采集
\subsection{Teleoperation Data Collection}
\label{app:teleoperation_collection}

% 对于每个选定的失败场景，仿真器会恢复到记录的失败状态，同时保留原始任务指令和场景配置。随后由人类遥操作员使用 SpaceMouse 从该状态尝试恢复任务。当存在多种有效的恢复行为时，每个场景可能会记录多条轨迹。
For each selected failure scenario, the simulator is restored to the recorded failure state while preserving the original task instruction and scene configuration. A human teleoperator is then asked to recover the task from this state using a SpaceMouse.For each scenario, multiple trajectories may be recorded when different valid recovery behaviors are possible. 

% 在每次推演之前，机器人位姿、夹爪状态、物体状态以及环境状态都会重置为该场景特有的配置。遥操作员不会获得原始的成功动作序列，必须根据当前观测和任务指令来决定恢复行为。如果在失败状态之后满足原始任务目标，则认为该恢复轨迹是成功的。
Before each rollout, the robot pose, gripper state, object states, and environment state are reset to the scenario-specific configuration. The teleoperator is not provided with the original successful action sequence and must determine the recovery behavior from the current observation and the task instruction. A recovery trajectory is considered successful if the original task goal is satisfied after the failure state.

% 轨迹录制与后处理
\subsection{Trajectory Processing}
\label{app:trajectory_processing}

% 遥操作演示数据以基于回合的 LeRobot v2.1 格式组织。每个恢复回合包含遥操作期间生成的同步第三人称与腕部相机观测、机器人状态以及末端执行器动作。除机器人观测和动作外，时间戳、帧索引、回合索引和任务索引等标准时序及索引字段用于保持每条轨迹内的时间顺序与任务对应关系。
The teleoperated demonstrations are organized in the episode-based LeRobot v2.1 format~\cite{cadene2026lerobot}. Each recovery episode contains synchronized third-person and wrist-camera observations, robot states, and end-effector actions generated during teleoperation. In addition to the robot observations and actions, standard temporal and indexing fields, such as timestamps, frame indices, episode indices, and task indices, are used to preserve temporal ordering and task correspondence within each trajectory.

% 在策略训练期间，LeRobot 数据流水线通过联合索引视觉观测、机器人状态、任务信息与动作来重构时间对齐的样本，使这些演示数据可直接兼容基于 LeRobot 的 VLA 训练流水线（例如 StarVLA）。重要的是，专家恢复轨迹作为训练和适配数据发布，而非作为基准测试输入的一部分。标准的 LIBERO-Recover 评估仅需要恢复场景资产，因此不会向受评测策略提供专家恢复动作序列。
During policy training, the LeRobot data pipeline reconstructs temporally aligned samples by jointly indexing visual observations, robot states, task information, and actions, making the demonstrations directly compatible with LeRobot-based VLA training pipelines such as StarVLA~\cite{starvla2026}. Importantly, the expert recovery trajectories are released as training and adaptation data rather than as part of the benchmark input. Standard LIBERO-Recover evaluation requires only the recovery-scene assets, and therefore does not provide the expert recovery action sequence to the evaluated policy.

% 数据集与发布格式
\section{Dataset and Release Format}
\label{sec:dataset_release_format}

% LIBERO-Recover 作为一组面向恢复的场景资产和可选训练数据集发布。该发布版本保留了 LIBERO 的任务语义和场景描述，同时额外记录了轨迹执行过程中选定时间戳处的仿真器状态。这种设计将任务定义与用于初始化恢复回合的物理状态分离开来。
LIBERO-Recover is released as a collection of recovery-oriented scene assets and optional training datasets. The release preserves the task semantics and scene descriptions of LIBERO, while additionally recording the simulator state at selected timestamps during trajectory execution. This design separates the task definition from the physical state used to initialize a recovery episode.

% LIBERO-Recover 由三个互补的数据组件组成：评估恢复场景、专家恢复轨迹和失败前时序历史。它们支持恢复流程的不同阶段，包括基于场景的评估、恢复策略训练和时序上下文建模。
LIBERO-Recover is organized into three complementary data components: \emph{Evaluation Recovery Scenarios}, \emph{Expert Recovery Trajectories}, and \emph{Pre-failure Temporal Histories}. They support different stages of the recovery pipeline, including scene-based evaluation, recovery policy training, and temporal-context modeling.

% LIBERO-Recover 覆盖四个源任务套件中的 130 个任务，对应五个 BDDL 目录；LIBERO-100 由 LIBERO-90 和 LIBERO-10 共同组成。
\paragraph{Source Task Suites.}
Table~\ref{tab:suites} summarizes the source task suites and their
corresponding BDDL directories. LIBERO-Recover covers 130 tasks across
four suites: LIBERO-Spatial, LIBERO-Object, LIBERO-Goal, and LIBERO-100.
The latter comprises LIBERO-90 and LIBERO-10, resulting in five
directories in the released task assets.

\begin{table}[t]
  \centering
  \caption{
    Source task suites and BDDL directories used by LIBERO-Recover,
    covering 130 tasks in total.
    LIBERO-100 comprises \texttt{libero\_90} (90 tasks) and
    \texttt{libero\_10} (10 tasks).
    Original task instructions and scene definitions are preserved,
    while recovery episodes are initialized from recorded failure states.
}
  \label{tab:suites}
  \begin{tabular}{@{}llcp{0.42\textwidth}@{}}
    \toprule
    Suite & bddl folder & Subtasks & Scope \\
    \midrule
    LIBERO-Spatial & \texttt{libero\_spatial} & 10 &
      One bowl, ten spatial relations (between / next to / on / inside) \\
    LIBERO-Object & \texttt{libero\_object} & 10 &
      Ten distinct objects, each placed into the basket \\
    LIBERO-Goal & \texttt{libero\_goal} & 10 &
      One scene, ten distinct goal states (open / close / on / turn on) \\
    LIBERO-100 & \texttt{libero\_90} + \texttt{libero\_10} & 100 &
      90 single-family long-horizon tasks (cabinets, stoves, baskets, trays,
      caddies) and 10 compositional multi-step tasks \\
    \midrule
    \textbf{Total} & & \textbf{130} & \\
    \bottomrule
  \end{tabular}
\end{table}

% 评估恢复场景
\subsection{Evaluation Recovery Scenarios}
\label{sec:evaluation_recovery_scenarios}

% 评估恢复场景为基准测试评估提供了核心资产。每个场景都与一个 LIBERO 任务相关联并包含 BDDL 任务描述。这些场景旨在用于标准化评估。它们允许不同的策略从相同的任务和记录的场景状态开始，同时通过初始化过程中的受控变体支持多次恢复试验。
The \emph{Evaluation Recovery Scenarios} provide the core assets for benchmark evaluation. Each scenario is associated with a LIBERO task and contains a BDDL task description. These scenarios are intended for standardized evaluation. They allow different policies to start from the same task and recorded scene state, while enabling multiple recovery trials through controlled variations in the initialization procedure.

% 每个恢复场景由一对文件表示：BDDL 任务描述和对应的仿真器状态文件：
Each recovery scene is represented by a pair of files: a BDDL task description and a corresponding simulator-state file:
\[
\mathcal{S}_i =
\left(
\mathcal{B}_i,\,
\mathcal{X}_i
\right),
\]
% 其中 $\mathcal{B}_i$ 是 BDDL 文件，$\mathcal{X}_i$ 是存储扁平化 MuJoCo 仿真状态的 NumPy 文件。这两个文件共享相同的文件名前缀。
where $\mathcal{B}_i$ is a BDDL file and $\mathcal{X}_i$ is a NumPy file storing the flattened MuJoCo simulation state. The filename stem is shared by the two files.

% 场景资产按任务套件、任务实例和回合进行组织。典型的目录结构如下所示：
The scene assets are organized by task suite, task instance, and episode. A typical directory structure is shown below:

\begin{verbatim}
selected_bddl_scenes_<suite>/
+-- <task_name>/
    +-- episode_<id>/
        |-- step_<timestamp>.bddl
        +-- step_<timestamp>_sim_state.npy
\end{verbatim}

% 例如，step_0400.bddl 和 step_0400_sim_state.npy 共同描述了在特定执行时间戳处记录的场景状态。
For example, \texttt{step\_0400.bddl} and \texttt{step\_0400\_sim\_state.npy} jointly describe the scene state recorded at a particular execution timestamp.

% BDDL 格式
\subsubsection{BDDL Format}
\label{sec:bddl_task_description}

% 评估恢复场景使用继承自 LIBERO 的 BDDL 格式。每个 BDDL 文件定义了符号化任务和相应的场景布局，而在记录的时间戳处的物理仿真器状态则单独存储在配对的 _sim_state.npy 文件中。典型的 BDDL 文件遵循以下模板：
The Evaluation Recovery Scenarios use the BDDL format inherited from LIBERO. Each BDDL file defines the symbolic task and the corresponding scene layout, while the physical simulator state at the recorded timestamp is stored separately in a paired \texttt{\_sim\_state.npy} file. A typical BDDL file follows the template below:

\begin{promptbox}[Simplified BDDL Template]
\label{box:bddl_template}
\begin{lstlisting}[
    language=Lisp,
    basicstyle=\small\ttfamily,
    keywordstyle=\color{black},
    commentstyle=\color{gray},
    stringstyle=\color{black},
    columns=fullflexible,
    keepspaces=true,
    showstringspaces=false,
    breaklines=true,
    breakatwhitespace=true,
    numbers=none,
    frame=none,
    aboveskip=0pt,
    belowskip=0pt,
    xleftmargin=1em
]
(define (problem <problem_name>)
  (:domain robosuite)
  (:language <task_instruction>)

  (:regions
    (<object_init_region>
      (:target <support_surface>)
      (:ranges ((x_min y_min x_max y_max)))
      (:yaw_rotation ((yaw_min yaw_max)))
    )
    ...
    (<task_region>
      (:target <target_fixture>)
    )
  )

  (:fixtures
    <fixture_1> - <fixture_type>
    ...
  )

  (:objects
    <object_1> - <object_type>
    ...
  )

  (:obj_of_interest
    <object_1>
    ...
  )

  (:init
    (On <object_1> <object_1_init_region>)
    ...
  )

  (:goal
    (And
      <goal_predicate_1>
      ...
    )
  )
)
\end{lstlisting}
\end{promptbox}

% :language 字段指定自然语言任务指令。:fixtures 和 :objects 字段定义静态场景元素和可移动物体。:regions 字段指定物体初始化区域和任务特定的目标区域。:init 块将物体与其对应的初始化区域相关联，而 :goal 块则保留了原始的符号化任务目标。
The \texttt{:language} field specifies the natural-language task instruction. The \texttt{:fixtures} and \texttt{:objects} fields define the static scene elements and movable objects. The \texttt{:regions} field specifies object initialization regions and task-specific target regions. The \texttt{:init} block associates objects with their corresponding initialization regions, whereas the \texttt{:goal} block preserves the original symbolic task objective.

% BDDL 解析约束
\subsubsection{BDDL Parsing Constraint.}

% 在数据集准备过程中，我们观察到当相同类型的物体在不同行声明时，BDDL 解析器可能会报错。例如，在包含两个盘子的场景中，以下声明可能会触发解析错误：
During dataset preparation, we observed that the BDDL parser may fail when objects of the same type are declared on separate lines. For example, in a scene containing two plates, the following declaration can trigger a parsing error:
\begin{lstlisting}[language=lisp]
(:objects
  plate_1 - plate
  plate_2 - plate
)
\end{lstlisting}

% 当发生该错误时，必须将声明重写为单个类型化列表：
When this error occurs, the declarations must be rewritten as a single typed list:
\begin{lstlisting}[language=lisp]
(:objects
  plate_1 plate_2 - plate
)
\end{lstlisting}

% 因此，在发布的 BDDL 文件中，共享相同类型的所有物体都应声明在同一行。此要求是由我们 LIBERO-Recover 流水线中使用的具体 BDDL 解析实现所决定的。它仅影响任务定义的文本表示形式，并不改变物体的身份标识、属性或语义。
\textbf{All objects sharing the same type should therefore be declared on one line in the released BDDL files.} This requirement is imposed by the specific BDDL parsing implementation used in our LIBERO-Recover pipeline. It only affects the textual representation of the task definition and does not change the identities, properties, or semantics of the objects.

\subsection{Expert Recovery Trajectories}
\label{sec:expert_recovery_trajectories}

% “专家恢复轨迹”提供了成功恢复行为的人类示范。每条轨迹记录恢复执行过程中的机器人观测、机器人状态、动作以及与任务相关的元数据。这些轨迹描述了人类专家如何响应受扰动的场景，处理由失败造成的异常状态，并继续朝原始任务目标执行。
The \emph{Expert Recovery Trajectories} provide human demonstrations of successful recovery behaviors. Each trajectory records robot observations, robot states, actions, and task-related metadata throughout the recovery process. These trajectories capture how human experts respond to disrupted scenes, correct states caused by execution failures, and continue execution toward the original task goal.

% 这些示范可用于监督策略训练、恢复策略自适应或恢复行为的定性分析。与定义基准测试初始条件的评估场景不同，专家恢复轨迹提供了从失败状态出发的动作级监督信号，从而支持模型学习如何实际执行恢复。
The demonstrations can be used for supervised policy training, recovery policy adaptation, or qualitative analysis of recovery behaviors. In contrast to the evaluation scenarios, which define the initial conditions for benchmarking, the expert trajectories provide action-level supervision from failure states, enabling models to learn how recovery should be executed.

\subsubsection{Data Format}
\label{sec:expert_recovery_format}

% 专家恢复轨迹采用 LeRobot 数据格式。该格式针对机器人学习中的多模态时序数据进行统一组织，将低维状态和动作与视觉观测分离存储，同时通过时间索引保持不同模态之间的同步关系。
The recovery demonstrations are released in the LeRobot format~\cite{cadene2026lerobot}, which provides a standardized representation for multimodal robot-learning trajectories. Following its episode-based organization, low-dimensional state--action data and visual observations are stored separately while remaining temporally synchronized within each recovery episode.

% 低维轨迹数据使用 Parquet 文件保存，包括机器人状态、末端执行器动作以及相应的时间索引；视觉信息以视频形式保存，包括第三人称相机和腕部相机两个同步视角。
The low-dimensional trajectory data are stored in Parquet files and contain robot states, end-effector actions, and temporal indices. Visual observations are encoded separately as video streams, including synchronized third-person and wrist-camera views. This separation follows the LeRobot design principle of storing high-frequency tabular robot signals and visual observations using storage formats suited to their respective modalities~\cite{cadene2026lerobot}.

% 数据整体由 data、videos 和 meta 三部分组成。其中 data 保存状态和动作轨迹，videos 保存各相机视角的视频，meta 保存任务、episode 和数据集级别的信息。
At the dataset level, the release is organized into three major components: \texttt{data}, \texttt{videos}, and \texttt{meta}. The \texttt{data} component stores state--action trajectories, \texttt{videos} stores the corresponding camera observations, and \texttt{meta} maintains dataset-, task-, and episode-level information. A representative organization is shown below:

\begin{verbatim}
LIBERO_Recovery_Expert/
|-- data/
|   `-- chunk-xxx/
|       |-- episode_xxxxxx.parquet
|       `-- ...
|-- videos/
|   `-- chunk-xxx/
|       |-- <third-person-camera>/
|       |   `-- episode_xxxxxx.mp4
|       `-- <wrist-camera>/
|           `-- episode_xxxxxx.mp4
`-- meta/
    |-- info.json
    |-- episodes.jsonl
    |-- episodes_stats.jsonl
    `-- tasks.jsonl
\end{verbatim}

% 这种标准化形式使恢复轨迹能够直接接入现有 LeRobot / VLA 数据加载与训练流程，并方便与原始 LIBERO 示范共同使用。
This standardized representation facilitates integration with existing LeRobot-based robot-learning pipelines and VLA training frameworks such as StarVLA~\cite{starvla2026}. It also makes the recovery demonstrations compatible with the broader LIBERO~\cite{libero} data ecosystem, enabling recovery data to be combined with conventional task demonstrations during policy adaptation.

\subsubsection{Dataset Scale and Coverage}
\label{sec:expert_recovery_scale}

% 数据由 4 名遥操作人员使用 SpaceMouse 从 413 个失败场景出发采集，共形成 3,184 条成功恢复轨迹、625,731 帧，并覆盖 110 种任务类型。
The released dataset contains 3,184 human recovery episodes collected by four teleoperators using a SpaceMouse from 413 failure scenarios. In total, the dataset contains 625,731 trajectory frames and covers 110 task types.

% 所有轨迹均以 20 Hz 记录，同时包含第三人称视角、腕部相机视角、机器人关节状态以及末端执行器动作。
All trajectories are recorded at 20\,Hz and provide two synchronized visual streams, including a third-person view and a wrist-camera view, together with robot joint states and end-effector actions. Consequently, the dataset preserves the complete observation--state--action evolution of the recovery process rather than only the final recovery outcome.

% 与 LIBERO 原始数据主要描述从标准初始状态完成任务的专家示范不同，LIBERO-Recover 的专家轨迹以失败状态为起点，重点记录模型如何重新建立可执行状态并继续完成原始任务。
In contrast to the original LIBERO demonstrations~\cite{libero}, which primarily capture task execution from standard task initializations, the recovery trajectories start from disrupted states and explicitly capture the sequence of actions required to restore progress toward the original task objective.

\subsection{Pre-failure Temporal Histories}
\label{sec:pre_failure_temporal_histories}

% “失败前时序历史”由 LIBERO-10 执行轨迹中失败发生前的短时序片段构成。我们使用 Qwen3.5-27B-Instruct 对完整执行视频进行分析，并对执行过程开始偏离预期任务进程的位置进行时间定位。根据模型给出的时间戳，将该时刻与原始轨迹对齐，并截取其之前的一段观测和机器人状态作为失败前历史。这些序列保留了机器人逐渐进入异常状态的动态上下文，包括物体位姿、末端执行器运动以及夹爪状态的变化。
The \emph{Pre-failure Temporal Histories} contain short temporal segments extracted from LIBERO-10 execution trajectories prior to failure. We use Qwen3.5-27B-Instruct to analyze the complete execution videos and temporally localize the point at which the rollout begins to deviate from the expected task progression. The resulting timestamp is aligned with the original trajectory, from which a preceding window of observations and robot states is extracted as the pre-failure history. These sequences preserve the dynamic context of how the robot approaches an abnormal state, including changes in object poses, end-effector motion, and gripper state.

% 失败前历史作为可选的时序上下文组件提供。策略既可以仅使用当前恢复观测，也可以在支持时序记忆时进一步访问失败发生前的历史。
The pre-failure histories are provided as an optional temporal-context component. A policy may use only the current recovery observation, or additionally access the preceding history when temporal memory is supported.

% 评估流程
\section{Recovery Evaluation Protocol}

% 评估流程
\subsection{Evaluation Pipeline}
\label{app:evaluation_pipeline}

\begin{figure*}[t]
    \centering
    \includegraphics[width=\textwidth]{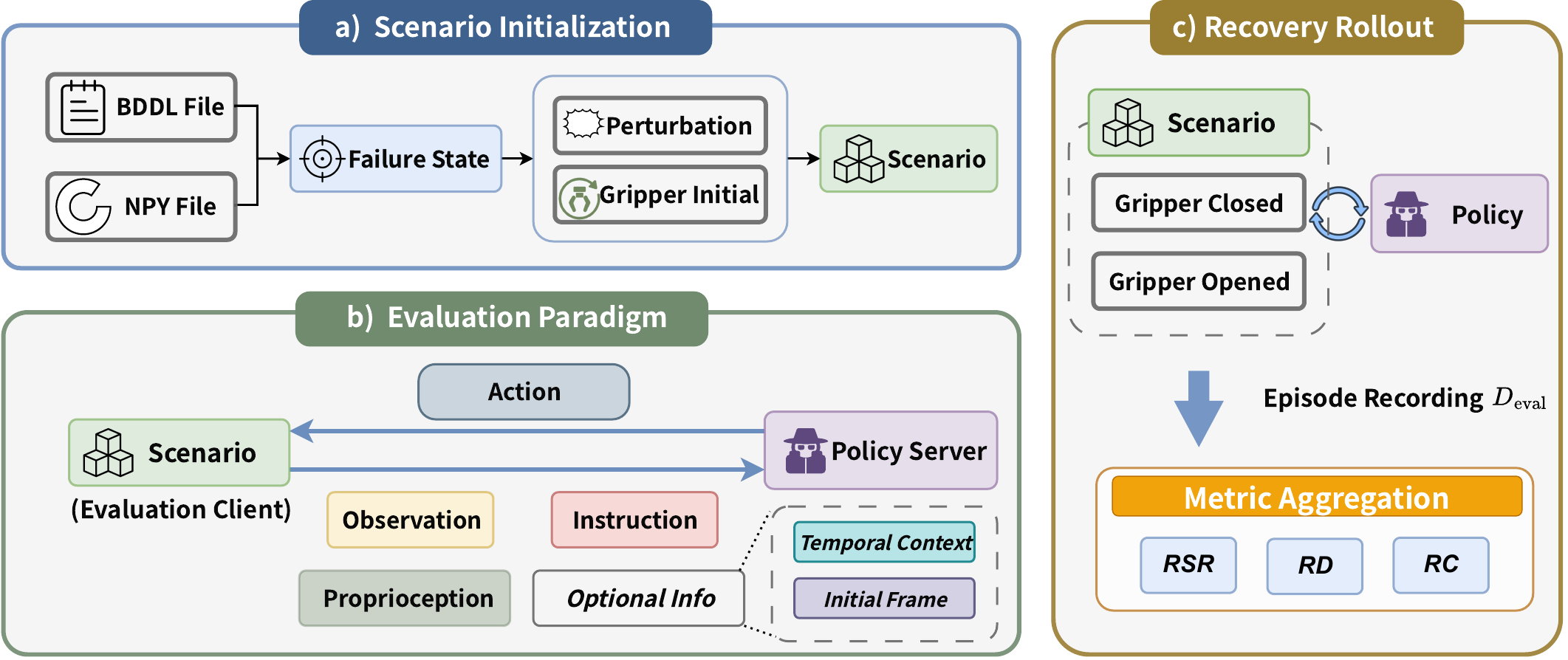}
    % LIBERO-Recover 评估流程概览。评估包含三个阶段：(a) 场景初始化：利用 BDDL 任务定义和记录的仿真器状态重建失败状态，随后通过物体扰动和夹爪初始化生成初始化的恢复场景。(b) 评估范式：评估客户端将当前观测、任务指令、机器人本体感觉以及可选的时序上下文发送给策略服务器，后者返回下一个动作。(c) 恢复推演：在张开和闭合夹爪条件下评估策略，生成汇总为 RSR、RD 和 RC 的回合级评估记录。
    \caption{\textbf{Overview of the LIBERO-Recover evaluation pipeline.} The evaluation consists of three stages. (a) \emph{Scenario Initialization}: the BDDL task definition and the recorded simulator state are used to reconstruct a failure state, after which object perturbation and gripper initialization produce an initialized recovery scenario. (b) \emph{Evaluation Paradigm}: the evaluation client sends the current observation, task instruction, robot proprioception, and optional temporal context to the policy server, which returns the next action. (c) \emph{Recovery Rollout}: the policy is evaluated under open- and closed-gripper conditions, producing episode-level evaluation records that are aggregated into RSR, RD, and RC.}
    \label{fig:evalpipeline}
\end{figure*}

% LIBERO-Recover 评估操作策略在执行进入中断状态后能否恢复执行原始任务。图~\ref{fig:evaluation\_pipeline} 总结了完整的评估协议，包括场景初始化、策略与环境交互、恢复推演以及指标汇总。与传统的从标称任务初始化开始的 LIBERO 评估不同，每个恢复回合均从与所记录执行失败相关的仿真器状态开始。原始任务指令和目标条件得以保留，而提供给策略的物理状态则反映了失败上下文以及受控初始化协议。
LIBERO-Recover evaluates whether a manipulation policy can resume an original task after execution has entered a disrupted state. Figure~\ref{fig:evalpipeline} summarizes the complete evaluation protocol, which consists of scenario initialization, policy--environment interaction, recovery rollout, and metric aggregation. Unlike conventional LIBERO evaluation, where a policy starts from a nominal task initialization, each recovery episode starts from a simulator state associated with a recorded execution failure. The original task instruction and goal condition are preserved, while the physical state provided to the policy reflects the failure context and the controlled initialization protocol.

% 对于策略 $\pi$ 和恢复场景 $\mathcal{S}_i$，评估流程构建一个初始状态：
For a policy $\pi$ and a recovery scenario $\mathcal{S}_i$, the evaluation procedure constructs an initialized state
\begin{equation}
\widetilde{s}_i
=
\mathcal{I}\!\left(
s_i^{\mathrm{fail}},
\delta_i^{\mathrm{obj}},
g_i^{\mathrm{init}}
\right),
\end{equation}
% 其中 $s_i^{\mathrm{fail}}$ 表示记录的失败状态，$\delta_i^{\mathrm{obj}}$ 表示物体状态扰动，而 $g_i^{\mathrm{init}}$ 表示夹爪初始化条件。
where $s_i^{\mathrm{fail}}$ denotes the recorded failure state, $\delta_i^{\mathrm{obj}}$ denotes the object-state perturbation, and $g_i^{\mathrm{init}}$ denotes the gripper initialization condition.

% 随后策略接收原始任务指令以及由 $\widetilde{s}_i$ 生成的观测，并生成闭环动作序列：
The policy then receives the original task instruction together with the observation generated from $\widetilde{s}_i$ and produces a closed-loop action sequence
\begin{equation}
\tau_i^{\mathrm{rec}}
=
\left(
\widetilde{s}_i,a_0,s_1,a_1,\ldots,s_T
\right).
\end{equation}
% 回合结果由最终状态是否在允许的交互步长内满足原始任务目标来决定。
The episode outcome is determined by whether the resulting state satisfies the original task goal within the allowed interaction horizon.

% 完整的评估流程包含四个逻辑有序的阶段：首先，恢复记录的失败状态以重建恢复开始时的物理构型；其次，应用受控初始化以生成评估条件，包括物体状态扰动和夹爪状态初始化，这些操作在保留任务指令和符号化目标的同时修改了物理起始条件；第三，策略以闭环方式与环境交互，重复利用当前观测生成动作，直到任务完成或推演终止；最后，将完整的轨迹、初始化条件和任务结果记录为回合级评估证据，并在回合、任务、恢复等级和任务套件之间进行汇总。
The complete evaluation process contains four logically ordered stages. First, the recorded failure state is restored to reconstruct the physical configuration from which recovery begins. Second, controlled initialization is applied to generate the evaluation condition, including object-state perturbation and gripper-state initialization. These operations modify the physical starting condition while preserving the task instruction and symbolic goal. Third, the policy interacts with the environment in a closed-loop manner, repeatedly using the current observation to generate actions until the task is completed or the rollout terminates. Finally, the complete trajectory, initialization condition, and task outcome are recorded as episode-level evaluation evidence and aggregated across episodes, tasks, recovery levels, and task suites.

% 我们的实现使用 StarVLA~\cite{starvla2026} 的统一视觉-语言-动作接口将策略与 LIBERO 环境相连接。评估协议本身独立于特定的策略架构：不同的策略被提供相同的任务指令、初始状态、扰动条件和交互预算。因此，测得的性能反映了策略理解受干扰任务状态并从中恢复的能力，而不是回合构建上的差异。
Our implementation uses the unified visual--language--action interface of StarVLA~\cite{starvla2026} to connect the policy with the LIBERO environment. The evaluation protocol itself is independent of a particular policy architecture: different policies are provided with the same task instructions, initialized states, perturbation conditions, and interaction budget. Consequently, the measured performance reflects the policy's ability to interpret and recover from disrupted task states rather than differences in episode construction.

% 失败状态扰动与初始化
\subsection{Failure-State Perturbation}

% 在恢复记录的仿真器状态后，我们通过两种受控干预来构建面向恢复的初始条件：物体状态扰动和夹爪状态初始化。对于每个自由关节物体，我们在水平面上独立采样均匀位移，并将所产生的偏移量应用于其位置。在默认设置下，最大位移为 $\pm 2\,\mathrm{cm}$。由铰链关节或滑动关节控制的物体（如抽屉和炉灶按钮）不参与此操作。
After restoring the recorded simulator state, we construct a recovery-oriented initial condition through two controlled interventions: object-state perturbation and gripper-state initialization. For each free-joint object, we independently sample a uniform displacement in the horizontal plane and apply the resulting offset to its position. In the default setting, the maximum displacement is $\pm 2\,\mathrm{cm}$. This small perturbation is intended to assess local generalization around recorded failure states while limiting changes to the underlying failure configuration. Objects controlled by hinge or slide joints, such as drawers and stove buttons, are excluded from this operation.

\begin{figure}[t]
    \centering
    \includegraphics[width=\linewidth]{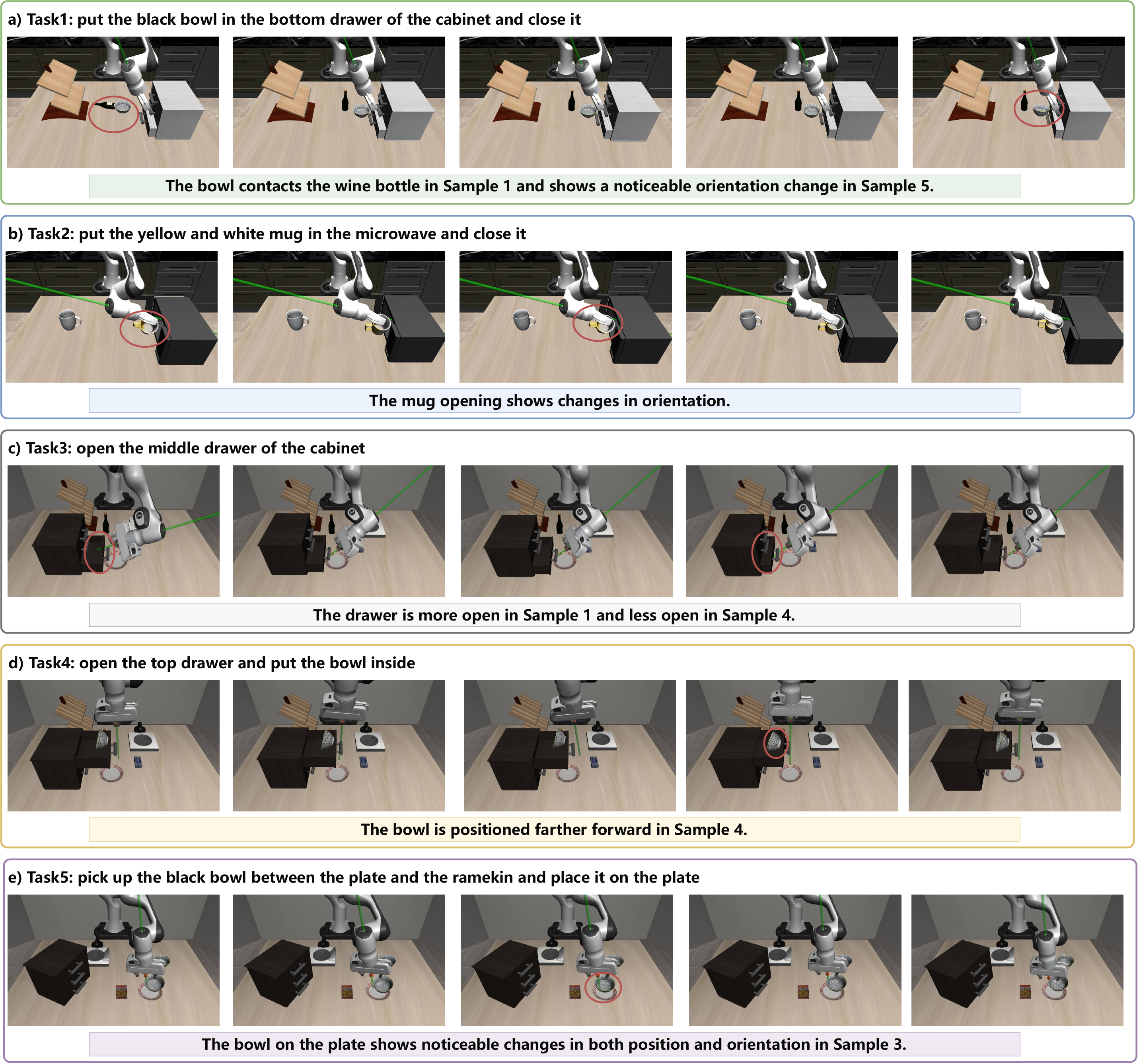}
    % 五个代表性任务中的受控物体扰动示例。对于每个任务，通过对自由关节物体施加微小的水平位移来采样五个扰动状态，同时任务指令和场景语义保持不变。圈出区域突出了物体位置、接触或朝向的代表性变化。
    \caption{Controlled object perturbations across five representative tasks, with five samples per task. Small horizontal displacements ($\pm 2$\,cm) introduce local variations while largely preserving the overall failure configurations shown here. This design probes local generalization around recorded failure states. Circled regions highlight changes in object position, contact, or orientation.}
    \label{fig:object_perturbation_example}
\end{figure}

% 图~\ref{fig:object_perturbation_example} 展示了五个代表性任务中各自的五个扰动样本。尽管所有样本均保留了原始任务指令和场景语义，但被扰动的物体呈现出不同的位置和朝向。例如，第一个样本中的碗发生了足够的位移以至于接触到了酒瓶，而第五个样本中的碗则表现出明显的朝向改变。这些示例说明了微小且受控的状态变化如何产生各不相同且面向恢复的初始条件。
Figure~\ref{fig:object_perturbation_example} presents five perturbation samples for each of five representative tasks. Although all samples preserve the original task instruction and scene semantics, the perturbed objects exhibit different positions and orientations. For example, the bowl in the first sample is displaced sufficiently to contact the wine bottle, while the bowl in the fifth sample shows a noticeable orientation change. These examples illustrate how small, controlled state variations produce distinct recovery-oriented initial conditions.

\subsection{Gripper Initialization}

\begin{figure}[t]
    \centering
    \includegraphics[width=\linewidth]{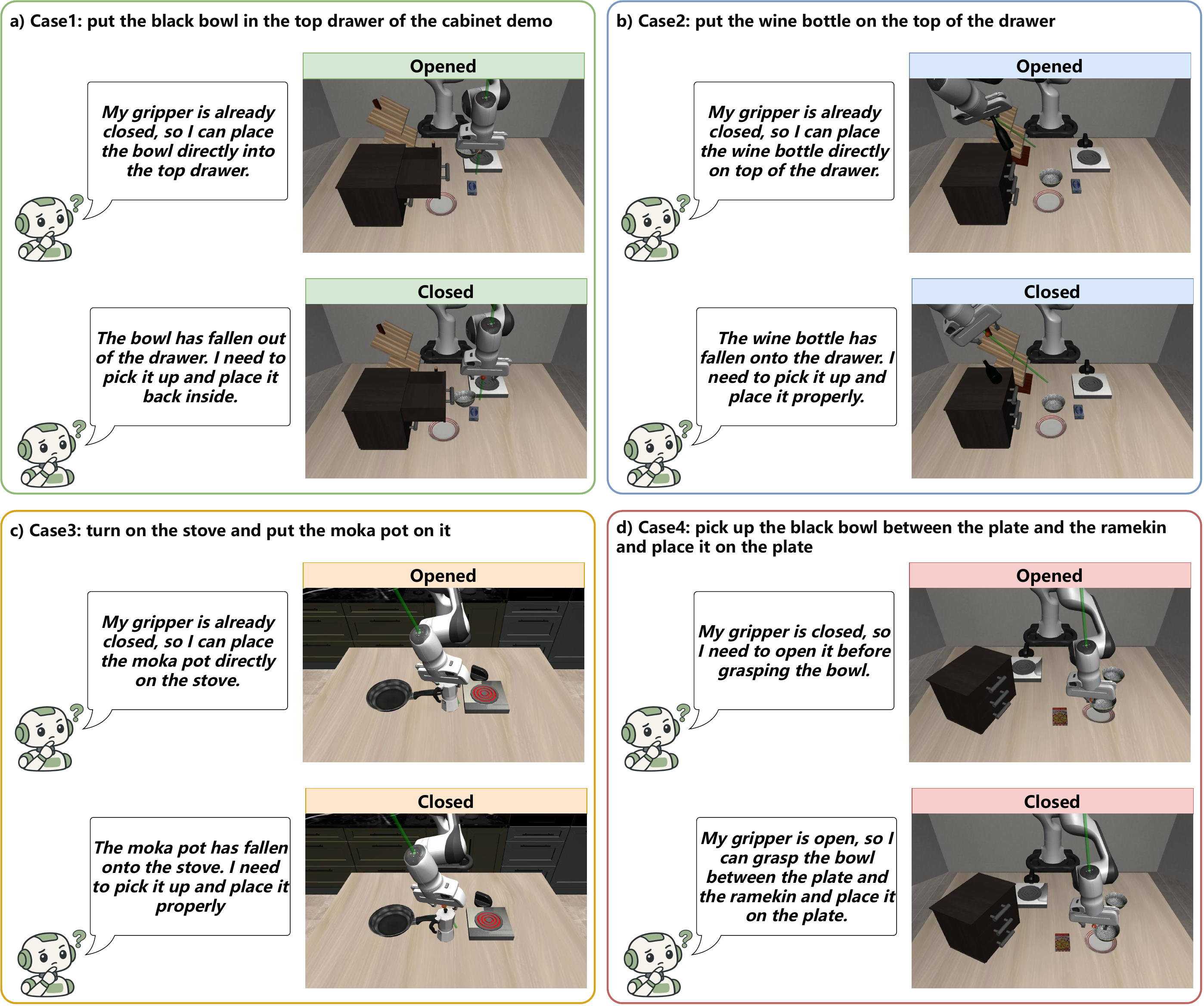}
    % 恢复评估中使用的两种夹爪初始化条件示例。机器人以张开或完全闭合的夹爪开始，而机械臂位姿和周围场景保持不变。这两种条件需要策略做出不同的响应，例如直接执行任务或先松开夹爪以恢复预期的操作状态。
    \caption{Examples of the two gripper initialization conditions used in recovery evaluation. The robot starts with either an open or a fully closed gripper, while the arm pose and surrounding scene remain unchanged. The two conditions require different responses from the policy, such as directly executing the task or first releasing the gripper to recover the intended manipulation state.}
    \label{fig:gripper_initialization}
\end{figure}

% 在状态扰动之后，机器人在两种夹爪条件之一中进行初始化。在张开夹爪设置下，夹爪被驱动至其标称张开构型。在闭合夹爪设置下，夹爪初始完全闭合，从而创造出一种在标准 LIBERO 执行中通常不会遇到的异常初始条件。目标夹爪构型是通过执行固定步数的零增量机械臂动作达到稳定状态来实现的，而不是通过直接瞬移关节状态。在扰动之后，控制器目标也会重新锚定到当前的末端执行器位姿，以防止推演开始时发生非自然的机械臂位移。图~\ref{fig:gripper_initialization} 展示了评估协议中使用的两种初始化条件。
Following the state perturbation, the robot is initialized under one of two gripper conditions. In the open-gripper setting, the gripper is driven to its nominal open configuration. In the closed-gripper setting, the gripper starts fully closed, creating an abnormal initial condition that is typically not encountered in standard LIBERO execution. The target gripper configuration is reached by executing zero-delta arm actions for a fixed number of settling steps rather than by directly teleporting the joint state. The controller goal is also re-anchored to the current end-effector pose after the perturbation, preventing an artificial arm displacement at the beginning of the rollout. Figure~\ref{fig:gripper_initialization} shows the two initialization conditions used in the evaluation protocol.

% 对于每个场景，评估器在相同的任务定义下执行固定数量的张开夹爪和闭合夹爪试验。所生成的评估子集隔离了物体状态扰动和异常夹爪初始化的影响，而固定的随机种子确保了采样的扰动能够在不同策略之间进行复现。
For each scenario, the evaluator performs a fixed number of open-gripper and closed-gripper trials under the same task definition. The resulting evaluation splits isolate the effects of object-state perturbations and abnormal gripper initialization, while the fixed random seed ensures that the sampled perturbations can be reproduced across different policies.

% 即使机械臂位姿和场景配置完全相同，这两种夹爪条件也可能导致不同的实际有效任务状态。在张开夹爪的情况下，策略可以直接开始预期的操作，例如抓取物体、举起物体或将其放置在目标位置。然而，在闭合夹爪的情况下，机器人可能需要在执行标称任务之前先释放夹爪，或者对初始化期间被移位或掉落的物体做出反应。因此，夹爪初始化不仅改变了执行器状态，还创造了独特的决策点，要求策略理解当前场景并选择恰当的恢复动作。图~\ref{fig:gripper_initialization} 中的示例说明了跨不同操作指令的这些取决于任务的响应。
The two gripper conditions may lead to different effective task states even when the arm pose and scene configuration are identical. With an open gripper, the policy can begin the intended manipulation directly, such as grasping an object, lifting it, or placing it at the target location. With a closed gripper, however, the robot may need to release the gripper before executing the nominal task, or respond to an object that has been displaced or dropped during initialization. Thus, gripper initialization does not simply change the actuator state; it creates distinct decision points that require the policy to interpret the current scene and select an appropriate recovery action. The examples in Figure~\ref{fig:gripper_initialization} illustrate these task-dependent responses across different manipulation instructions.

% 恢复推演
\subsection{Recovery Rollout}
\label{app:recovery_rollout}

% 在恢复场景初始化后，策略从所产生的状态 $\widetilde{s}_0$ 开始并尝试完成原始任务。在每个控制步 $t$，策略接收原始语言指令 $l$、当前视觉观测 $o_t$ 以及机器人本体感觉状态 $r_t$。当启用时序上下文时，还会额外提供历史表示 $\mathcal{H}_t$。随后策略根据下式生成下一个动作：
After the recovery scenario has been initialized, the policy starts from the resulting state $\widetilde{s}_0$ and attempts to complete the original task. At each control step $t$, the policy receives the original language instruction $l$, the current visual observation $o_t$, and the robot proprioceptive state $r_t$. When temporal context is enabled, a history representation $\mathcal{H}_t$ is additionally provided. The policy then generates the next action according to
\begin{equation}
a_t =
\pi_{\theta}
\left(
l, o_t, r_t, \mathcal{H}_t
\right),
\end{equation}
% 对于在没有时序上下文的情况下进行评估的策略，$\mathcal{H}_t$ 则被省略。
where $\mathcal{H}_t$ is omitted for policies evaluated without temporal context.

% 预测的动作在环境中执行，产生新的物理状态和相应的观测：
The predicted action is executed in the environment, resulting in a new physical state and the corresponding observation:
\begin{equation}
s_{t+1}=f_{\mathrm{env}}(s_t,a_t),
\qquad
(o_{t+1},r_{t+1})=h(s_{t+1}).
\end{equation}
% 新生成的观测随后反馈给策略以进行下一次决策。该过程形成了一个闭环恢复交互，其中策略必须根据自身先前决策所产生的状态不断调整其动作。与重放固定的成功轨迹相比，这种推演允许策略对物体位移、夹爪不匹配、意外接触以及由恢复初始化引入的其他状态变化做出反应。
The newly generated observation is then fed back to the policy for the next decision. This process forms a closed-loop recovery interaction in which the policy must continuously adapt its actions to the state produced by its own previous decisions. In contrast to replaying a fixed successful trajectory, the rollout therefore allows the policy to respond to object displacement, gripper mismatch, unexpected contact, and other state changes introduced by the recovery initialization.

% 每个场景均在张开夹爪和闭合夹爪两种初始化条件下进行评估。除另有说明外，每种条件执行五次试验，每个场景共进行十次恢复推演。策略最多允许与环境交互 $H=520$ 个控制步。当环境在达到最大步长前发出任务完成信号时，推演被标记为成功。如果未能在线路预算内达到任务完成条件，则该推演被标记为不成功。
Each scenario is evaluated under both open-gripper and closed-gripper initialization conditions. Unless otherwise specified, five trials are performed for each condition, resulting in ten recovery rollouts per scenario. The policy is allowed to interact with the environment for at most $H=520$ control steps. A rollout is marked as successful when the environment emits the task-completion signal before the horizon is reached. If the task-completion condition is not reached within the available interaction budget, the rollout is marked as unsuccessful.

% 对于第 $i$ 次推演，我们记录轨迹：
For the $i$-th rollout, we record the trajectory
\begin{equation}
\tau_i =
\left(
s_0^i,a_0^i,s_1^i,a_1^i,\ldots,s_{T_i}^i
\right),
\end{equation}
% 以及其任务标识符、恢复等级、夹爪初始化、物体扰动和二值结果 $y_i\in\{0,1\}$。这些回合级记录构成了指标汇总阶段所使用的基础依据。
together with its task identifier, recovery level, gripper initialization, object perturbation, and binary outcome $y_i\in\{0,1\}$. These episode-level records constitute the basic evidence used by the metric aggregation stage.

% 指标汇总
\subsection{Metric Aggregation}
\label{app:metric_aggregation}

% 指标汇总阶段将回合级的恢复结果转化为正文中报告的基准测试统计数据。每次评估的推演都与任务指令、恢复场景、恢复难度等级、任务套件标签、夹爪初始化条件以及二值任务结果相关联。记录的轨迹还包含观测、机器人状态、动作以及基于阶段分析所需的时序信息。
The metric aggregation stage converts episode-level recovery outcomes into the benchmark statistics reported in the main paper. Each evaluated rollout is associated with a task instruction, a recovery scenario, a recovery difficulty level, a task-suite label, a gripper initialization condition, and a binary task outcome. The recorded trajectory additionally contains the observations, robot states, actions, and the temporal information required for phase-based analysis.

% 回合级结果。
\paragraph{Episode-level outcome.}

% 对于第 $i$ 次恢复推演，我们根据最终任务状态定义二值成功指示变量 $y_i$：
For the $i$-th recovery rollout, we define a binary success indicator $y_i$ according to the final task state:
\begin{equation}
y_i =
\begin{cases}
1, & \text{if the original task goal is satisfied},\\
0, & \text{otherwise}.
\end{cases}
\end{equation}
% 因此，只有当策略从初始化的恢复状态开始并完成原始任务时，推演才被视为成功。如果策略在未满足任务目标的情况下达到最大交互步长，则该推演被记录为不成功的恢复尝试。
A rollout is therefore considered successful only when the policy completes the original task after starting from the initialized recovery state. If the policy reaches the maximum interaction horizon without satisfying the task goal, the rollout is recorded as an unsuccessful recovery attempt.

% 令 $\mathcal{E}$ 表示在特定评估条件下评估的恢复回合集合。该集合可以对应于一个模型的所有回合、特定任务套件、单一恢复等级、单一夹爪初始化条件或这些条件的任意组合。恢复成功率（RSR）计算为：
Let $\mathcal{E}$ denote a set of evaluated recovery episodes under a particular evaluation condition. This set may correspond to all episodes of a model, a specific task suite, one recovery level, one gripper initialization condition, or any combination of these conditions. The Recovery Success Rate (RSR) is computed as
\begin{equation}
\mathrm{RSR}(\mathcal{E})
=
\frac{1}{|\mathcal{E}|}
\sum_{i\in\mathcal{E}} y_i .
\end{equation}
% 因此，每次推演贡献一个等权重的成功或失败样本。总体 RSR 是在所有有效恢复回合上计算的，而特定等级和特定套件的 RSR 值则是通过将 $\mathcal{E}$ 限制在相应子集上获得的。
Thus, each rollout contributes one equally weighted success or failure sample. The overall RSR is computed over all valid recovery episodes, whereas the level-specific and suite-specific RSR values are obtained by restricting $\mathcal{E}$ to the corresponding subset.

% 例如，$\mathrm{RSR}_{L_3}$ 仅使用分配给物体状态恢复的恢复回合进行计算。它度量了当失败需要恢复任务相关物体状态时，受评估策略完成原始任务的频率。同样地，张开夹爪和闭合夹爪的结果是使用与相应初始化条件相关的回合计算的。这种分组使得报告的分数能够将总体恢复能力与特定失败难度或初始化条件下的表现区分开来。
For example, $\mathrm{RSR}_{L_3}$ is computed using only the recovery episodes assigned to object-state recovery. It measures how often the evaluated policy can complete the original task when the failure requires the restoration of a task-relevant object state. Likewise, the open-gripper and closed-gripper results are computed using the episodes associated with the corresponding initialization condition. This grouping allows the reported scores to distinguish overall recovery ability from performance under a particular failure difficulty or initialization condition.

% 场景级与任务级汇总。
\paragraph{Scenario- and task-level aggregation.}

% 一个恢复场景可以通过多次试验进行评估，因为同一初始化场景可以在不同的随机物体扰动和夹爪条件下进行测试。当报告单个场景的结果时，与该场景关联的所有有效试验都包含在相应的场景级恢复率中。
A recovery scenario may be evaluated through multiple trials because the same initialized scene can be tested with different random object perturbations and gripper conditions. When the result is reported for one scenario, all valid trials associated with that scenario are included in the corresponding scenario-level recovery rate.

% 任务级汇总是针对与同一原始 LIBERO 任务相关联的所有恢复场景进行的。令 $\mathcal{E}_k$ 表示源自任务 $k$ 的所有恢复回合的集合。该集合可能包含多个失败状态、多个扰动实例以及在不同夹爪初始化下的重复试验。因此，任务 $k$ 的恢复率为：
Task-level aggregation is performed over all recovery scenarios associated with the same original LIBERO task. Let $\mathcal{E}_k$ denote the set of all recovery episodes derived from task $k$. This set may contain multiple failure states, multiple perturbation instances, and repeated trials under different gripper initializations. The recovery rate of task $k$ is therefore
\begin{equation}
R_k
=
\frac{1}{|\mathcal{E}_k|}
\sum_{i\in\mathcal{E}_k} y_i .
\end{equation}
% 换句话说，$R_k$ 代表与任务 $k$ 相关的所有失败场景上的恢复成功率，而不是单个代表性失败场景的结果。当分别分析两种夹爪条件时，$\mathcal{E}_k$ 相应地限制在张开夹爪或闭合夹爪子集中。
In other words, $R_k$ represents the recovery success rate over all failure scenarios related to task $k$, rather than the result of one representative failure scenario. When the two gripper conditions are analyzed separately, $\mathcal{E}_k$ is correspondingly restricted to the open-gripper or closed-gripper subset.

% 恢复退化度。
\paragraph{Recovery degradation.}

% 恢复退化度（RD）度量执行进入失败状态后损失了多少执行能力。每次推演分为三个时序阶段：失败前、失败中和失败后。失败前阶段描述首次出现任务相关偏差之前的执行过程，而失败后阶段则描述随后的执行过程，在此过程中策略必须对受干扰的状态做出反应并尝试完成原始任务。
Recovery Degradation (RD) measures how much execution capability is lost after the execution enters a failure state. Each rollout is divided into three temporal phases: Pre-Failure, During-Failure, and Post-Failure. The Pre-Failure phase describes the execution before the first task-relevant deviation, whereas the Post-Failure phase describes the subsequent execution in which the policy must respond to the disrupted state and attempt to complete the original task.

% 令 $m_{\mathrm{pre}}^i$ 和 $m_{\mathrm{post}}^i$ 分别表示推演 $i$ 在失败前和失败后的特定阶段执行性能。这些量在相同的评估子集上进行汇总：
Let $m_{\mathrm{pre}}^i$ and $m_{\mathrm{post}}^i$ denote the phase-specific execution performance of rollout $i$ before and after the failure, respectively. These quantities are aggregated over the same evaluation subset:
\begin{equation}
M_{\mathrm{pre}}
=
\frac{1}{|\mathcal{E}|}
\sum_{i\in\mathcal{E}}m_{\mathrm{pre}}^i,
\qquad
M_{\mathrm{post}}
=
\frac{1}{|\mathcal{E}|}
\sum_{i\in\mathcal{E}}m_{\mathrm{post}}^i .
\end{equation}
% 这里，$M_{\mathrm{pre}}$ 是在失败发生前保持的平均执行性能，而 $M_{\mathrm{post}}$ 是失败发生后的平均性能，此时策略必须理解改变的状态、生成纠正动作并恢复任务。这两个值是使用匹配的任务和场景子集计算的，因此它们的差异反映了失败的影响，而非任务构成的改变。
Here, $M_{\mathrm{pre}}$ is the average execution performance maintained before the failure occurs, and $M_{\mathrm{post}}$ is the average performance after the failure, when the policy must interpret the altered state, generate corrective actions, and resume the task. The two values are computed using matched task and scenario subsets so that their difference reflects the effect of failure rather than a change in task composition.

% RD 随后定义为：
RD is then defined as
\begin{equation}
\mathrm{RD}
=
1-
\frac{M_{\mathrm{post}}+\epsilon}
     {M_{\mathrm{pre}}+\epsilon},
\end{equation}
% 其中 $\epsilon$ 是用于数值稳定性的微小常数。如果失败后的性能接近失败前的性能，RD 接近于零，表明策略保留了大部分执行能力。如果失败后的性能趋近于零而失败前的性能保持为正，RD 趋近于 1，表明严重的性能退化。因此，RD 度量的是相对能力保持率，而非绝对恢复成功率。如果策略在失败前的性能显著更高，即使它在失败后取得了非零的成功率，仍可能获得较大的 RD。
where $\epsilon$ is a small constant for numerical stability. If the post-failure performance is close to the pre-failure performance, RD is close to zero, indicating that the policy retains most of its execution capability. If the post-failure performance approaches zero while the pre-failure performance remains positive, RD approaches one, indicating severe degradation. Therefore, RD measures relative capability retention rather than absolute recovery success. A policy may achieve a nonzero post-failure success rate while still obtaining a large RD if its pre-failure performance is substantially higher.

% 恢复一致性。
\paragraph{Recovery consistency.}

% 恢复一致性（RC）度量策略在不同任务之间是否表现出相似的恢复行为。对于每个任务 $k$，$R_k$ 是针对与该任务相关的所有失败场景和试验计算的恢复成功率。RC 定义为：
Recovery Consistency (RC) measures whether a policy exhibits similar recovery behavior across different tasks. For each task $k$, $R_k$ is the recovery success rate computed over all failure scenarios and trials associated with that task. RC is defined as
\begin{equation}
\mathrm{RC}
=
1-
\operatorname{Std}_{k}(R_k),
\end{equation}
% 其中 $\operatorname{Std}_{k}$ 表示跨任务级恢复率的标准差。
where $\operatorname{Std}_{k}$ denotes the standard deviation across task level recovery rates.

% 较高的 RC 表明策略的恢复性能在不同任务之间差异很小。较低的 RC 表明策略在某些任务上表现良好，但在其他任务上严重受挫。RC 刻画的是跨任务恢复行为的稳定性，而非其绝对水平。因此，RC 需要与 RSR 结合解读。例如，一个在几乎所有任务上都失败的策略，如果其失败率保持一致，仍可能获得相对较高的 RC；而同时具有高 RSR 和高 RC 的策略则表现出更强且更稳定的恢复能力。
A high RC indicates that the policy's recovery performance varies little from one task to another. A low RC indicates that the policy performs well on some tasks but fails substantially on others. RC captures the stability of recovery behavior across tasks, rather than its absolute level. Consequently, RC is interpreted jointly with RSR. For example, a policy that fails on nearly all tasks may still obtain a relatively high RC if its failure rates are consistent, whereas a policy with both high RSR and high RC demonstrates stronger and more stable recovery capability.

% 分层报告。
\paragraph{Stratified reporting.}

% 最终结果沿着几个互补的维度进行汇总。首先，报告每个任务套件的结果，以刻画任务构成和场景多样性的影响。其次，将结果按四个恢复等级进行区分，以展示随着恢复需要越来越深层次的状态推理时性能的变化情况。第三，在分析初始执行器条件的影响时，分别报告张开夹爪和闭合夹爪的试验结果。最后，利用任务级恢复率计算 RC，同时利用阶段级性能统计数据计算 RD。
The final results are aggregated along several complementary dimensions. First, results are reported for each task suite to characterize the effect of task composition and scene diversity. Second, results are separated by the four recovery levels to show how performance changes as recovery requires increasingly deeper state reasoning. Third, open-gripper and closed-gripper trials are reported separately when analyzing the effect of the initial actuator condition. Finally, task-level recovery rates are used to compute RC, while phase-level performance statistics are used to compute RD.

% 所有报告的指标均根据相同的回合级记录计算。汇总流程不依赖于策略是零样本评估、使用专家恢复轨迹微调适应，还是提供了可选的时序上下文。这些选择会影响每次推演期间生成的观测、动作和结果，而从所生成的记录到 RSR、RD 和 RC 的映射规则保持固定。
All reported metrics are computed from the same episode-level records. The aggregation procedure does not depend on whether a policy is evaluated zero-shot, adapted with expert recovery trajectories, or provided with optional temporal context. These choices affect the observations, actions, and outcomes generated during each rollout, while the mapping from the resulting records to RSR, RD, and RC remains fixed.

\section{Experimental Implementation Details}
\label{app:exp_details}

% 说明六个模型如何适配 LIBERO 机械臂和输入接口。实验采取的推理精度、控制频率、相机输入、action horizon 和 chunk size，列出对应的表格，这里假设GPU是
% 实验配置
\subsection{Model Adaptation}
\label{app:model_adaptation}

% 我们在 LIBERO 环境中评估了六种具有代表性的操作策略：OpenVLA-OFT、GR00T-N1.5、$\pi_0$、$\pi_0$-Fast、Wan2-Policy 以及 Cosmos-Predict2-Policy。所有策略均部署在相同的 Franka 操作环境中，并接收原始 LIBERO 语言指令以及视觉观测和机器人状态信息。由于这些策略使用不同的输入和动作接口，我们为每个模型实现了一个轻量级的环境适配器。
We evaluate six representative manipulation policies on the LIBERO environment: OpenVLA-OFT\cite{openvla}, GR00T-N1.5\cite{gr00t}, $\pi_0$~\cite{pi_0}, $\pi_0$-Fast~\cite{pi0_fast}, Wan2-Policy\cite{wan}, and Cosmos-Predict2-Policy\cite{cosmos-policy}. All policies are deployed on the same Franka manipulation environment and receive the original LIBERO language instruction together with visual observations and robot state information. Since these policies use different input and action interfaces, we implement a lightweight environment adapter for each model.

% 该适配器将 LIBERO 观测转换为对应策略所期望的格式，包括图像缩放、通道排序、语言指令格式化、本体感觉状态归一化以及动作归一化。预测出的动作被转换回 LIBERO 控制空间并由仿真器执行。该适配器不修改策略架构，也不引入额外的任务特定信息。
The adapter converts LIBERO observations into the format expected by the corresponding policy, including image resizing, channel ordering, language-instruction formatting, proprioceptive-state normalization, and action normalization. The predicted actions are converted back to the LIBERO control space and executed by the simulator. The adapter does not modify the policy architecture or introduce additional task-specific information.

% 对于 OpenVLA-OFT、GR00T-N1.5、$\pi_0$ 和 $\pi_0$-Fast，适配器将语言指令、所需的相机观测以及机器人本体感觉状态提供给相应的 VLA 接口。模型输出一组（chunk）底层机器人动作，在下一次策略查询前按序执行。
For OpenVLA-OFT, GR00T-N1.5, $\pi_0$, and $\pi_0$-Fast, the adapter provides the language instruction, the required camera observations, and the robot proprioceptive state to the corresponding VLA interface. The models output a chunk of low-level robot actions, which are executed sequentially before the next policy query.

% Wan2-Policy 和 Cosmos-Predict2-Policy 使用基于世界模型的动作接口。对于这些模型，适配器根据官方模型格式组织当前的视觉观测和语言指令，并将预测的动作表示映射到 LIBERO 所使用的 7 自由度 Franka 控制空间和夹爪命令。所有模型均在相同的仿真器状态、任务指令、恢复场景、超时规则和成功判定标准下进行评估。
Wan2-Policy and Cosmos-Predict2-Policy use a world-model-based action interface. For these models, the adapter organizes the current visual observation and language instruction according to the official model format and maps the predicted action representation to the seven-degree of freedom Franka control space and the gripper command used by LIBERO. All models are evaluated under the same simulator states, task instructions, recovery scenarios, timeout rules, and success criterion.

% 表~\ref{tab:inference_config} 总结了各策略所使用的推理配置。除非另有说明，推理均在单张 NVIDIA H100 80GB GPU 上以 bfloat16 精度进行，控制频率为 20 Hz。
Table~\ref{tab:inference_config} summarizes the inference configuration used for each policy. Unless otherwise stated, inference is performed with bfloat16 precision on one NVIDIA H100 80GB GPU at a control frequency of 20 Hz.
\begin{table*}[t]
    \centering
    \caption{Inference configuration and LIBERO interface adaptation for
    the six evaluated policies.}
    \label{tab:inference_config}
    \scriptsize
    \setlength{\tabcolsep}{3.5pt}
    \renewcommand{\arraystretch}{1.08}
    \resizebox{\textwidth}{!}{
    \begin{tabular}{lccccc|c}
        \toprule
        \textbf{Model}
        & \textbf{Policy type}
        & \textbf{Input interface}
        & \textbf{Precision}
        & \textbf{Control frequency}
        & \textbf{Chunk size}
        & \textbf{Training steps} \\
        \midrule

        OpenVLA-OFT
        & VLA
        & Instruction + image + proprioception
        & BF16
        & 20 Hz
        & 8/16/32
        & 50K \\

        GR00T-N1.5
        & VLA
        & Instruction + image + proprioception
        & BF16
        & 20 Hz
        & 8/16/32
        & 30K \\

        $\pi_0$
        & VLA
        & Instruction + image + proprioception
        & BF16
        & 20 Hz
        & 8/16/32
        & 100K \\

        $\pi_0$-Fast
        & VLA
        & Instruction + image + proprioception
        & BF16
        & 20 Hz
        & 8/16/32
        & 30K \\

        Wan2-Policy
        & WAM
        & Instruction + visual observation
        & BF16
        & 20 Hz
        & 8/16/32
        & 60K \\

        Cosmos-Predict2-Policy
        & WAM
        & Instruction + visual observation
        & BF16
        & 20 Hz
        & 8/16/32
        & 50K \\
        \bottomrule
    \end{tabular}}
\end{table*}

% 中文：所有策略均预测 8 步的动作块。预测的动作块以 20 Hz 开环执行，每消耗完一个动作块后策略重新查询一次环境。动作块大小分析中仅改变每次查询所执行的动作数量，预测接口保持不变。
All policies predict an 8-step action chunk per query. The predicted chunk is executed open-loop at 20\,Hz, and the policy is re-queried after the chunk has been consumed, i.e., every eight control steps. In the chunk-size analysis of the main paper, only the number of executed actions per policy query is varied from 4 to 32, while the observation interface, prediction horizon, and rollout protocol remain unchanged.

% 调整主页，开始与附录打配合。比如我们上面对指标进行了详细讲解

% 1.文章没有说明 16 个 evaluation dimensions 的具体定义，这里需要用短篇幅补一点这样的设计，以防审稿人拿AI挑刺，人好好读就知道啥叫16维了，就是4个套件的4个L级别。
% 2.Key1--这里需要补充，图5的信息，指标的含义：
% 3.key2--表1的数据有点怪，
% 4.key3--每个模型的具体表现（同时图6 RSR RD中的M是怎么算的，是怎么定义RD的两个M）列一个表，然后对正文做补充的分析
% 5.key4--每个模型的具体表现

% 统一训练框架
\subsection{Unified Training of the Evaluated Policies}
\label{app:unified_training}

% 中文：为了保证公平比较，六个被评估的策略都在同一个统一的训练框架 StarVLA 中训练，共享相同的数据流水线、动作表示和评估接口。四个 VLA 策略（OpenVLA-OFT、π0、π0-Fast、GR00T-N1.5）与两个基于世界模型的策略（Wan2-Policy、Cosmos-Predict2-Policy）仅在视觉骨干与动作头结构上不同。
To ensure a controlled comparison, all six evaluated policies are trained within a single unified training framework~\cite{starvla2026} and share the same data pipeline, action representation, and rollout interface. The four VLA policies (OpenVLA-OFT, $\pi_0$, $\pi_0$-Fast, and GR00T-N1.5) and the two world-model-based policies (Wan2-Policy and Cosmos-Predict2-Policy) differ only in their visual backbones and action heads, as summarized in Table~\ref{tab:baseline_training}.

% 中文：所有策略均在初始 LIBERO 数据（四个任务套件训练集的并集，共 6,193 条示范、942,508 帧）上训练。其中 LIBERO-100 套件由 libero_10（379 条示范）与 libero_90（4,500 条示范）两部分共同构成。动作为 7 自由度增量末端位姿加夹爪指令，采用 min-max 归一化；每个策略接收 224×224 的第三人称与腕部相机图像以及机器人本体状态，控制频率 20 Hz。
All policies are trained on the initial LIBERO demonstrations, i.e., the union of the four training suites (\texttt{libero\_spatial}, \texttt{libero\_object}, \texttt{libero\_goal}, and \texttt{libero\_100}), comprising 6{,}193 episodes and 942{,}508 frames, where the LIBERO-100 suite jointly covers the \texttt{libero\_10} (379 episodes) and \texttt{libero\_90} (4{,}500 episodes) directories. Actions are 7-DoF delta end-effector poses with a gripper command, normalized with min--max statistics computed over the training set. Each policy receives synchronized third-person and wrist-camera images resized to $224\times224$ together with the robot proprioceptive state, and outputs an 8-step action chunk at the 20\,Hz control frequency.

\begin{table}[t]
    \centering
    \caption{Backbones and action heads of the six evaluated policies. All policies share the same LIBERO training data, action space, and evaluation interface. Training steps are reported in Table~\ref{tab:baseline_hyperparam}.}
    \label{tab:baseline_training}
    \scriptsize
    \setlength{\tabcolsep}{4pt}
    \renewcommand{\arraystretch}{1.15}
    \resizebox{\linewidth}{!}{
    \begin{tabular}{lcccc}
        \toprule
        \textbf{Policy} & \textbf{VLM backbone} & \textbf{Auxiliary backbone} & \textbf{Action head} \\
        \midrule
        OpenVLA-OFT & Qwen3-VL-4B-Instruct & -- & DiT-B \\
        $\pi_0$ & Qwen3-VL-4B-Instruct & DINOv2 ViT-S/14 & DiT-B flow-matching \\
        $\pi_0$-Fast & Qwen2.5-VL-3B-Instruct-Action & DINOv2 ViT-S/14 & FAST tokenizer \\
        GR00T-N1.5 & Qwen2.5-VL-3B-Instruct & DINOv2 ViT-S/14 & DiT-B flow-matching \\
        Wan2-Policy & Qwen3-VL-4B-Instruct & Wan2.2-TI2V-5B & MLP \\
        Cosmos-Predict2-Policy & Qwen3-VL-4B-Instruct & Cosmos-Predict2-2B & DiT-B flow-matching \\
        \bottomrule
    \end{tabular}}
\end{table}

% 中文：各策略家族的头部结构：扩散类动作头为 16 层 DiT-B（隐藏维 1024，交叉注意力维 2048），训练时使用 8 步流匹配，推理时使用 4 步去噪；FAST 类策略使用离散动作词元上的自回归解码；Wan2-Policy 通过多层感知机回归动作。
The flow-matching action heads used by $\pi_0$, GR00T-N1.5, and Cosmos-Predict2-Policy are 16-layer DiT-B transformers with a hidden size of 1024 and a cross-attention dimension of 2048, trained with 8 flow-matching steps and executed with 4 denoising steps at inference. $\pi_0$-Fast discretizes the action chunk with the FAST tokenizer and decodes action tokens autoregressively, and Wan2-Policy regresses the action chunk with an MLP head over world-model features.

% 中文：表 X 逐策略列出优化超参数。所有策略均使用 AdamW（β1=0.9，β2=0.95，eps=1e-8，权重衰减 1e-8），学习率采用余弦调度并衰减至最小学习率 1e-6，5000 步线性预热，梯度裁剪 1.0。学习率按模块分组设置：基础模块为 2.5e-5（OpenVLA-OFT、Wan2-Policy、Cosmos-Predict2-Policy）或 3e-5（π0、π0-Fast、GR00T-N1.5），VLM 接口为 1e-5，动作头为 1e-4。VLA 损失权重 1.0，VLM 协同训练损失权重 0.1。梯度检查点与 BF16 混合精度，随机种子 42。
Table~\ref{tab:baseline_hyperparam} reports the per-policy optimization hyperparameters. All policies are optimized with AdamW ($\beta_1{=}0.9$, $\beta_2{=}0.95$, $\epsilon{=}1\mathrm{e}{-8}$, weight decay $1\mathrm{e}{-8}$). The learning rate follows a cosine schedule that decays to a minimum learning rate of $1\mathrm{e}{-6}$ after 5{,}000 linear warmup steps, and gradients are clipped at a global norm of 1.0. Learning rates are assigned per module group: $2.5\mathrm{e}{-5}$ for the base modules of OpenVLA-OFT, Wan2-Policy, and Cosmos-Predict2-Policy, and $3\mathrm{e}{-5}$ for those of $\pi_0$, $\pi_0$-Fast, and GR00T-N1.5; the VLM interface uses $1\mathrm{e}{-5}$ and the action head uses $1\mathrm{e}{-4}$ in all cases. Training uses a per-device VLA batch size of 16 for the VLA policies and 8 for the two WAM policies, with a single gradient-accumulation step, gradient checkpointing, and BF16 mixed precision, under a fixed random seed of 42. The VLA imitation loss has weight 1.0, and the VLM co-training loss on image--grounding corpora has weight 0.1 to preserve visual--language alignment during action training.

\begin{table}[t]
    \centering
    \caption{Optimization hyperparameters of the six evaluated policies. All policies use AdamW, a cosine learning-rate schedule decaying to a minimum learning rate of $1\mathrm{e}{-6}$ with 5{,}000 linear warmup steps, gradient clipping at 1.0, BF16 mixed precision, and seed 42.}
    \label{tab:baseline_hyperparam}
    \scriptsize
    \setlength{\tabcolsep}{4pt}
    \renewcommand{\arraystretch}{1.15}
    \resizebox{\linewidth}{!}{
    \begin{tabular}{lccccc}
        \toprule
        \textbf{Policy} & \textbf{Base LR} & \textbf{VLM LR} & \textbf{Action LR} & \textbf{Batch/GPU} & \textbf{Steps} \\
        \midrule
        OpenVLA-OFT & $2.5\mathrm{e}{-5}$ & $1\mathrm{e}{-5}$ & $1\mathrm{e}{-4}$ & 16 & 50K \\
        $\pi_0$ & $3\mathrm{e}{-5}$ & $1\mathrm{e}{-5}$ & $1\mathrm{e}{-4}$ & 16 & 100K \\
        $\pi_0$-Fast & $3\mathrm{e}{-5}$ & $1\mathrm{e}{-5}$ & $1\mathrm{e}{-4}$ & 16 & 30K \\
        GR00T-N1.5 & $3\mathrm{e}{-5}$ & $1\mathrm{e}{-5}$ & $1\mathrm{e}{-4}$ & 16 & 30K \\
        Wan2-Policy & $2.5\mathrm{e}{-5}$ & $1\mathrm{e}{-5}$ & $1\mathrm{e}{-4}$ & 8 & 60K \\
        Cosmos-Predict2-Policy & $2.5\mathrm{e}{-5}$ & $1\mathrm{e}{-5}$ & $1\mathrm{e}{-4}$ & 8 & 50K \\
        \bottomrule
    \end{tabular}}
\end{table}

% 中文：按照各策略家族的官方训练配方，预训练 VLM 骨干在基线训练中保持冻结，仅更新动作头。
Following the official recipes of each policy family, all VLA is updated.

% 中文：表中报告的训练步数为每个策略发布检查点所对应的优化器步数，检查点按 LIBERO 验证性能选取。
The training steps reported in Tables~\ref{tab:inference_config} and~\ref{tab:baseline_hyperparam} denote the optimizer steps of the released checkpoint of each policy, selected by the standard LIBERO validation protocol of the corresponding recipe.

% 对应

% 恢复数据混合微调配置
\subsection{Recovery-Aware Fine-Tuning Configuration}
\label{app:recovery_finetune}

% 中文：正文中的 Combined 模型通过在原始 LIBERO 训练数据与 LIBERO-Recover 专家恢复轨迹的混合数据上联合微调得到。表 X 给出混合数据的组成，其中 LIBERO-100 将 libero_10 与 libero_90 两个目录合并统计。所有分量以相同权重采样、不做重加权，专家恢复轨迹约占混合数据帧数的 39.9%、episode 数的 34.0%。
The recovery-aware variants (\emph{Combined}) in the main paper are obtained by jointly fine-tuning the policies on a mixture of the original LIBERO training data and the LIBERO-Recover expert recovery trajectories. Table~\ref{tab:mix_data} details the composition of the mixture, in which LIBERO-100 aggregates the \texttt{libero\_10} and \texttt{libero\_90} directories. All components are sampled with equal weights and no re-weighting, so the expert recovery trajectories account for 39.9\% of the mixture frames and 34.0\% of its episodes.

\begin{table}[t]
    \centering
    \caption{Composition of the recovery-aware fine-tuning mixture. LIBERO-100 merges the \texttt{libero\_10} and \texttt{libero\_90} directories, and all components are sampled with equal weight.}
    \label{tab:mix_data}
    \scriptsize
    \setlength{\tabcolsep}{4pt}
    \renewcommand{\arraystretch}{1.15}
    \resizebox{\linewidth}{!}{
    \begin{tabular}{lccc}
        \toprule
        \textbf{Component} & \textbf{Episodes} & \textbf{Frames} & \textbf{Frame share} \\
        \midrule
        LIBERO-Spatial demonstrations & 432 & 52{,}970 & 3.4\% \\
        LIBERO-Object demonstrations & 454 & 66{,}984 & 4.3\% \\
        LIBERO-Goal demonstrations & 428 & 52{,}042 & 3.3\% \\
        LIBERO-100 (\texttt{libero\_10} + \texttt{libero\_90}) & 4{,}879 & 770{,}512 & 49.1\% \\
        Expert recovery trajectories & 3{,}184 & 625{,}731 & 39.9\% \\
        \midrule
        \textbf{Total} & \textbf{9{,}377} & \textbf{1{,}568{,}239} & \textbf{100\%} \\
        \bottomrule
    \end{tabular}}
\end{table}

% 中文：表 X 总结混合微调的完整配置：4 张 NVIDIA H20（96 GB）GPU，accelerate + DeepSpeed ZeRO Stage 2 分布式训练；每卡批量 1、梯度累积 4 步，有效批量 16；共 50,000 个优化器步，约相当于混合数据的 0.5 个 epoch。
Table~\ref{tab:finetune_config} summarizes the complete fine-tuning configuration. Fine-tuning is performed on 4 NVIDIA H20 (96\,GB) GPUs with DeepSpeed ZeRO Stage 2 distributed training. The per-device batch size is 1 with 4 gradient-accumulation steps, yielding an effective batch size of 16, and training runs for 50{,}000 optimizer steps, corresponding to roughly half an epoch over the 1{,}568{,}239-frame mixture.

\begin{table}[t]
    \centering
    \caption{Recovery-aware fine-tuning configuration shared by the \emph{Combined} variants.}
    \label{tab:finetune_config}
    \scriptsize
    \setlength{\tabcolsep}{4pt}
    \renewcommand{\arraystretch}{1.2}
    \resizebox{\linewidth}{!}{
    \begin{tabular}{ll}
        \toprule
        \textbf{Item} & \textbf{Setting} \\
        \midrule
        Hardware & 4$\times$ NVIDIA H20 (96\,GB) \\
        Distributed training & accelerate + DeepSpeed ZeRO Stage 2 \\
        Effective batch size & 16 (4 GPUs $\times$ 1 per-device $\times$ 4 accum.) \\
        Optimizer & AdamW ($\beta_1{=}0.9$, $\beta_2{=}0.95$, $\epsilon{=}1\mathrm{e}{-8}$, wd $1\mathrm{e}{-8}$) \\
        Learning rate (base / VLM / action head) & $2.5\mathrm{e}{-5}$ / $1\mathrm{e}{-5}$ / $1\mathrm{e}{-4}$ \\
        LR schedule & cosine decay to $1\mathrm{e}{-6}$, 3{,}000 warmup steps \\
        Training steps & 50{,}000 ($\approx$0.5 epoch over the mixture) \\
        Checkpointing & save 10{,}000 / eval 5{,}000 steps \\
        Frozen modules & vision encoder only \\
        Precision & BF16, gradient checkpointing, clip 1.0 \\
        Wall-clock time & $\approx$5\,h ($\approx$1\,h per 10{,}000 steps) \\
        Random seed & 42 \\
        \bottomrule
    \end{tabular}}
\end{table}

% 中文：与基线训练（整个 VLM 骨干冻结）不同，混合微调仅冻结视觉编码器，语言模型与动作头均参与更新，以便模型适应恢复场景中的状态分布。检查点每 10,000 步保存一次，每 5,000 步评估一次；最终采用第 50,000 步的检查点。
Unlike baseline training, where the full VLA is updated, the recovery-aware fine-tuning freezes only the vision encoder, and both the language model and the action head are updated, allowing the policy to adapt to the state distribution induced by failures. Checkpoints are saved every 10{,}000 steps and evaluated every 5{,}000 steps; the checkpoint at step 50{,}000 is used for all reported results.

% 中文：GR00T-N1.5 的混合微调沿用与 OpenVLA-OFT 完全相同的协议（数据混合、优化器、调度、步数与冻结策略），仅框架相关设置继承自其在初始 LIBERO 数据上的训练配置（Qwen2.5-VL-3B 骨干、DINOv2 视觉编码、DiT-B 流匹配动作头以及 3e-5 的基础学习率）。
The GR00T-N1.5 recovery-aware run adopts exactly the same protocol as the OpenVLA-OFT run, i.e., the identical data mixture, optimizer, schedule, number of steps, and freezing strategy, while its framework-specific settings are inherited from its initial-LIBERO recipe: the Qwen2.5-VL-3B backbone with the DINOv2 visual encoder, the DiT-B flow-matching action head, and a base learning rate of $3\mathrm{e}{-5}$.

% 中文：temporal 变体使用相同的训练配方，区别在于训练与评估时额外以失败前时序历史（当前帧之前的 8 帧观测）作为时序上下文提供给策略。
The \emph{temporal} variants are trained with the same recipe, except that the pre-failure temporal history, i.e., the 8 observation frames preceding the current frame, is additionally provided to the policy as temporal context during both training and evaluation.

\section{Failure Analysis}

% 中文：从数据集构建过程到恢复实验（见 Sec.~\ref{sec:construction} 和 Sec.~\ref{exp}），我们观察到了一些反复出现的现象：策略产生的失败并非完全随机，而是可以归纳为若干具有代表性的失败模式。这些失败不仅与底层控制不够精确有关，也反映出当执行状态发生变化后，策略难以及时更新原有的动作规划。
From the dataset construction process to the recovery experiments (Secs.~\ref{sec:construction} and~\ref{exp}), we observed several recurring patterns in the failures produced by the evaluated policies. These failures are not entirely random and can be roughly grouped into a small number of representative failure modes. They are related not only to imprecise low-level control, but also to the difficulty of updating the original action plan after the execution state has changed.

\subsection{Failure to Verify Action Outcomes}
\label{app:action_outcome_verification}

\begin{figure*}[t]
    \centering
    \includegraphics[width=\textwidth]{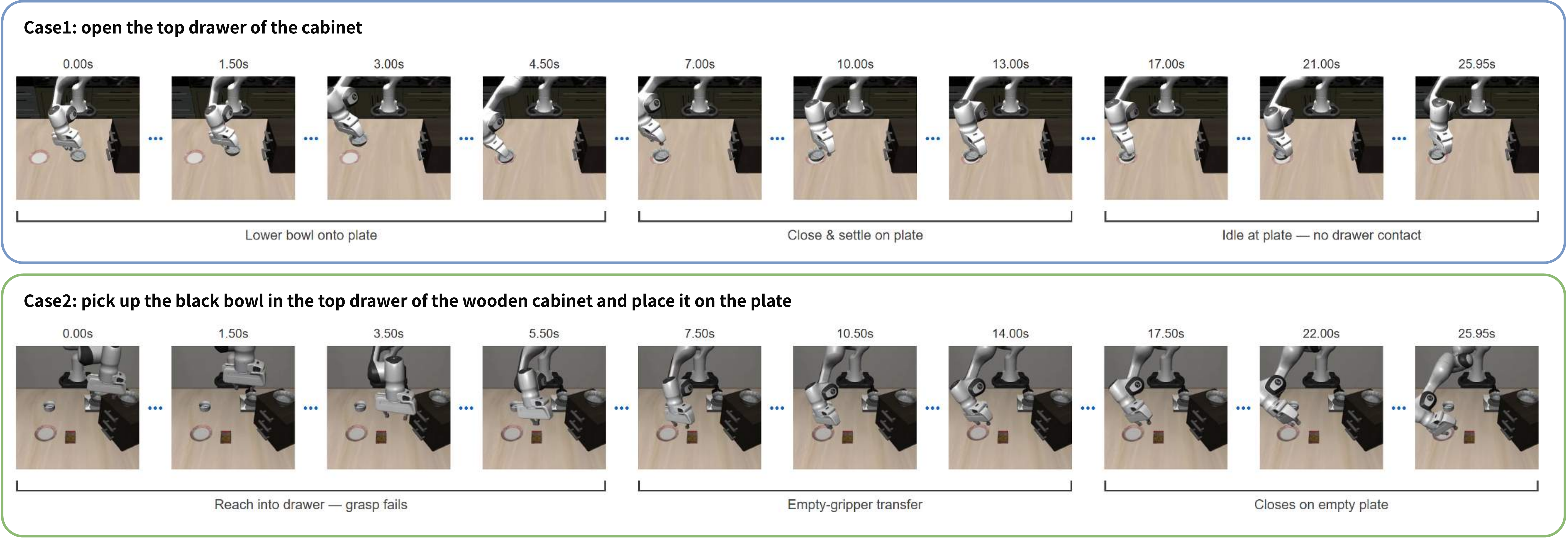}
    \caption{Representative failure cases in VLA rollouts.}
    \label{fig:failure_cases}
\end{figure*}

% 第一个问题是动作结果与后续动作之间缺乏一致性：即使前一步动作未达到预期结果，策略仍可能继续执行原本的动作序列。已有研究指出，当前 VLA 存在视觉定位与动作生成脱节的问题。例如，ReconVLA 发现模型的视觉注意力未能充分聚焦于操作目标，而 Tsai 等人的研究表明，直接向动作头提供目标的三维位置信息能够改善空间与任务泛化。这些发现提示，策略可能过度依赖训练中熟悉的动作模式，而未充分利用当前观测调整动作。在收集失败场景的过程中，我们也观察到了两个具有代表性的案例，如图~\ref{fig:failure_cases} 所示。
The first issue is a mismatch between action outcomes and subsequent behavior: a policy may continue the original action sequence even when the preceding action has not achieved its intended result. Prior studies have identified related limitations in grounding action generation in visual observations. ReconVLA~\cite{song2025reconvla} reports that visual attention is insufficiently focused on manipulation targets, while Tsai et al.~\cite{tsai2026grounded3d} show that directly providing target 3D locations to the action head improves spatial and task generalization. These findings suggest that policies may rely too heavily on familiar action patterns without sufficiently using current observations to adjust their actions. During failure-scenario collection, we observed two representative cases, shown in Figure~\ref{fig:failure_cases}.

% 中文：在图~\ref{fig:failure\_cases} 的第一个案例中，任务要求打开橱柜顶部抽屉，但机器人随后在盘子附近执行了放置动作。视频中没有看到与打开抽屉相关的有效接触，后续动作也没有根据当前场景进行调整。
In the first example, the instruction is \texttt{open\_the\_top\_drawer\_of\_the\_cabinet}, but the robot subsequently performs a placement motion near the plate. The video does not show an effective interaction with the drawer, and the later actions are not adjusted according to the state observed in the scene.

% 中文：第二个案例中，机器人没有抓住黑色碗，却继续沿着朝向盘子的搬运轨迹移动空夹爪。此时问题并不只是抓取没有成功，而是后续动作默认了一个并未发生的事件，即机器人已经持有碗。
In the second example, the robot fails to grasp the black bowl but continues moving the empty gripper along the transport trajectory toward the plate. The problem is therefore not only the missed grasp. The later motion assumes that the bowl has already been acquired, although this prerequisite has not been satisfied.

% 中文：这两个案例表明，模型没有稳定地检查“动作是否产生了预期结果”。一旦抓取、放置或接触没有成功，后续动作仍可能按照原始计划继续执行，导致错误动作被连续放大。
These two cases show that the policy does not reliably check whether an action produced its intended result. When a grasp, placement, or contact fails, the policy may still follow the original plan, allowing the initial error to propagate through the remaining actions.

% 细粒度控制不足
\subsection{Insufficient Fine-Grained Control}
\label{app:fine_grained_control}

% 中文：第二类失败出现在物体仍然可见，但其位置、朝向或可抓取区域已经发生变化的情况下。这里展示了两个 L2 级别的失败案例，均来自 LIBERO 与 LIBERO-Recover 联合训练后的模型。可以看到，在 recovery 场景中，模型需要根据变化后的物体构型选择更加具体的抓取角度和接近方式，因而对细粒度控制提出了更高要求。
The second type of failure occurs when the target object remains visible, but its position, orientation, or graspable region has changed. We show two $L_2$ failure cases produced by a model jointly trained on LIBERO and LIBERO-Recover, as reported in the main results. In the recovery scenarios, the model must select a specific grasping angle and approach direction according to the changed object configuration, which places greater demands on fine-grained control.

\begin{figure*}[t]
    \centering
    \includegraphics[width=\textwidth]{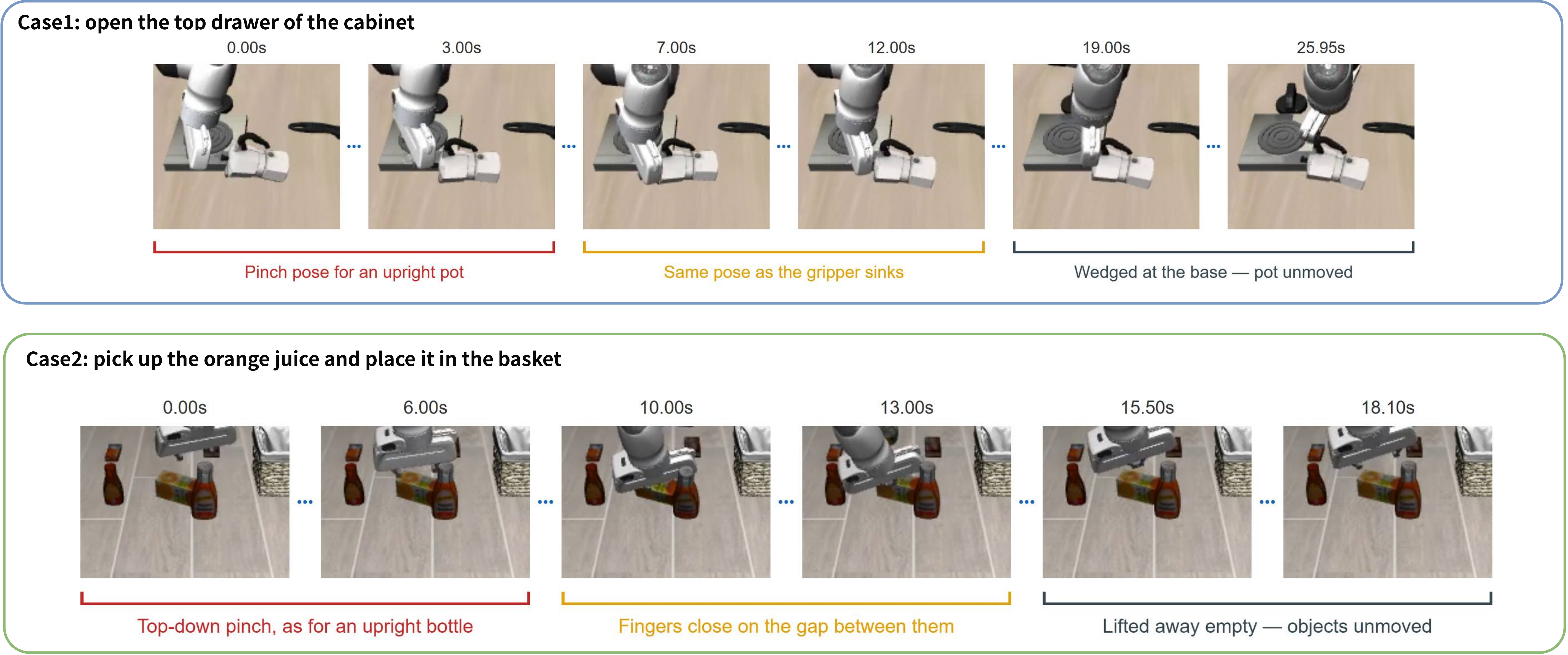}
    \caption{Two fine-grained grasping failures after object configuration changes.}
    \label{fig:orange_juice_recovery}
\end{figure*}

% 中文：图~\ref{fig:fine_grained_control_cases} 展示了两个代表性示例。橙汁容器和摩卡壶侧倒后，它们的高度、朝向以及可接近的抓取区域均发生了变化。然而，策略仍然采用适用于原始直立构型的抓取动作接近物体。在数次不成功的尝试后，夹爪轨迹逐渐偏离目标，但策略没有识别出能够建立稳定接触的新抓取位置或接近方向。
Figure~\ref{fig:orange_juice_recovery} shows two representative examples. After the orange-juice container and the moka pot fall onto their sides, their height, orientation, and accessible grasp regions change. However, the policy continues to approach them with grasping motions that appear suitable for their original upright configurations. After several unsuccessful attempts, the gripper trajectory increasingly deviates from the target, but the policy does not identify a new grasp location or approach direction that could establish stable contact.

% 中文：这类失败案例说明，仅仅检测到目标物体并不足以完成恢复。策略还需要根据当前构型判断哪些位置可供抓取，以及机械臂应从哪个方向接近。对于已经发生移位或旋转的物体，这一步通常决定了后续恢复是否能够继续。
These failure cases show that detecting the target object is not sufficient for recovery. The policy must also determine which regions are graspable under the current configuration and from which direction the robot should approach. For an object that has been displaced or rotated, this decision often determines whether recovery can proceed.

% 缺失前置恢复动作
\subsection{Recovery Requires an Intermediate Action}
\label{app:missing_prerequisite_actions}

% 中文：第三，部分失败无法通过直接重复标称任务动作来解决。机器人必须首先执行前置恢复动作，例如重新放置物体、释放夹爪或清除障碍物。这里展示了两个 L4 级别的失败案例，均来自 LIBERO 与 LIBERO-Recover 联合训练后的模型。可以看到，在 recovery 场景中，模型虽然已经接受了恢复轨迹的训练，但仍然倾向于直接执行原始任务动作，而没有先恢复原始任务所依赖的物体状态。
Third, some failures cannot be resolved by directly repeating the nominal task action. The robot must first perform a prerequisite recovery action, such as repositioning an object, releasing the gripper, or removing an obstruction. We show two $L_4$ failure cases produced by a model jointly trained on LIBERO and LIBERO-Recover data. Although the model has been exposed to recovery trajectories, it still tends to execute the original task action directly in the recovery scenes instead of first restoring the object state required by the task.

\begin{figure*}[t]
    \centering
    \includegraphics[width=\textwidth]{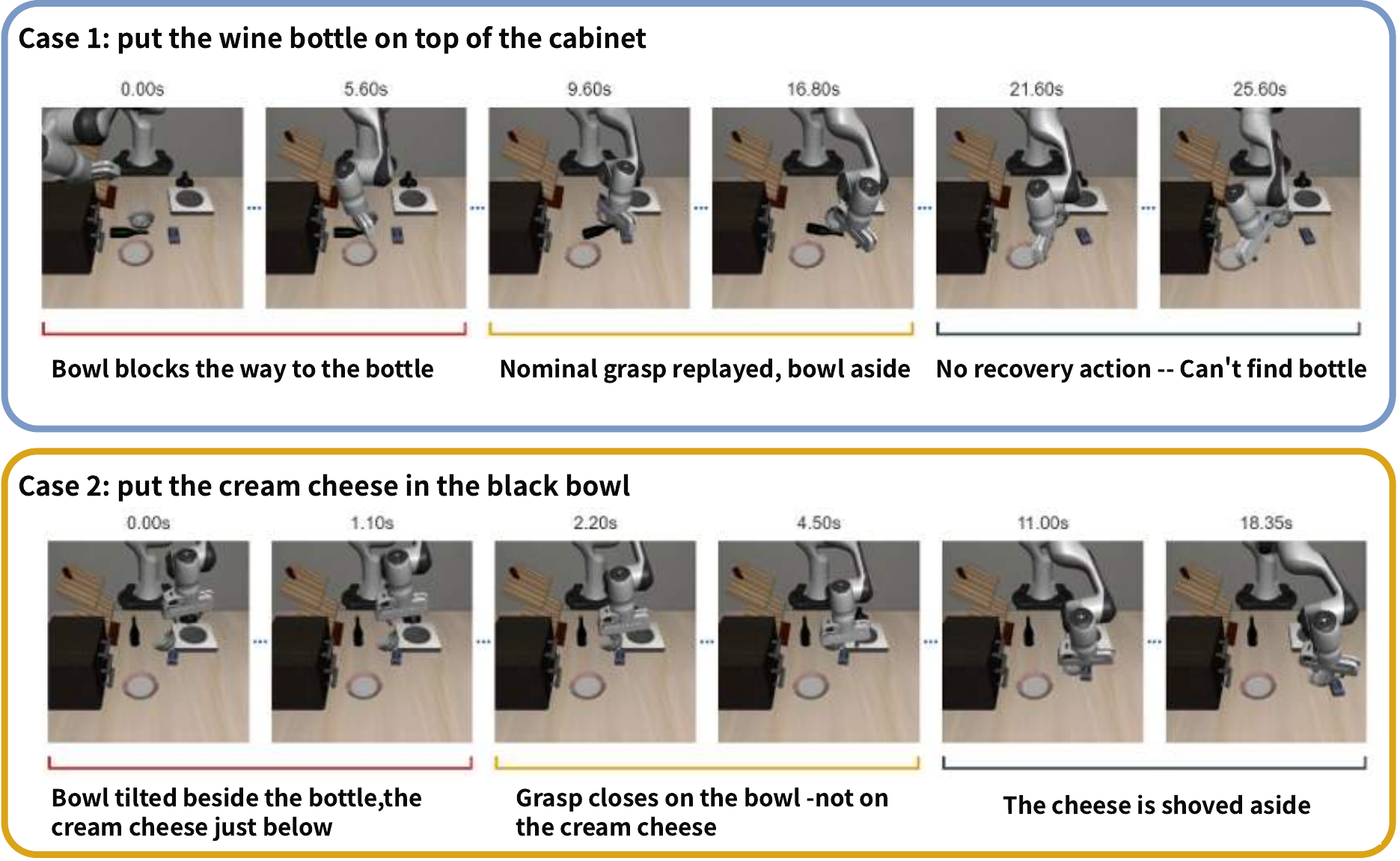}
    \caption{Examples of recovery failures that require prerequisite actions.}
    \label{fig:prerequisite_recovery}
\end{figure*}

% 中文：在图~\ref{fig:prerequisite_recovery}(a) 所示的奶油奶酪案例中，初始失败状态使机器人抓住了碗而不是奶油奶酪。恢复推演开始时，夹爪已经打开，但碗仍然靠近奶油奶酪。模型直接重复抓取奶油奶酪的动作，结果多次抓到碗，而没有先将碗移开或重新建立两个物体之间合适的空间关系。
In the cream-cheese example in Figure~\ref{fig:prerequisite_recovery}(a), the initial failure leaves the robot holding the bowl instead of the cream cheese. At the beginning of the recovery rollout, the gripper is open, but the bowl remains close to the cheese. The model directly repeats the grasping motion toward the cheese and repeatedly contacts the bowl, without first moving the bowl aside or restoring a suitable spatial relation between the two objects.

% 中文：在图~\ref{fig:prerequisite_recovery}(b) 所示的酒瓶案例中，一个碗挡住了机器人接近酒瓶的路径。要完成原始任务，机器人应先将碗移开，再重新规划酒瓶的抓取动作。然而，模型直接执行了原始的抓瓶动作，并在接近过程中与碗发生了额外碰撞，使物体状态进一步偏离可恢复构型。
In the wine-bottle example in Figure~\ref{fig:prerequisite_recovery}(b), a bowl blocks the robot's approach to the bottle. To complete the original task, the robot should first move the bowl aside and then replan the grasp for the bottle. Instead, the model directly executes the original bottle-grasping motion and makes additional contact with the bowl, causing the object configuration to move farther from a recoverable state.

% 中文：这两个案例中的共同问题不是模型完全没有看到目标物体，而是没有按照恢复所需的顺序组织动作。模型需要先建立一个中间状态，例如释放被错误抓住的物体、分离相互遮挡的物体或清除接近路径上的障碍，然后才能重新执行原始任务动作。
The common problem in these two cases is not that the model fails to see the target object. Rather, it does not organize the actions in the order required for recovery. The model must first establish an intermediate state, for example by releasing an incorrectly held object, separating objects that interfere with each other, or clearing the approach path, before attempting the original task action again.

% 中文：这些案例表明，L3 恢复要求模型首先判断哪些状态需要被修改，才能重新满足原始任务的执行条件。如果缺少这种前置条件推理，直接重复原始动作不仅无法完成任务，还可能引入新的碰撞和物体移位，使场景进一步偏离熟悉的状态，增加后续恢复的难度。因此，模型需要具备一定的状态评估与动作排序能力，才能在不断变化的交互环境中保持更强的鲁棒性。
These cases show that $L_3$ recovery first requires the model to determine which states must be changed to restore the conditions for the original task. Without this prerequisite reasoning, directly repeating the original action may fail to complete the task and may introduce additional collisions or object displacement, making the scene increasingly different from familiar configurations and further complicating recovery. Robust behavior in such settings therefore requires the model to assess the current state and order its corrective actions before resuming the nominal task.

\subsection{Recovery Degradation under Repeated Failures}
\label{app:progressive_degradation}

% 中文：为了直观展示 RD 所反映的性能变化，我们选取了 GR00T 在任务 \texttt{stack\_the\_middle\_black\_bowl\_on\_the\_back\_black\_bowl} 中的一条失败轨迹。沿着原始轨迹，我们每隔 20 个控制步保存一次仿真器状态，并从每个记录状态重新运行同一个任务。图中的箭头表示从对应状态重新开始推演后的任务结果。
To qualitatively illustrate the performance change measured by RD, we select a failed GR00T rollout on the task \texttt{stack\_the\_middle\_black\_bowl\_on\_the\_back\_black\_bowl}. Along the original rollout, we save the simulator state every 20 control steps and rerun the same task from each recorded state. The arrows in the figure indicate the outcome of the rerun initialized from the corresponding state.

\begin{figure*}[t]
    \centering
    \includegraphics[width=\textwidth]{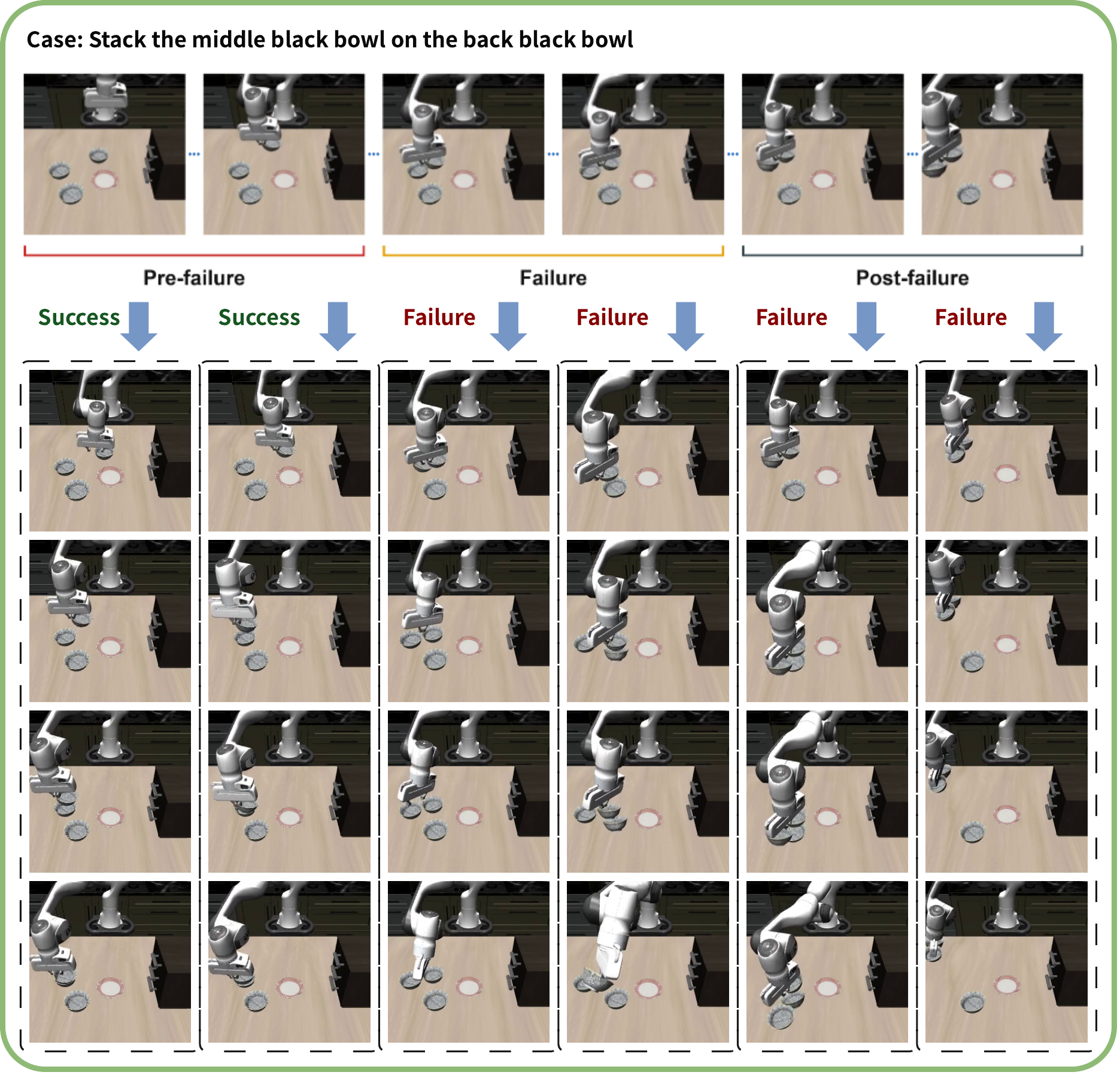}
    % 中文：重复失败后的恢复退化案例。我们从 GR00T 在任务 \texttt{stack\_the\_middle\_black\_bowl\_on\_the\_back\_black\_bowl} 中的一条失败轨迹出发，每隔 20 个控制步记录一次状态，并从各状态重新运行任务。Pre-failure 状态重新运行成功，而 Failure 和 Post-failure 状态重新运行失败。
    \caption{Qualitative illustration of recovery degradation after failure.}
    \label{fig:ualitative}
\end{figure*}

% 中文：图~\ref{fig:rd\_qualitative} 将这条轨迹划分为 Pre-failure、Failure 和 Post-failure 三个阶段。前两个状态仍处于失败发生之前，从这些状态重新运行时任务可以成功完成。进入 Failure 阶段后，策略已经无法从当前状态完成原始任务；在 Post-failure 阶段，重新运行仍然失败。
Figure~\ref{fig:ualitative} divides the rollout into three phases: Pre-failure, Failure, and Post-failure. The first two checkpoints are taken before the failure occurs, and the task can still be completed when rerun from these states. After entering the Failure phase, the policy can no longer complete the original task from the resulting state, and the reruns remain unsuccessful in the Post-failure phase.

% 中文：这个案例说明，失败会改变的不只是某一个物体的位姿，还可能改变物体在场景中的分布、相互之间的空间关系以及机器人后续可用的接近路径。随着执行继续进行，原本可以完成任务的场景逐渐变成策略不熟悉的构型，模型需要重新判断物体的位置关系和可行的操作方式。
This case shows that failure changes more than the pose of an individual object. It can also alter the distribution of objects in the scene, their spatial relations, and the approach paths available to the robot. As execution continues, a scene that was initially solvable gradually becomes less familiar to the policy, which must reassess object relations and feasible ways of interacting with them.

% 中文：因此，这一案例为正文中 RD 从 During-Failure 到 Post-Failure 的进一步下降提供了直观的定性解释。失败后的重复交互没有把场景恢复到更容易执行的状态，反而使后续恢复面临更大的状态偏移和空间推理压力。该案例用于说明 RD 的行为含义，而不是单独估计 RD 指标。
This example provides a qualitative explanation for the further decrease in RD-related performance from During-Failure to Post-Failure reported in the main paper. The repeated interactions after failure do not return the scene to an easier configuration; instead, they introduce larger state shifts and stronger spatial reasoning demands for subsequent recovery. This case is intended to illustrate the behavior captured by RD rather than to estimate the metric independently.

% 失败恢复中的挑战
\section{Challenges in Failure Recovery}
\label{app:recovery_challenges}

% 中文：恢复任务的难点不只在于场景中物体发生了变化。失败还可能改变夹爪状态、末端执行器位姿以及机器人与物体之间的接触关系。
The difficulty of recovery does not come only from changes in the scene. A failure may also change the gripper state, the end-effector pose, and the contact relationship between the robot and the objects.

% 场景与机器人状态变化
\subsection{Scene and Robot State Changes}
\label{app:scene_robot_changes}

% 中文：执行失败通常同时改变外部场景和机器人自身的交互状态。为了说明这两类变化，我们选取了四个收集到的失败案例，其中两个主要表现为场景和物体状态的改变，另外两个主要表现为机器人末端位姿的改变。
Execution failures can change both the external scene and the robot's own interaction state. To illustrate these two types of changes, we select the collected failure cases. 
% 场景变化。
\paragraph{Scene changes.}
% 中文：在前两个案例中，物体发生了移位或旋转，原有的支撑关系、接触关系和接近路径随之发生变化。即使任务指令和目标物体保持不变，机器人也不能直接沿用失败前的操作方式，而需要重新判断物体之间的空间关系以及可行的接近区域。
Objects are displaced or rotated, which changes their support relations, contact relations, and approach paths. Although the task instruction and target object remain unchanged, the robot cannot directly reuse the manipulation strategy from before the failure. It must reassess the spatial relations between objects and identify a feasible approach region.

% 机器人与交互状态变化。
\paragraph{Robot state changes.}
% 中文：后两个案例中的主要变化发生在机器人自身。失败后，机械臂末端可能停留在不自然的姿态，或者以较大的偏转角度靠近工作空间边界。此时，物体本身未必发生明显移动，但后续控制动作已经不再容易平滑执行，原本适合的动作序列也可能因此失效。
The main change occurs in the robot itself. After failure, the end effector may remain in an awkward pose or approach the workspace boundary with a large orientation deviation. The objects may not have moved substantially, but the subsequent actions are no longer easy to execute smoothly, and the original action sequence may become ineffective.

% 中文：这一点在遥操作过程中也较为明显。在少数案例中，遥操作员并没有立即操作目标物体，而是先调整机器人的末端位置和朝向，使机械臂重新回到更容易控制的状态，然后再继续执行恢复动作。这说明恢复不仅需要处理场景中的物体变化，有时还需要先恢复机器人自身的可操作状态。
This effect is also visible during teleoperation. In a small number of cases, the teleoperator does not immediately manipulate the target object, but first adjusts the end-effector position and orientation to bring the robot back to a more controllable configuration before continuing the recovery action. Recovery therefore sometimes requires restoring the robot's own operable state in addition to handling changes in the scene.

% 中文：这四个案例说明，失败后的状态变化同时具有显性的场景变化和相对隐蔽的机器人状态变化。前者要求策略重新理解物体及其空间关系，后者要求策略判断当前末端位姿是否仍然适合执行下一步动作。忽略任意一类变化，都可能导致后续动作难以执行。
These cases show that post-failure changes include both visible changes in the scene and less obvious changes in the robot state. The former requires the policy to reinterpret the objects and their spatial relations, while the latter requires it to determine whether the current end-effector pose is still suitable for the next action. Ignoring either type of change can make subsequent actions difficult to execute.

\begin{figure*}[t]
    \centering
    % 替换为实际图片路径。
    \includegraphics[width=\textwidth]{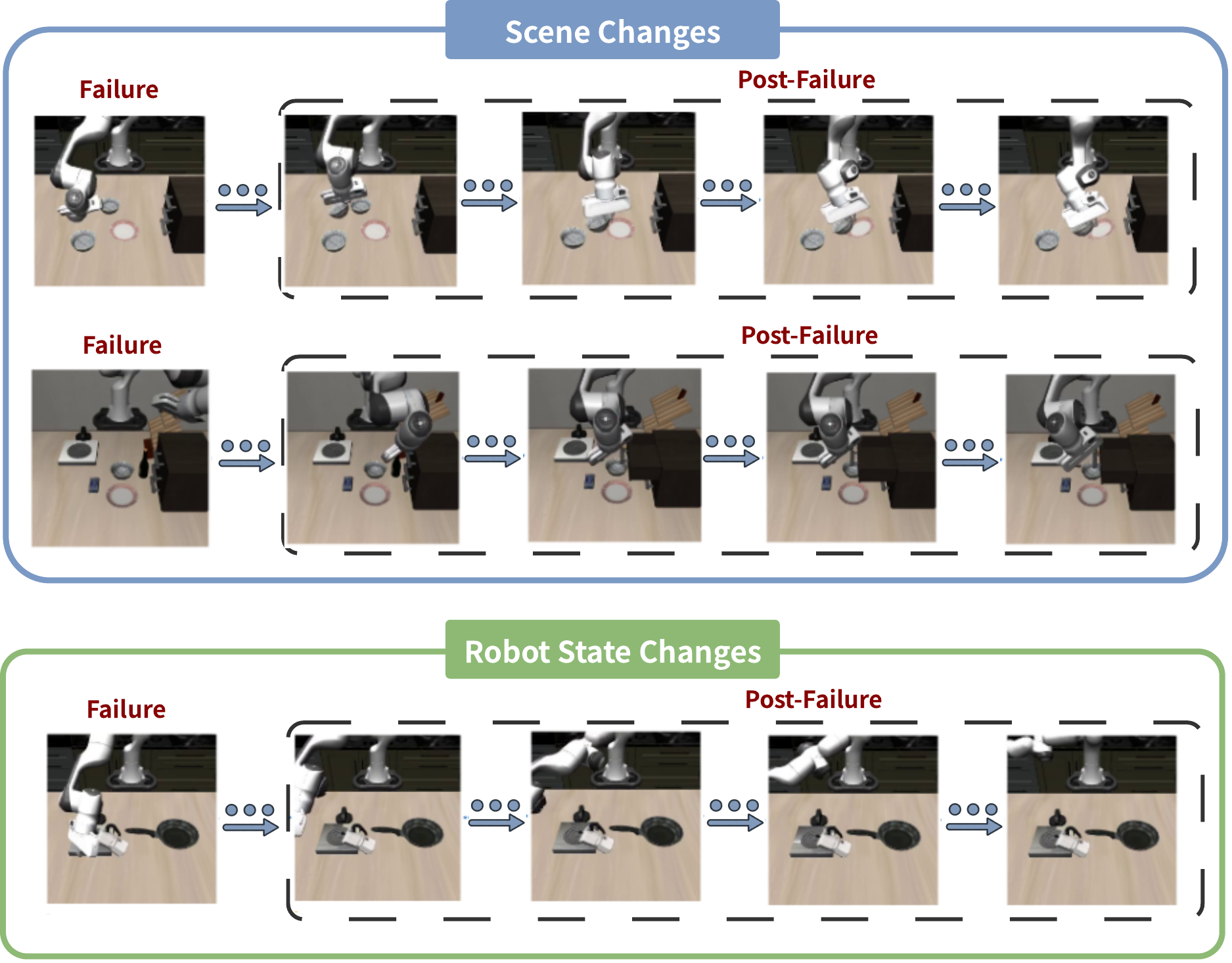}
    % 中文：失败后的场景与机器人状态变化。前两个案例展示物体位姿及物体间关系的变化，第三个案例展示不利于后续操作的机器人姿态。
    \caption{Scene and robot-state changes after failure. The first two cases show changes in object poses and relations between objects. The third shows an awkward robot configuration that hinders subsequent manipulation.}
    \label{fig:scene_robot_changes}
\end{figure*}

% 不同物体构型下的抓取
\subsection{Grasping under Different Object Configurations}
\label{app:recovery_affordances}

\begin{figure*}[t]
    \centering
    \includegraphics[width=\textwidth]{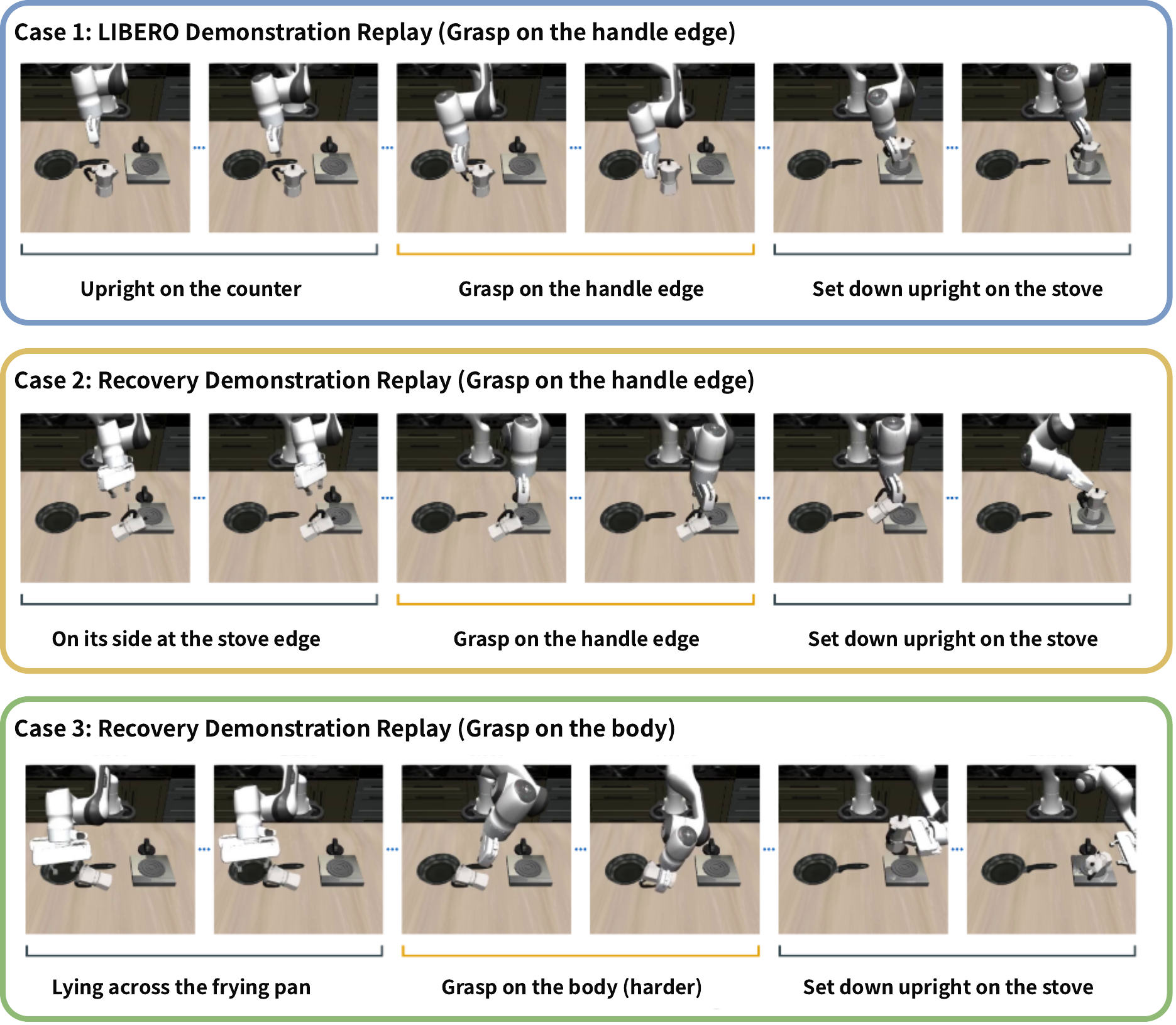}
    % 已录制示范轨迹的回放：Case 1来自LIBERO官方数据集，Case 2和Case 3来自我们采集的遥操作恢复示范。
    \caption{Replay of recorded demonstrations: Case 1 uses an official
    LIBERO demonstration, while Cases 2 and 3 use our teleoperated
    recovery demonstrations.}
    \label{fig:moka_recovery_states}
\end{figure*}

% 中文：在不同的失败后状态下，同一任务指令可能需要不同的恢复动作。为直观展示这一点，我们选取任务 ``put the moka pot on the stove''，并分别回放 LIBERO 官方轨迹数据和我们采集的恢复轨迹数据。图~\ref{fig:moka\_recovery\_states} 展示了三段轨迹中抓取摩卡壶的关键片段，其中 Case 1 来自 LIBERO 官方数据，Case 2 和 Case 3 来自我们的遥操作恢复数据。
The same task instruction may require different recovery actions under different post-failure states. To illustrate this point, we select the task ``put the moka pot on the stove'' and replay both an official LIBERO trajectory and our recorded recovery trajectories. Figure~\ref{fig:moka_recovery_states} shows the grasping segments from three replayed trajectories: Case 1 is taken from the official LIBERO~\cite{libero} data, whereas Cases 2 and 3 are from our teleoperated recovery data.

% 中文：三段轨迹的最终目标相同，但摩卡壶在抓取阶段的位置、朝向以及与周围物体的相对关系并不一致。因此，适用于标准初始构型的抓取位置和接近方向，在失败后的其他构型中可能不再有效。
Although the final task goal is identical across the three trajectories, the position, orientation, and relative spatial relations of the moka pot during grasping are different. A grasp location or approach direction that is suitable for the standard initial configuration may therefore become ineffective under another post-failure configuration.

% 中文：从图中可以看到，机器人需要根据摩卡壶当前的姿态选择不同的末端接近方式，而不是简单地复用同一段抓取动作。某些构型允许夹爪直接从上方接近，另一些构型则要求机器人调整腕部朝向，或从侧面寻找更容易建立稳定接触的位置。
As shown in the figure, the robot must choose the end-effector approach according to the current pose of the moka pot rather than replaying the same grasping motion. Some configurations allow the gripper to approach from above, whereas others require a different wrist orientation or a lateral approach to establish stable contact.

% 中文：这个回放对比说明，恢复任务中的抓取动作不能只由任务指令和物体类别决定，还需要结合当前物体构型及周围可用空间进行判断。也就是说，即使任务目标没有改变，失败后的有效恢复动作仍可能与原始成功轨迹中的动作存在明显差异。
This replay comparison shows that grasping in recovery cannot be determined solely by the task instruction and object category. The policy must also consider the current object configuration and the available space around it. Thus, even when the task goal remains unchanged, the effective recovery action may differ substantially from the action used in the original successful trajectory.

% 对恢复性能的启示
\subsection{Implications for Recovery Performance}
\label{app:recovery_performance_discussion}

% 中文：上述案例说明，恢复推演与标准 LIBERO 评测之间的差别并不只是初始物体位置发生了变化。策略还需要判断失败后哪些物体关系仍然有效、当前末端位姿是否适合继续操作，以及原始抓取动作是否仍然可执行。
The cases above show that recovery differs from standard LIBERO evaluation in more than the initial object positions. The policy must determine which object relations remain usable after failure, whether the current end-effector pose is suitable for further interaction, and whether the original grasping action is still executable.

% 中文：当物体发生移位或旋转时，策略需要重新建立对场景的空间理解；当机器人末端处于不利姿态时，策略还需要先恢复自身的可操作状态。这些变化可能同时出现，使得策略不能只根据当前语言指令直接生成原始动作，而需要根据实际状态重新安排动作顺序。
When an object is displaced or rotated, the policy must reconstruct its spatial understanding of the scene. When the robot is left in an unfavorable pose, it may also need to restore its own operable configuration first. These changes can occur together, so the policy cannot simply generate the original action from the language instruction. It must reorder its actions according to the actual state.

% 中文：这一点也解释了为什么 L1/L2 与 L3/L4 之间存在明显的性能差异。L1 和部分 L2 失败通常只需要重试或调整下一步动作，而 L3 和 L4 失败往往需要先恢复物体、机器人或环境状态，再继续执行原始任务。恢复过程中增加的中间步骤也为后续动作引入了更多可能的失败点。
This observation helps explain the performance difference between $L_1/L_2$ and $L_3/L_4$. Many $L_1$ failures and some $L_2$ failures can be handled by retrying or adjusting the next action, whereas $L_3$ and $L_4$ failures often require the robot to restore an object, robot, or environmental state before resuming the original task. These additional intermediate steps introduce more opportunities for failure during recovery.

% 中文：上述案例并不能单独量化场景变化、机器人状态变化和抓取精度对最终成功率的贡献，但它们与正文中观察到的 RSR 下降和 RD 增大是一致的。恢复任务要求策略持续检查动作结果，并根据新的状态更新后续计划，而不是将成功轨迹中的动作顺序完整重放。
The cases above do not separately quantify the contribution of scene changes, robot-state changes, and grasping precision to the final success rate. However, they are consistent with the lower RSR and higher RD observed in the main experiments. Recovery requires the policy to continuously check action outcomes and update its subsequent plan from the new state, rather than replaying the action sequence from a successful trajectory.

\section{Limitations and Discussion}
\label{app:limitations}

%LIBERO-Recover 当前在仿真环境中构建，失败定位和类别标注也采用了视觉语言模型辅助并经过人工复核。因此，结果可能无法完全反映真实机器人中的传感器噪声、控制误差和硬件限制，且自然失败分布的不均衡也使高难度恢复等级的样本相对较少。
LIBERO-Recover is currently constructed in simulation, with failure localization and categorization assisted by a vision-language model and verified by human reviewers. The results may therefore not fully reflect sensor noise, control errors, or hardware limitations in physical robots. In addition, the naturally imbalanced failure distribution leads to fewer examples at the higher recovery levels.

% 中文：尽管如此，基准中的定量结果和案例分析共同说明，失败后的恢复需要模型重新理解物体状态、空间关系和机器人姿态，而不能只重复原始动作。后续工作将进一步在真实机器人上验证这些现象，并将评测扩展到不同机械臂和多机械臂协同场景。
Nevertheless, the quantitative results and qualitative cases consistently show that recovery requires the policy to reinterpret object states, spatial relations, and robot pose after failure rather than simply repeat the original action. Future work will validate these observations on physical robots and extend the evaluation to different robot arms and multi-arm cooperative manipulation.

\end{document}